\documentclass[review]{elsarticle}

\usepackage[margin=0.618in]{geometry}
\usepackage{subcaption}
\usepackage{multirow}
\usepackage{enumitem}
\usepackage{amsmath}
\usepackage{url}
\usepackage{float}
\usepackage{graphicx}
\usepackage{fancyhdr}
\usepackage[colorlinks,linkcolor=blue]{hyperref}
\journal{Journal of \LaTeX\ Templates}

\graphicspath{{figs/}{images/}}

\begin{document}

\begin{frontmatter}

\title{Efficient Misalignment-Robust Multi-Focus Microscopical Images Fusion}

\author{Yixiong Liang}
\ead{yxliang@csu.edu.cn}
\author{Yuan Mao}
\ead{yuanmao@csu.edu.cn}
\author{Zhihong Tang}
\ead{zhihongtang@csu.edu.cn}
\author{Meng Yan}
\ead{bryant@csu.edu.cn}
\author{Yuqian Zhao}
\ead{zyq@csu.edu.cn}
\author{Jianfeng Liu\corref{mycorrespondingauthor}}
\cortext[mycorrespondingauthor]{Corresponding author at: School of Information Science and Engineering, Central South University, Changsha 410083, China}
\ead{ljf@csu.edu.cn}
\address{School of Information Science and Engineering, Central South University, Changsha 410083, China}

\begin{abstract}
In this paper we propose a very efficient method to fuse the unregistered multi-focus microscopical images based on the speed-up robust features (SURF). Our method follows the pipeline of first registration and then fusion. However, instead of treating the registration and fusion as two completely independent stage, we propose to reuse the determinant of the approximate Hessian generated in SURF detection stage as the corresponding salient response for the final image fusion, thus it enables nearly \emph{cost-free} saliency map generation. In addition, due to the adoption of SURF scale space representation, our method can generate scale-invariant saliency map which is desired for scale-invariant image fusion. We present an extensive evaluation on the dataset consisting of several groups of unregistered multi-focus 4K ultra HD microscopic images with size of $4112 \times 3008$. Compared with the state-of-the-art multi-focus image fusion methods, our method is much faster and achieve better results in the visual performance. Our method provides a flexible and efficient way to integrate complementary and redundant information from multiple multi-focus ultra HD unregistered images into a fused image that contains better description than any of the individual input images. Code is available at  \url{ https://github.com/yiqingmy/JointRF}.
\end{abstract}


\end{frontmatter}

\section{Introduction}
\begin{figure}[t]
    \centering
    \begin{subfigure}[b]{0.24\textwidth}
        \includegraphics[width=\textwidth]{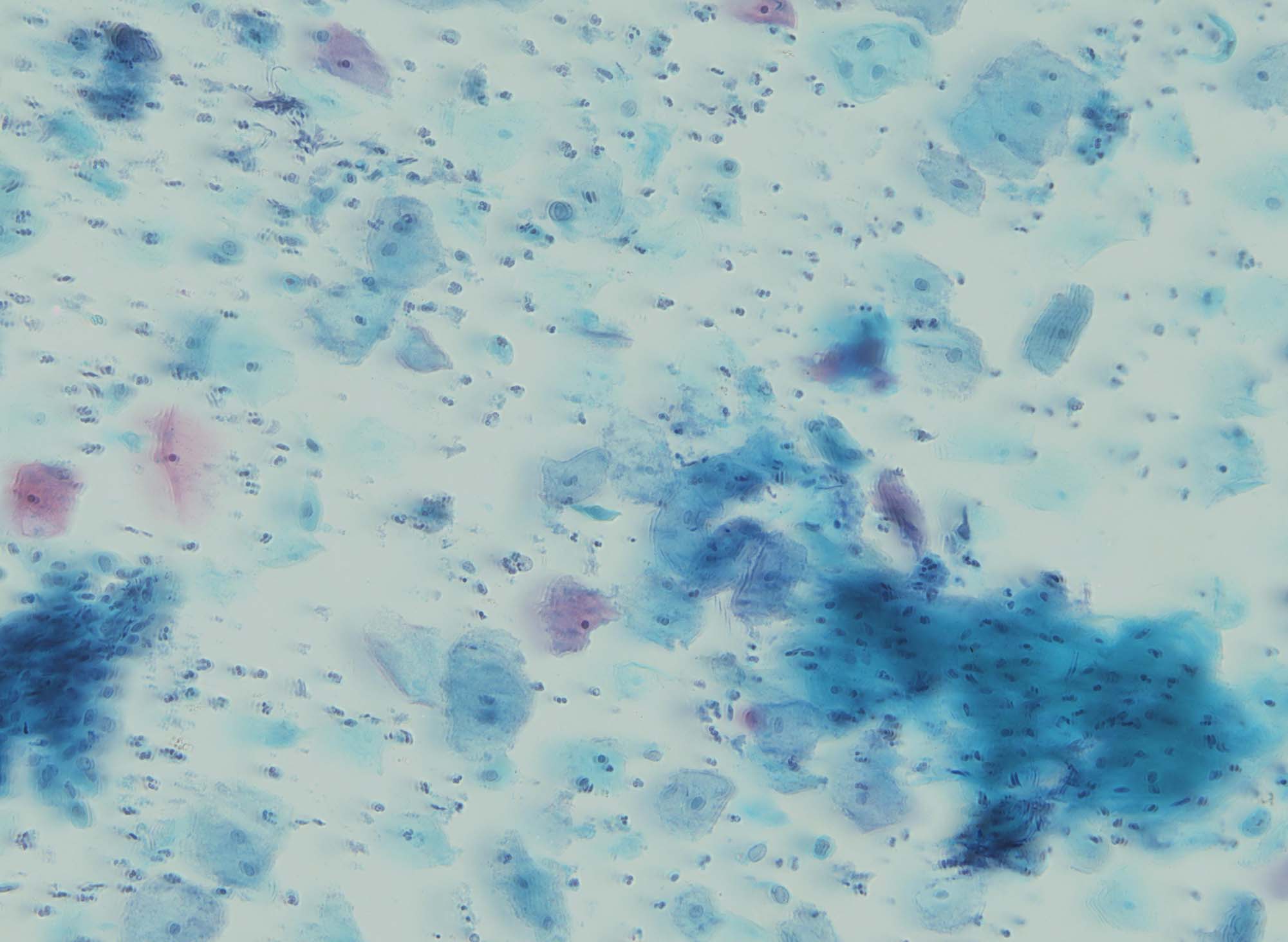}
        \caption{GFF \cite{li2013imageGFF}}    \label{007gff}
    \end{subfigure}
    \begin{subfigure}[b]{0.24\textwidth}
        \includegraphics[width=\textwidth]{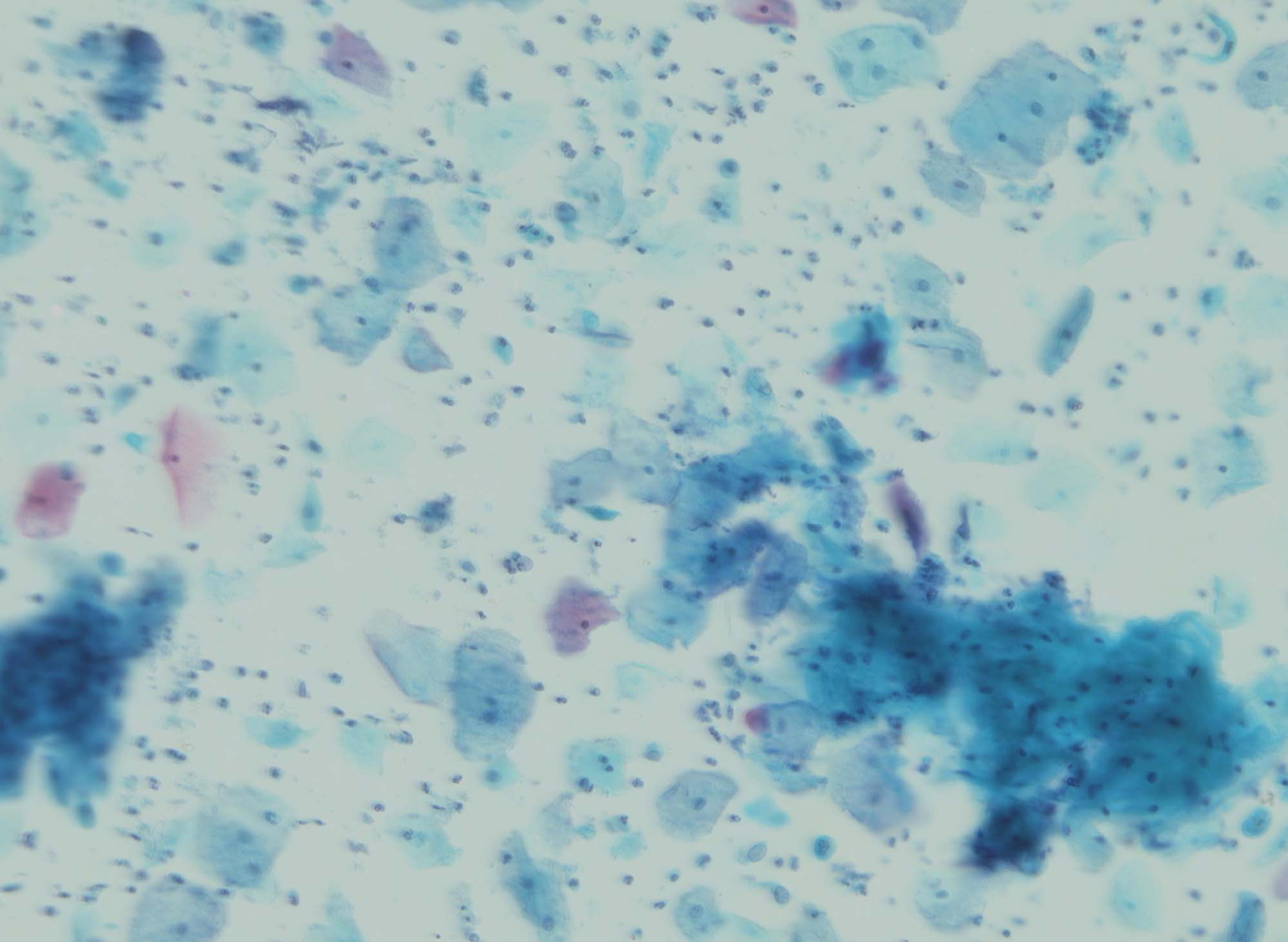}
        \caption{DSIFT \cite{liu2015multi}} \label{007dsift}
    \end{subfigure}
    \begin{subfigure}[b]{0.24\textwidth}
        \includegraphics[width=\textwidth]{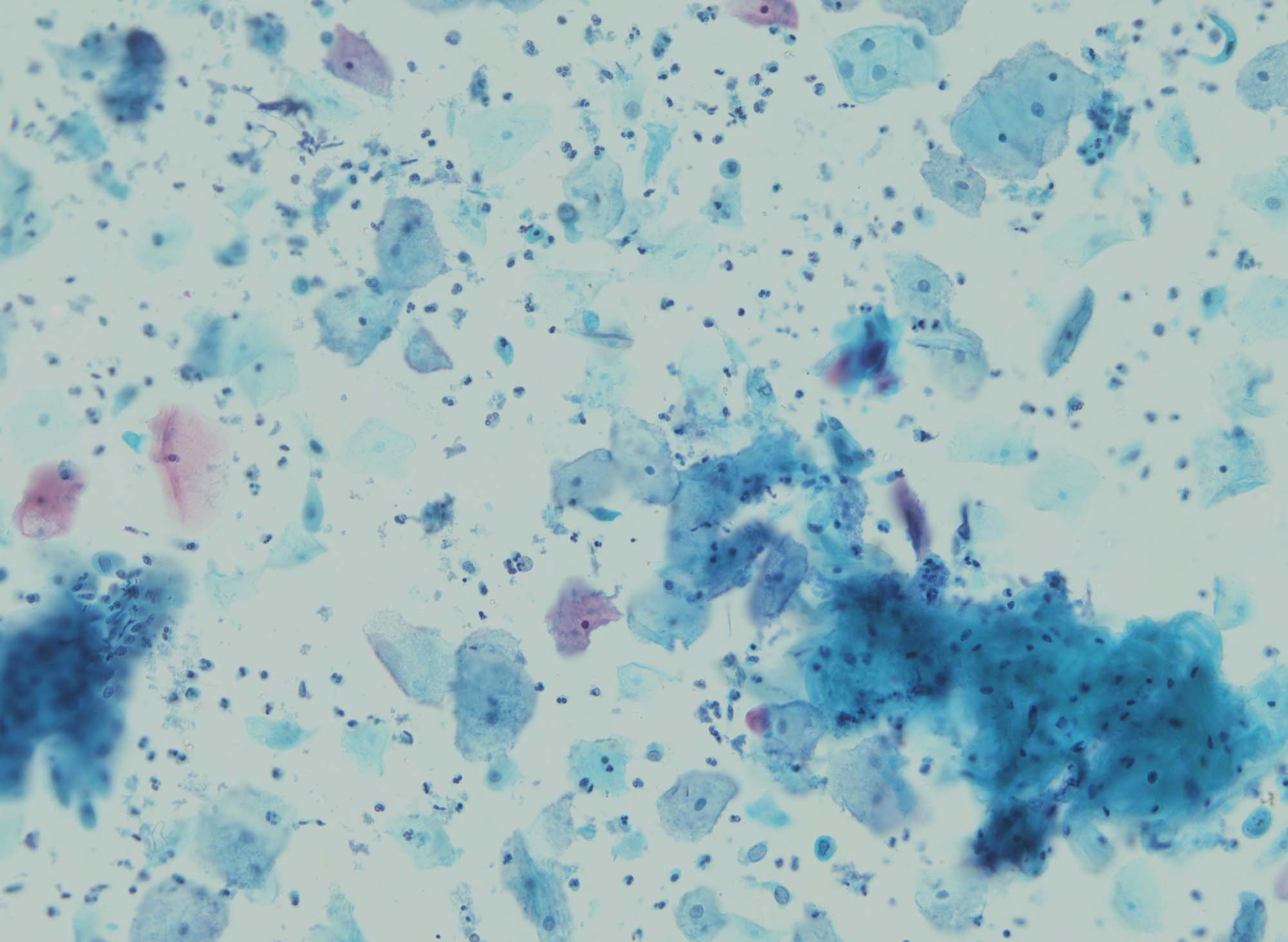}
        \caption{MWGF \cite{zhou2014multi}}  \label{007mwgf}
    \end{subfigure}
    \begin{subfigure}[b]{0.24\textwidth}
        \includegraphics[width=\textwidth]{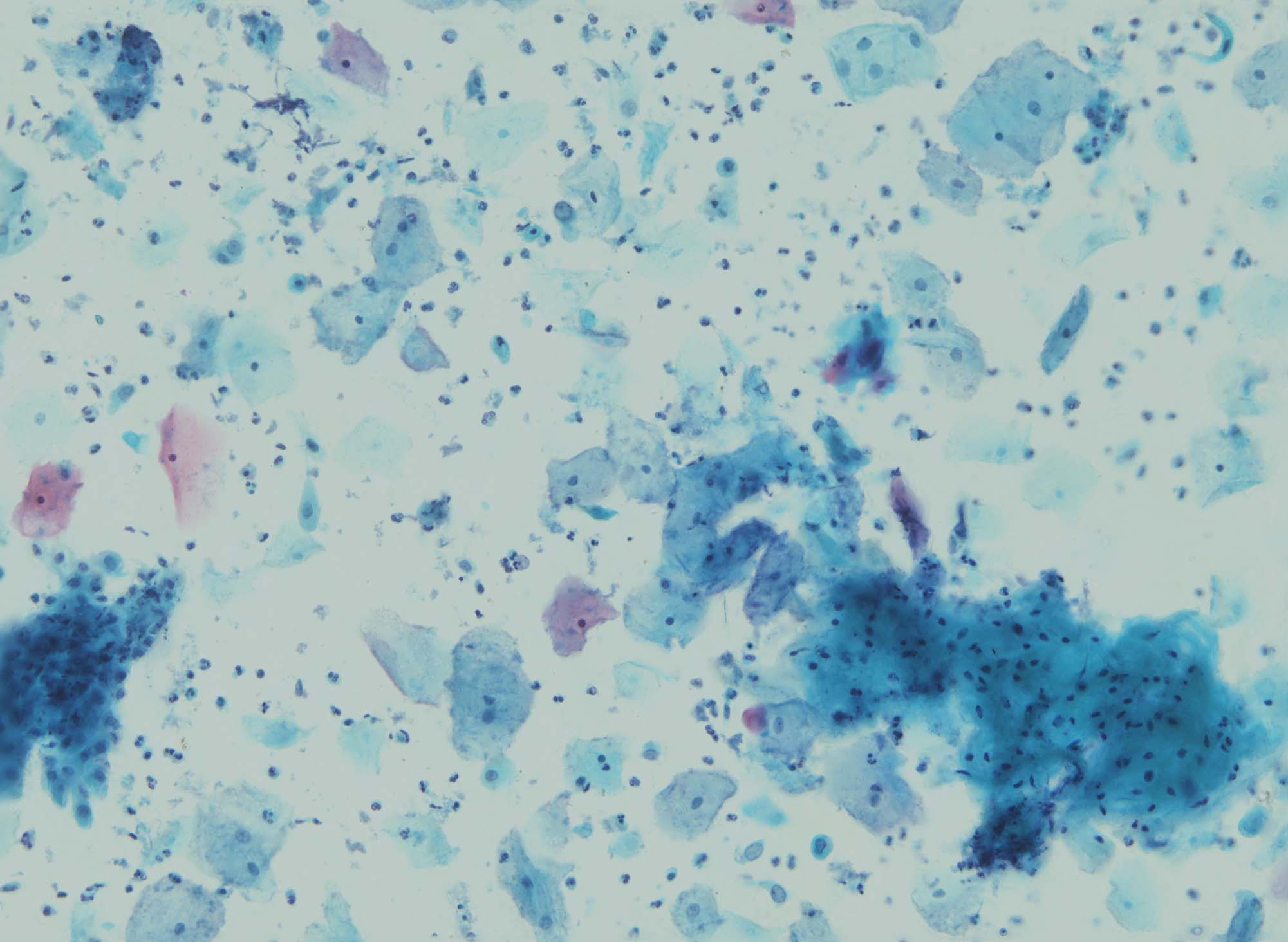}
        \caption{The proposed}   \label{Fig1(d)}
    \end{subfigure}
    \caption{Comparative fused images.}\label{Fig1}
\end{figure}

As the depth-of-field (DoF) of the bright-field microscopy is often limited, we can often capture images focused on the partial scene while the parts of the specimen that lie outside the object plane are blurred. These images are not appropriate for further diagnose analysis. Multi-focus image fusion provides one versatile solution to reconstruct an all-in-focus image by merging a series of local-focused images taken under different distance from the object to the lens. Existing multi-focus image fusion methods are usually based on the assumption that the source images are well-aligned \cite{miao2011novel, li2017pixel,li2013imageGFF} or with only small mis-registration \cite{liu2015multi,zhou2014multi}. But the conditions are not always satisfied for the multi-focus microscopical images which are captured by moving the specimen using a step motor, where slight vibration will result in large mis-registration. Figure \ref{Fig1} shows the fused results by several state-of-the-art multi-focus image fusion methods, including guided filtering (GFF) \cite{li2013imageGFF}, the dense SIFT (DSIFT) \cite{liu2015multi}, and multi-scale weighted gradient (MWGF) \cite{zhou2014multi}, where there are obvious visual artifacts and distortion.

The mis-registration of images has become one of the serious bottleneck to improve the visual quality for multi-focus image fusion. A straightforward solution is to register all input images first and then perform image fusion based on the registered images \cite{liu2015automatic,laliberte2003registration,kessler2006image}. Especially, the images can be registered first using the SIFT-based image registration methods \cite{Gong2014A,Guo2017Automatic}
, then the registered images can be fused using DSIFT \cite{liu2015multi}. However, this scheme regards the registration and fusion as two completely independent parts, resulting in redundant operation during the process of the registration and fusion. In fact, during the SIFT feature detection for registration, it records the difference of gaussian (DoG) response for each pixel, which can also be regarded as salient response for the subsequent image fusion.

In this paper, we propose a novel method based on Speed-Up Robust Features (SURF) \cite{bay2006surf} to fuse the unregistered multi-focus microscopical images in a very efficient way. The key idea is to use the determinant of the approximate Hessian for both SURF detection and the saliency map generation. We first register all input images using the SURF-based registration method and then merge them by regarding the determinant of the approximated Hessian as the corresponding salient response, thus it enables nearly cost-free saliency map generation. In addition, due to the adoption of SURF scale space representation, our method can generate scale-invariant saliency map which is desired for scale-invariant image fusion. Experimental results demonstrate our method is very fast to fuse the high-resolution microscopical images. Furthermore, as shown in Figure \ref{Fig1(d)}, the fused image is of desirable visual quality that there is no geometric distortion and information artifacts.

\subsection{Related Work}
\textbf{Registration Method}. The goal of it is to establish the correspondence between sensed and reference images to maximize the similarity of the two images with the common content \cite{zitova2003image}. It can be roughly classified into intensity-based and feature-based methods \cite{goshtasby20052}. The intensity-based methods compare intensity patterns in images via correlation metrics. The feature-based methods find correspondence between image feature, such as points, T-junction, edge and contour to formulate the feature points, which are widely used due to their simplicity \cite{laliberte2003registration, hill2001medical}.

\textbf{Fusion Method}. It can be mainly divided into transformation domain method and spatial domain method \cite{goshtasby2007image}. The transformation methods are based on the framework "decomposition-fusion-reconstruction", which first transform the images by some mathematical transformation such as pyramid decomposition, wavelet decomposition or sparse coding \cite{pajares2004wavelet, liu2015general}, and then reconstruct the fused image based on the transform coefficients. They are often not efficient due to the multi-decomposition and inverse construction \cite{aslantas2014pixel}. The spatial domain methods try to take each pixel in the fused image as the weighted average of the corresponding pixels in the input images, where the weights or activity map are often determined according to the saliency of different pixels. These methods are often fast and easy to be implemented, but the fused performance extremely depends on the accurate estimation of the weights of the pixels. The proposed method belongs to this category. Recently, there are some methods \cite{Liu2018Deep} based on the convolution neural networks (CNNs) to regard the image fusion as a classification problem to avoid the handcraft design. However, it is not efficient and the performance highly depends on the network and the size of training dataset.

\textbf{Joint Fusion and Registration}. Recently, there are several joint methods \cite{chen2015sirf, chen2011maximum, ofir2017registration} which combine the registration and fusion into a single frame and then use the alternating optimization between registration and fusion to minimize the resulting energy function. However, the repeated iteration is often time-consuming, which makes these joint methods not appropriate to fuse the high resolution images in real-time or nearly real-time systems.

\begin{figure}[t]
	\centering
	\includegraphics[width=0.5\linewidth]{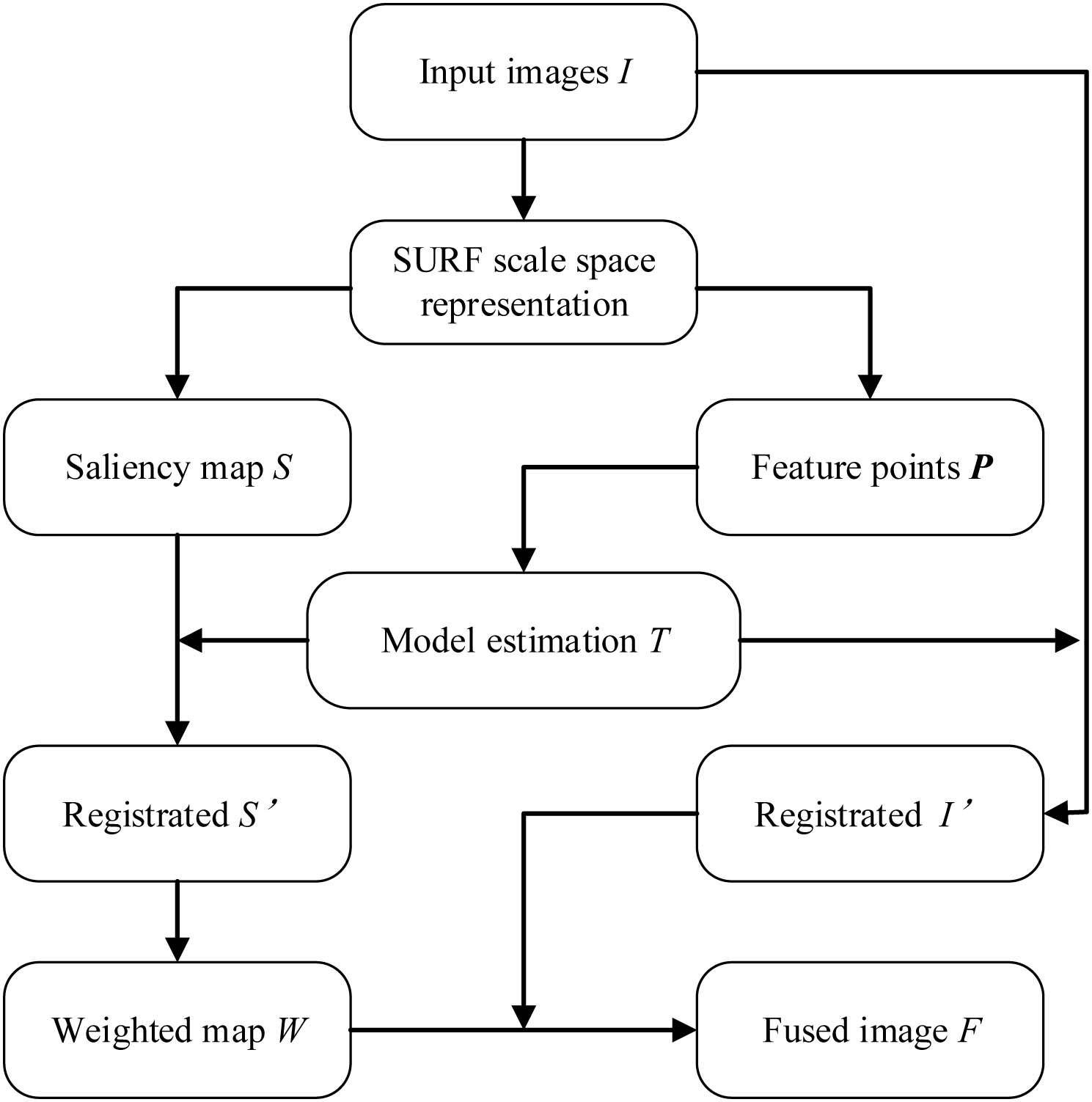}
	\caption{The scheme of the proposed multi-focus image fusion method.}
	\label{Fig2}
\end{figure}

\begin{figure}[t]
  \centering
  \includegraphics[width=0.75\textwidth]{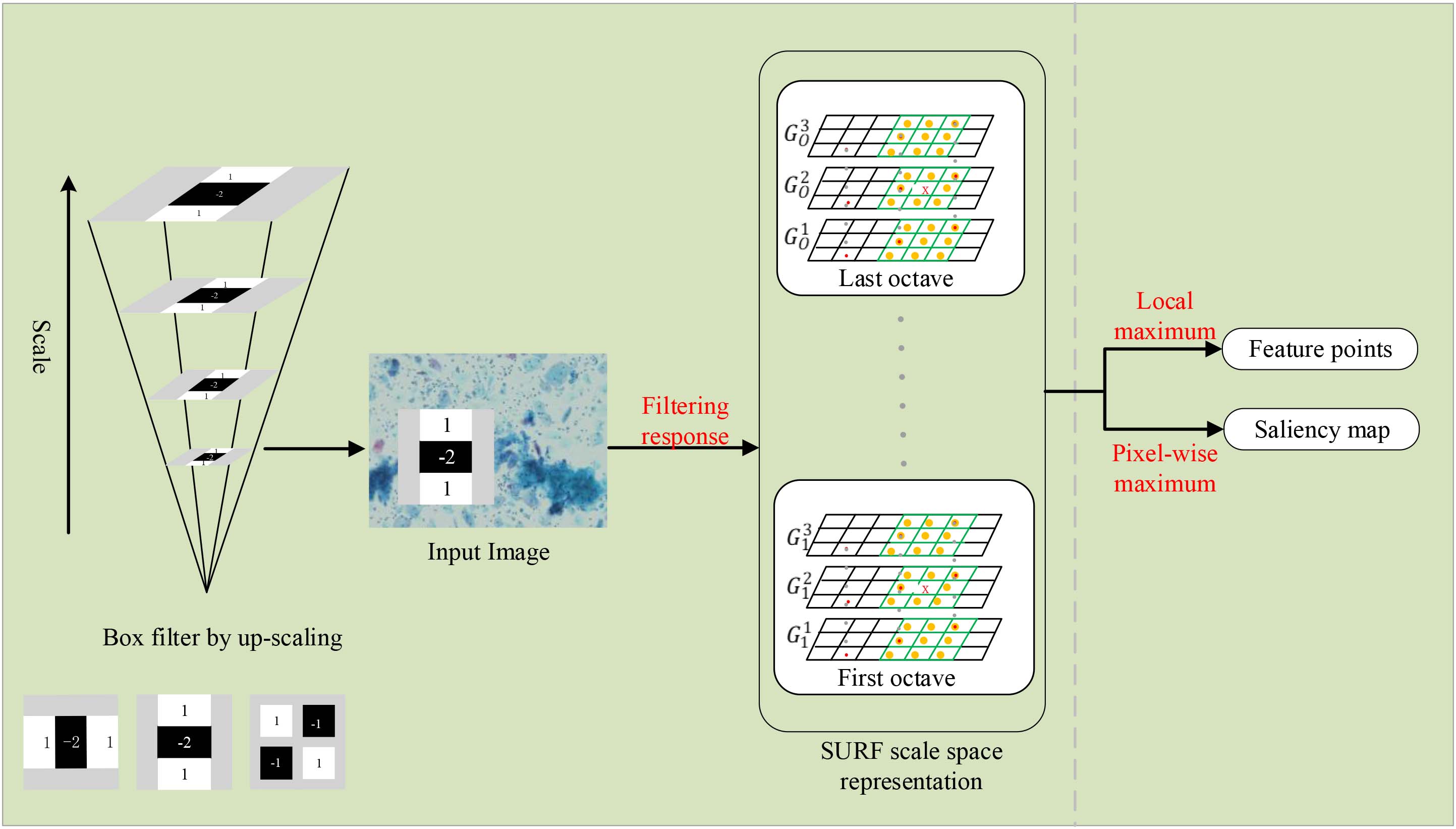}
  \caption{The schematic diagram of SURF scale space representation.}
  \label{scale-space}
\end{figure}

\section{Methodology }
The scheme of the proposed method is shown in Figure \ref{Fig2}. First, following SURF \cite{bay2006surf} we first approximate the Gaussian second order derivatives with box filters to calculate the Hessian matrix and construct a scale space representation for each image by simply varying the box filters size. Second, based on the SURF scale space representation, we can detect the repeatable feature points and generate the scale-invariant saliency map simultaneously. Third, the simplified registration transformation model can be estimated using SURF matching followed by a Hough-like voting scheme. Based on the estimated registration model, all input images and the corresponding scale-invariant saliency map can be registered individually. Then a simple non-max suppression operation is applied to these registered saliency maps and the resulting mask images are further refined by a single-scale guided filtering \cite{He2013Guided} to generate the optimal weighted maps. Finally, the fused image is constructed by combining the registered source images with the corresponding weighted maps.

\subsection{SURF Scale Space Representation }\label{SURF}
The SURF relies on the determinant of the Hessian for both location and scale selection \cite{bay2006surf}. However, during calculating the Hessian the SURF simply replaces the second order Gaussian partial derivative by box filters shown in the bottom-left of Figure \ref{scale-space}, whereas these approximations can be evaluated very fast using integral images  independently of size. Based on these approximations, we can efficiently construct the SURF scale space representation by up-scaling the filter size rather than iteratively reducing the image size.
%
To be specific, the scale space includes several octaves and each octave is filled with constant layers and for the $o$-th octave and $k$-th layer, the corresponding salient response image $G_{o}^{k}(x,y)$ is then determined by the determinant of approximate Hessian matrix $\mathcal{H}_{\mathrm{approx}}$, i.e.,
\begin{equation}
G_{o}^{k}(x,y)=\mathrm{det}\left( \mathcal{H}_{\mathrm{approx}}(x,y,w(o,k)) \right),
\end{equation}
\begin{equation}\label{H}
   \mathcal{H}_{\mathrm{approx}}(x,y,w)=\left [\begin{array}{ccc} D_{xx}(x,y,w) & \quad \alpha D_{xy}(x,y,w) \\
\alpha D_{xy}(x,y,w) & \quad D_{yy}(x,y,w)\end{array} \right ],
\end{equation}
where the $D_{xx}(x,y,w)$ is the convolution of the box filter in $x$-direction of size $w$ with the image in position $(x,y)$, similar to $D_{yy}(x,y,w)$ and $D_{xy}(x,y,w)$. The $\alpha$ is the relative weight to balance the expression for the actual determinant of Hessian matrix, which is often set as $\alpha=0.9$, and $w(o,k)$ is the size of box filters corresponding to the $o$-th octave and the $k$-th layer which is often determined by
\begin{equation}\label{s}
w(o,k) =(2^{o} \times k+1)\times 3.
\end{equation}
The smallest size of box filters is set to $w(1,1) = 9$ which are the approximations for Gaussian second order derivatives with $\sigma = 1.2$ and represents the lowest scale or the highest spatial resolution.

\subsection{Feature-Based Registration}
\subsubsection{Registration Model}
The feature-based registration \cite{hill2001medical} aims to set up a parametric transformation $T$ that can relate the position $(x_s,y_s)$ of features in the sensed image or coordinate space with the position $(x_r,y_r)$ of the corresponding feature in the reference image or coordinate space. The general 2D
planar transformation is modeled as perspective transform or \emph{homography}, which operates on homogeneous coordinates
\begin{equation}
\begin{bmatrix}
x_r\\
y_r\\
1
\end{bmatrix}=T\begin{bmatrix}
x_s\\
y_s\\
1
\end{bmatrix},
\end{equation}
where $T$ is an arbitrary $3\times3$ matrix. Note that $T$ is homogeneous, it has only eight degrees of freedom. As for our microscopical scenarios where the multi-focus images are taken by moving the specimen using a step motor, the misregistration is mainly generated from the vibration of the step motor and the scale change caused by varying the distance from the specimen to the lens. Due to our scale-invariant saliency map generation (Section \ref{sismg}), we neglect the scale change and further approximate the motor's vibration as translation in vertical and horizonal directions for reasons of simplicity. Thus, the finally transformation $T$ can be simplified as
\begin{equation}\label{T}
  T=\left [\begin{array}{ccc} 1 & 0 & t_x \\
0 & 1 & t_y \\
0 & 0 &1
\end{array} \right ],
\end{equation}
where the $t_x$ and $t_y$ are the only two freedoms that need to be estimated.

To determine the model $T$ and register the series of input images, we follow the basic pipeline of feature-based registration \cite{zitova2003image}, i.e., feature detection, feature matching, and image registration.

\subsubsection{Feature Detection and Description}
Based on the SURF scale space representation, we can detect the feature points by applying \emph{non-maximum suppression} in the $3\times3\times3$ local region, as shown in Figure \ref{scale-space}. To cover a complete octave, there are two additional layers added to each octave. After searching through all the layers, the scale-invariant feature points can be detected to prepare for the feature matching. Then, the description for each feature point is calculated based on Haar wavelet response (refer to \cite{bay2006surf} for details). Because there is no rotation transformation in the multi-focus images, we adopt the upright SURF (U-SURF) to accelerate the calculation. Usually, the dimension of the descriptor is 64 ($4\times4\times4$) and can be easily changed to 36 ($3\times3\times4$) or 100 ($5\times5\times4$) by splitting up each region into square sub-regions with different size. 
\subsubsection{Reference Image selection and Feature Matching}
We first need to choose an reference image from the series of input multi-focus images. A reasonable way is to select the image with least blurred region or highest sharpness. A more sophisticated method \cite{liu2015automatic} is to choose the image whose information entropy is closest to the root mean square values of all images. However, these methods will introduce additional computation to evaluate the sharpness or distances in terms of information entropy. Here we choose one with the largest number of detected feature points as the reference image and the others as the sensed images. The rationale behind this is the reference image should be selected to match the others as possible. Such a cost-free alternative is a very simple but also effective in practice.
%

In the matching stage, for each feature point in the sensed image, we search the most similar point in the reference image by \emph{Euclidean-Distance} with the corresponding descriptors. In our implementation, we reject all matches in which the ratio of closest to second-closest neighbors is greater than 0.8. As pointed out in \cite{bay2006surf}, the matching can be accelerated by taking advantage of the sign of the Laplacian (i.e. the trace of the Hessian matrix) which distinguishes bright blobs on dark backgrounds from the reverse situation. If the trace of the $\mathcal{H}_{\mathrm{approx}}$ for two feature points is opposite, there is no need to calculate the distance between these points and therefore can accelerate the feature matching.

\subsubsection{Model Estimation and Image Registration}
\begin{figure}[t]
    \centering
    \begin{subfigure}[b]{0.4\textwidth}
        \includegraphics[width=\textwidth]{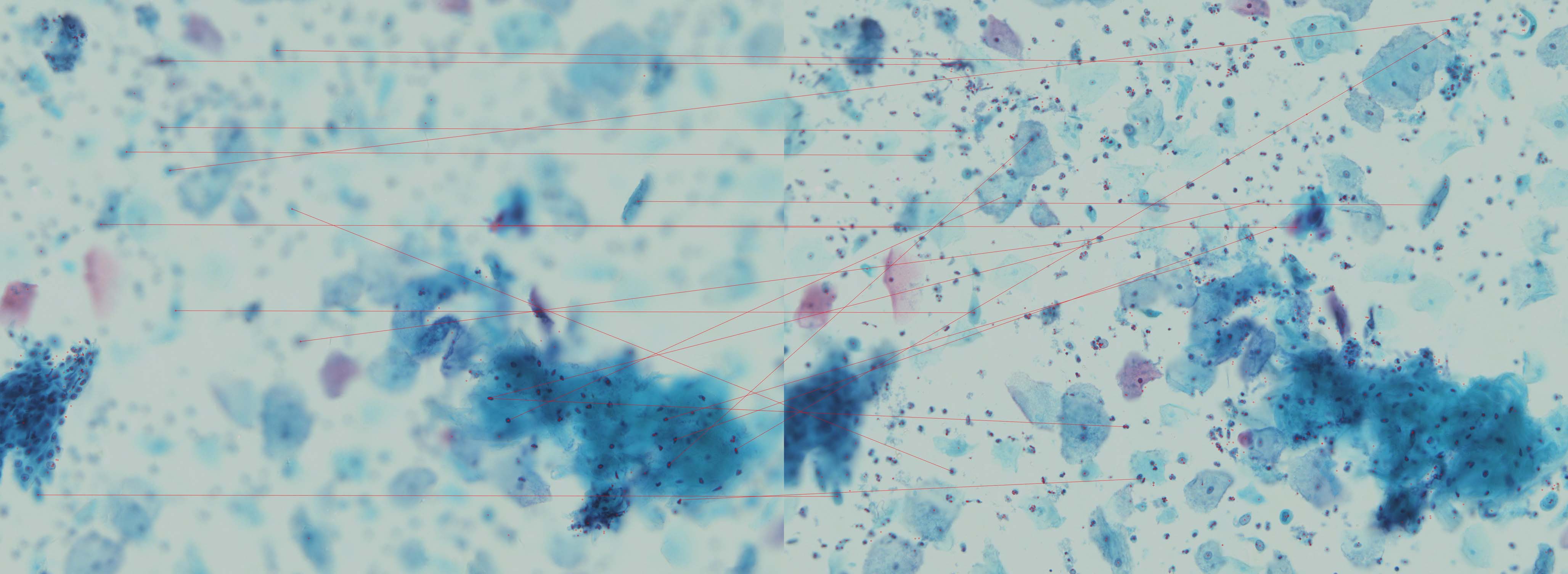}
        \caption{} \label{Fig5(a)}
    \end{subfigure}
    \begin{subfigure}[b]{0.4\textwidth}
        \includegraphics[width=\textwidth]{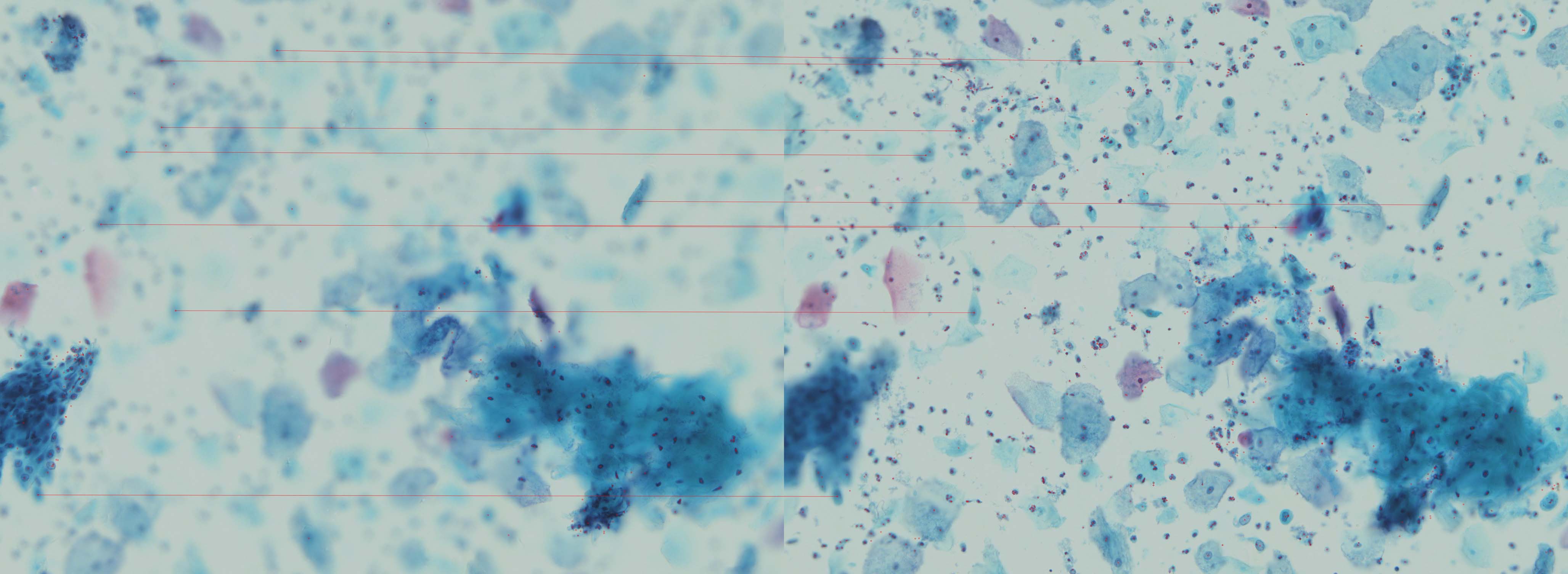}
        \caption{} \label{Fig5(b)}
    \end{subfigure}
    \caption{ (a) The top 20 matched pairs of feature points. (b) The final matched pairs.}\label{Fig5}
\end{figure}

Based on the matched pairs, we can estimate the transformation parameters and then perform the registration between each sensed image and reference image. As shown in Figure \ref{Fig5(a)}, there are some obvious mismatched pairs due to the similar local appearance and the simplistic matching strategy. The RANdom SAmple Consensus (RANSAC) algorithm \cite{Fischler1981Random} is widely used to estimate the transformation parameters which are robust to these ``outliers''. However, since RANSAC needs to iteratively fit the model and identify the ``outliers'', it is a little time-consuming. As our model only involves two parameters (i.e. $t_x$, $t_y$), here we adopt an alternative Hough-like voting scheme which is much faster. More precisely, we divide the parameter space into several grids constrained by an error tolerance, and calculate the $(t_x, t_y)$ for each matched pair which votes for the corresponding grid. The matched pairs who vote for the grid with the highest value are regarded as ``inliers'', as shown in Figure \ref{Fig5(b)}. Then the final optimal $(t_x, t_y)$ is simply determined by the median or mean of the ones of these ``inliers'', respectively corresponding to solution of least absolute deviations ($\ell_1$) or least square errors ($\ell_2$) loss. Based on the estimated translation parameters, the sensed images can be easily aligned to reference image.

\subsection{Spatial Domain Fusion}
The spatial-domain fusion method often consists of the following three steps, i.e., saliency map generation, weighted map construction and image fusion.
\subsubsection{Scale-invariant Saliency Map Generation}\label{sismg}
The saliency map records the salient response for each pixel. A good saliency map should be robust to the noise and the varies of objects size. As mentioned above, we directly use the pixel-wise determinant of the approximated Hessian as the corresponding salient response, therefore we can generate the scale-invariant saliency map almost without extra computation. Specifically, for each pixel in the image $I(x,y)$ , we achieve the scale-invariant saliency by searching across the scale space
\begin{equation}\label{saliency}
  S(x,y)=\mathop{max} \limits_{o: 1\rightarrow O}\left \{\mathop{max} \limits_{k:1\rightarrow L} \left \{G_{o}^{k}(x,y)\right \}\right \},
\end{equation}
where $L$ and $O$ are the number of layers and octaves, respectively. It should be noted that the scale space is constructed based on the unregistered input images and thereby the resulting scale-invariant saliency is also needed aligned based on the above estimated translation parameters, resulting in the registered saliency map $\hat{S}(x,y)$. 
%

\subsubsection{Weighted Map Construction and Fusion}
A straightforward approach is to take the registered scale-invariant saliency of each pixels $\hat{S}(x,y)$ as the corresponding weight. However, it will introduce blur in the fused image. We adopt the simple \emph{non-max suppression} to alleviate this problem. The initial weight map $W^i(x,y)$ for the $i$-th image can be obtained with the aligned saliency map $\hat{S}^{1}(x,y)$
\begin{equation}\label{}
  W^{i}(x,y) = \left\{
                \begin{array}{cl}
                   1, & \mathrm{if}~ {\hat{S}}^{i}(x,y) = \max \left({\hat{S}}^{1}(x,y), {\hat{S}}^{2}(x,y), \cdots, {\hat{S}}^{n}(x,y) \right) \\
                   0, & \mathrm{otherwise} \\
                 \end{array}
                 ,\right.
\end{equation}
where $n$ is the number of input images.

Notice that the above procedure compares pixels individually without considering the spatial context information, which will introduce some geometric distortion and artifacts in the final fused image. Following \cite{li2013imageGFF}, we determine the final weight $\hat{W}(x,y)$ by applying guided filtering \cite{He2013Guided} on the initial weight map $W(x,y)$ as follows
\begin{equation}\label{gif}
  \hat{W}(x,y) =  \frac{1}{|\omega|} \sum_{(u,v) \in \omega(x,y)}[ a(u,v) \hat{I}(x,y) + b(u,v)],
\end{equation}
where $|\omega|$ is the number of pixels in window $\omega(x,y)$ with size of $(2r+1)\times(2r+1)$ which is centered at pixel $(x,y)$, $\hat{I}(x,y)$ is the aligned image, $a(x,y)$ and $b(x,y)$ are the const coefficients of window $\omega(x,y)$ which are determined by \emph{ridge regression}
\begin{eqnarray}
   a(x,y) &=& \frac{\frac{1}{|\omega|} \sum_{(u,v) \in \omega(x,y)} \hat{I}(u,v) W(u,v) - \mu(x,y) \bar{W}(x,y)}{\delta^2(x,y) + \epsilon} \\
   b(x,y) &=&  \bar{W}(x,y) - a(x,y)\mu(x,y).
\end{eqnarray}
Here $\mu(x,y)$ and $\delta^2(x,y)$ are the mean and variance of registered image $\hat{I}(x,y)$ in window $\omega(x,y)$, $\bar{W}(x,y)$ is the mean of the initial weight map $W(x,y)$ in window $\omega(x,y)$ and $\epsilon$ denotes the regularization parameter penalizing large $a(x,y)$.

According to Eq. \ref{gif}, we can determine the corresponding weight map $\hat{W}^i(x,y)$ for each registered image $\hat{I}^i(x,y)$ and thereby obtain the final fused image $F(x,y)$ by
\begin{equation}\label{fused}
  F(x,y)=\frac{\sum_{i=1}^{n} \hat{W}^{i}(x,y) \hat{I}^{i}(x,y)}{\sum_{i=1}^{n} \hat{W}^{i}(x,y)}.
\end{equation}
\begin{figure}[t]
    \centering
    \begin{subfigure}[b]{0.24\textwidth}
        \includegraphics[width=\textwidth]{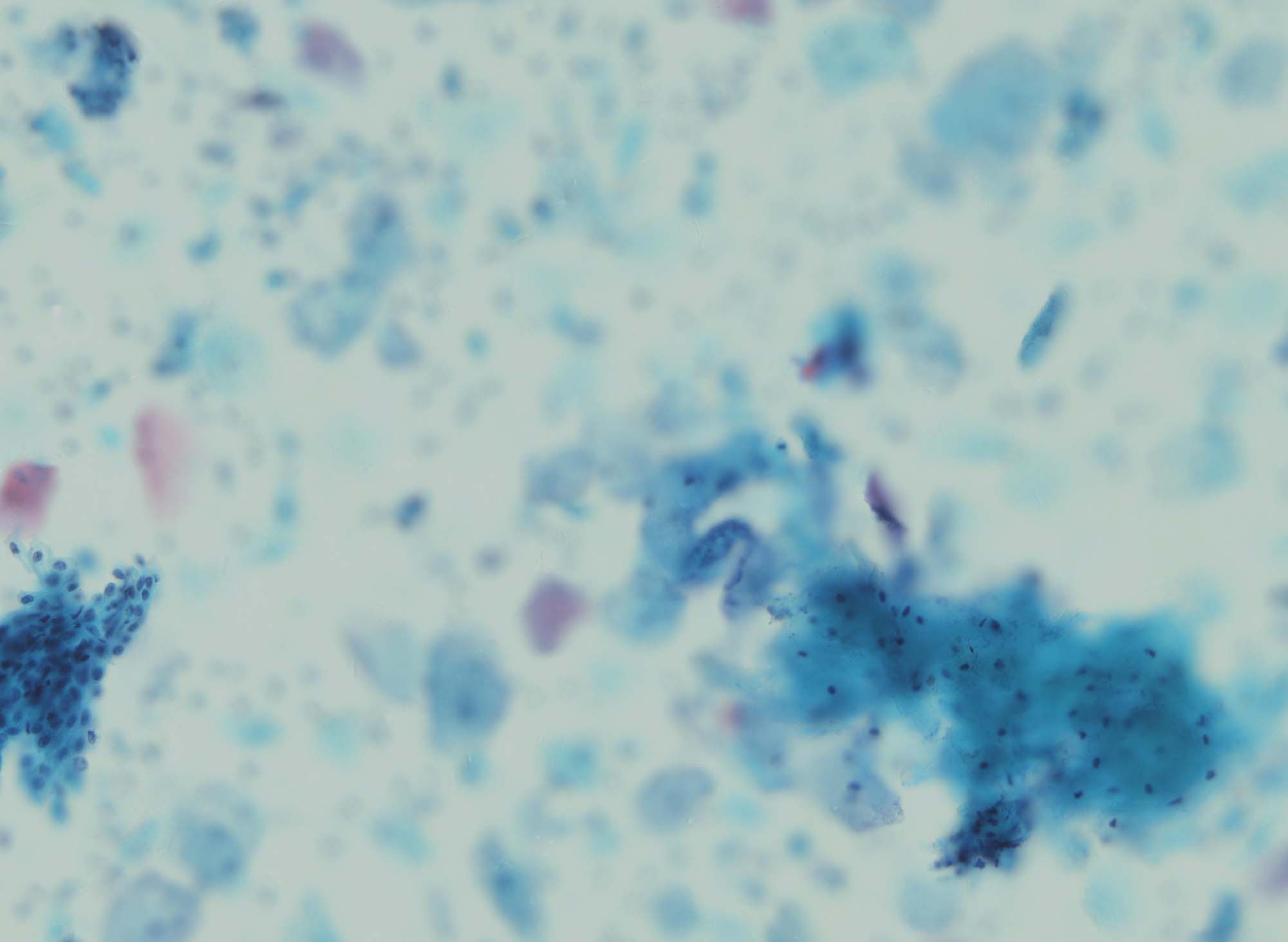}
        \caption{}    \label{s1}
    \end{subfigure}
        \begin{subfigure}[b]{0.24\textwidth}
        \includegraphics[width=\textwidth]{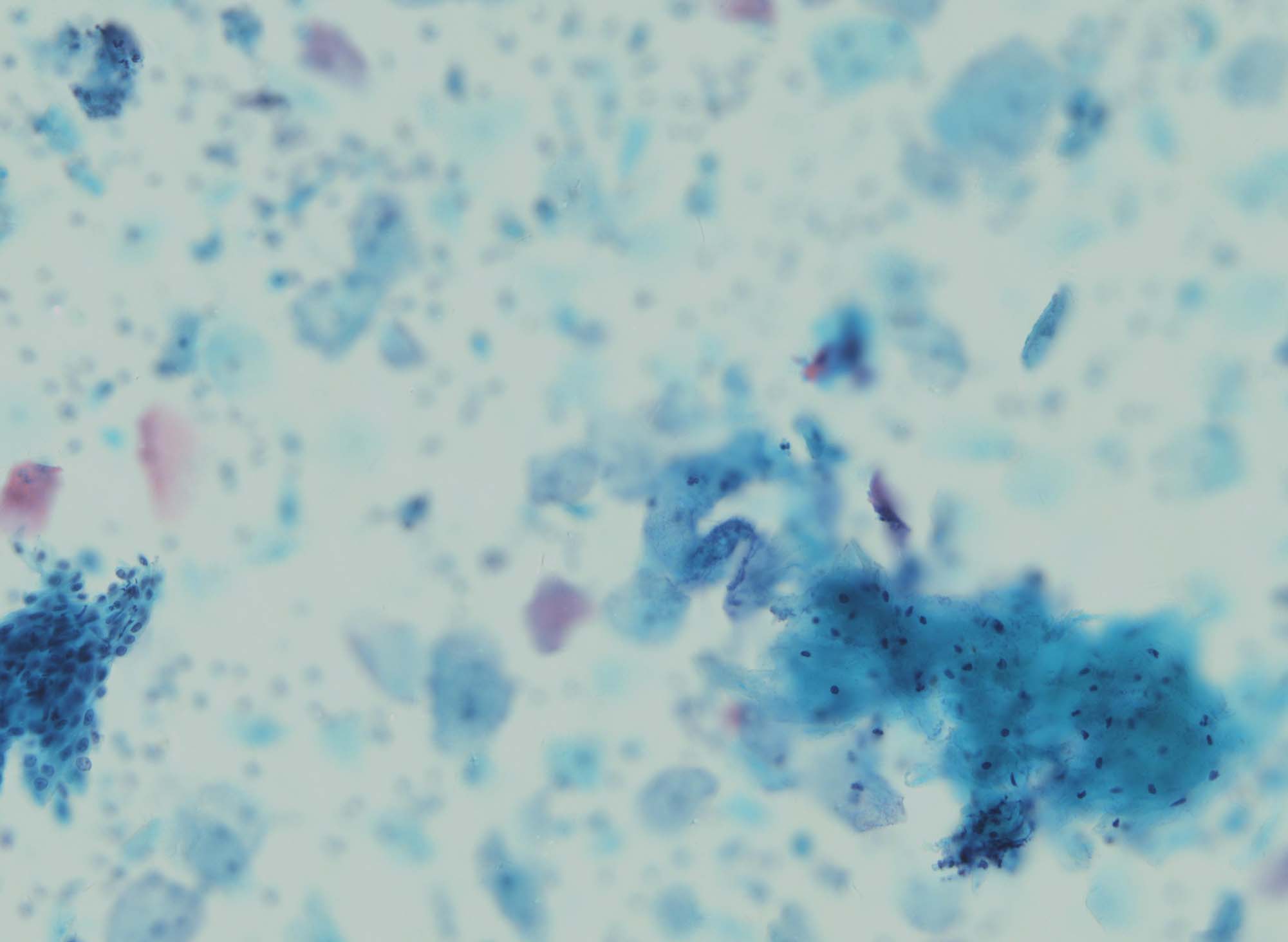}
        \caption{}   \label{s2}
    \end{subfigure}
    \begin{subfigure}[b]{0.24\textwidth}
        \includegraphics[width=\textwidth]{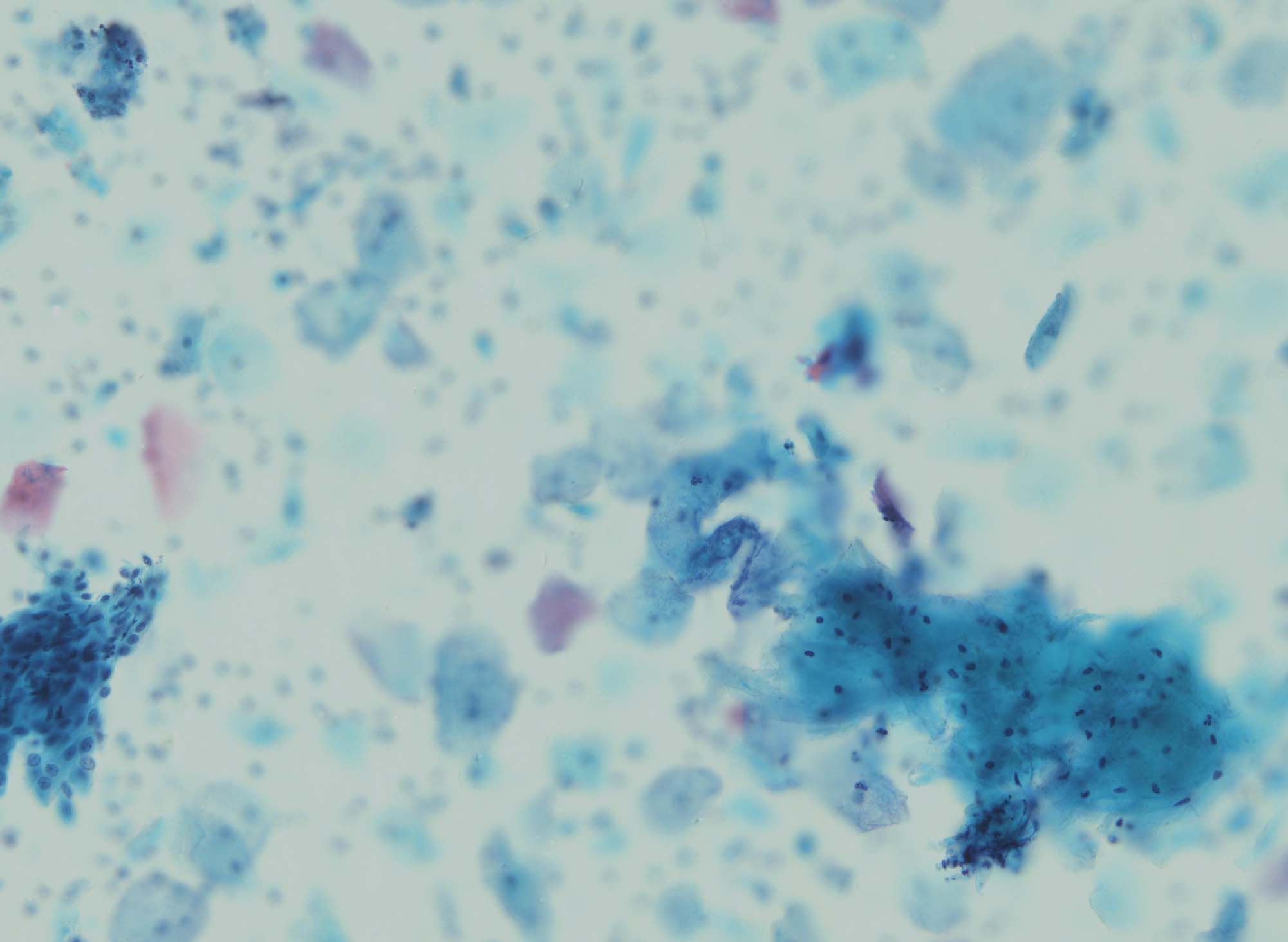}
        \caption{}   \label{s3}
    \end{subfigure}
    \begin{subfigure}[b]{0.24\textwidth}
        \includegraphics[width=\textwidth]{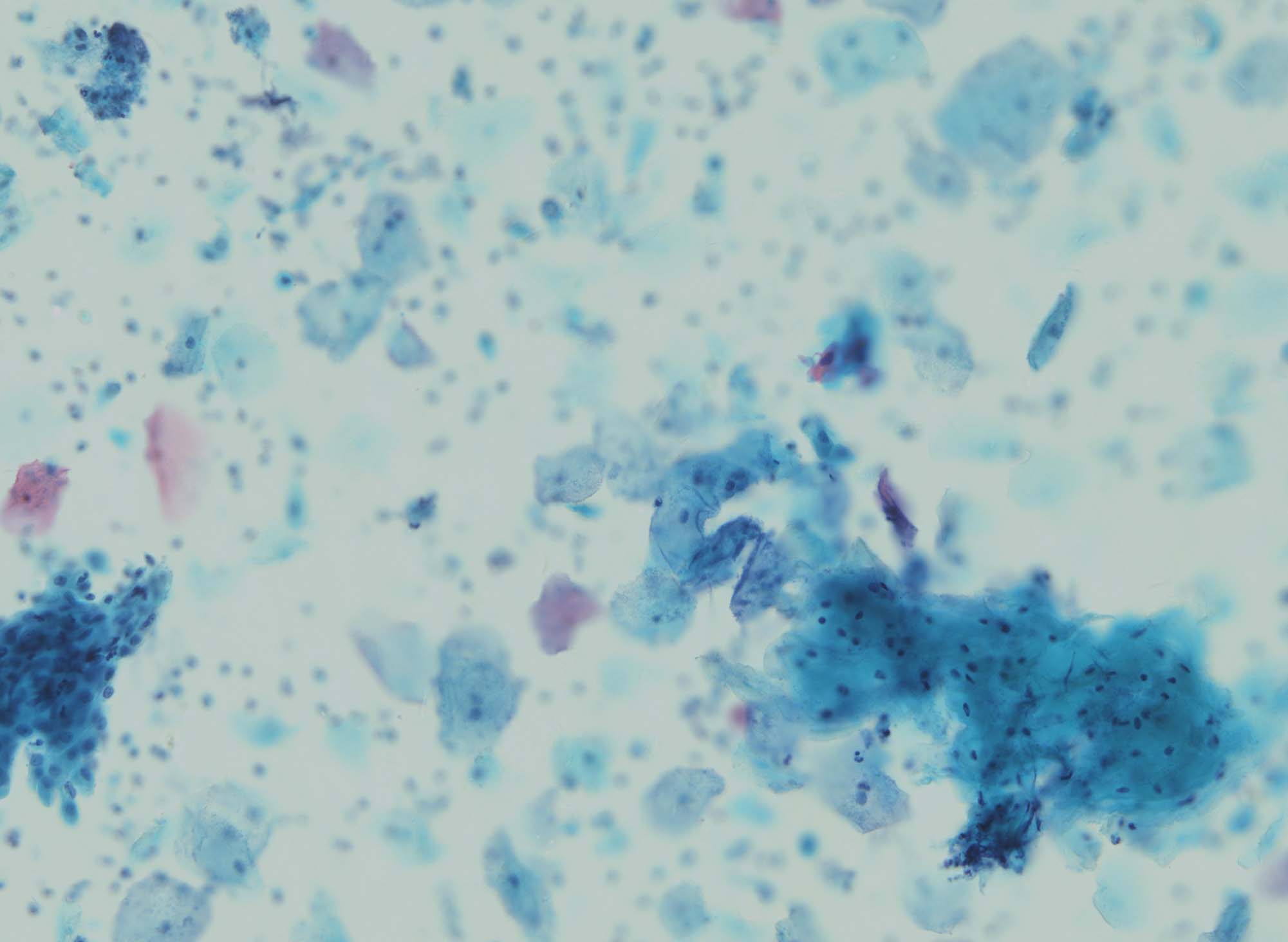}
        \caption{}   \label{s4}
    \end{subfigure}\\
    \begin{subfigure}[b]{0.24\textwidth}
        \includegraphics[width=\textwidth]{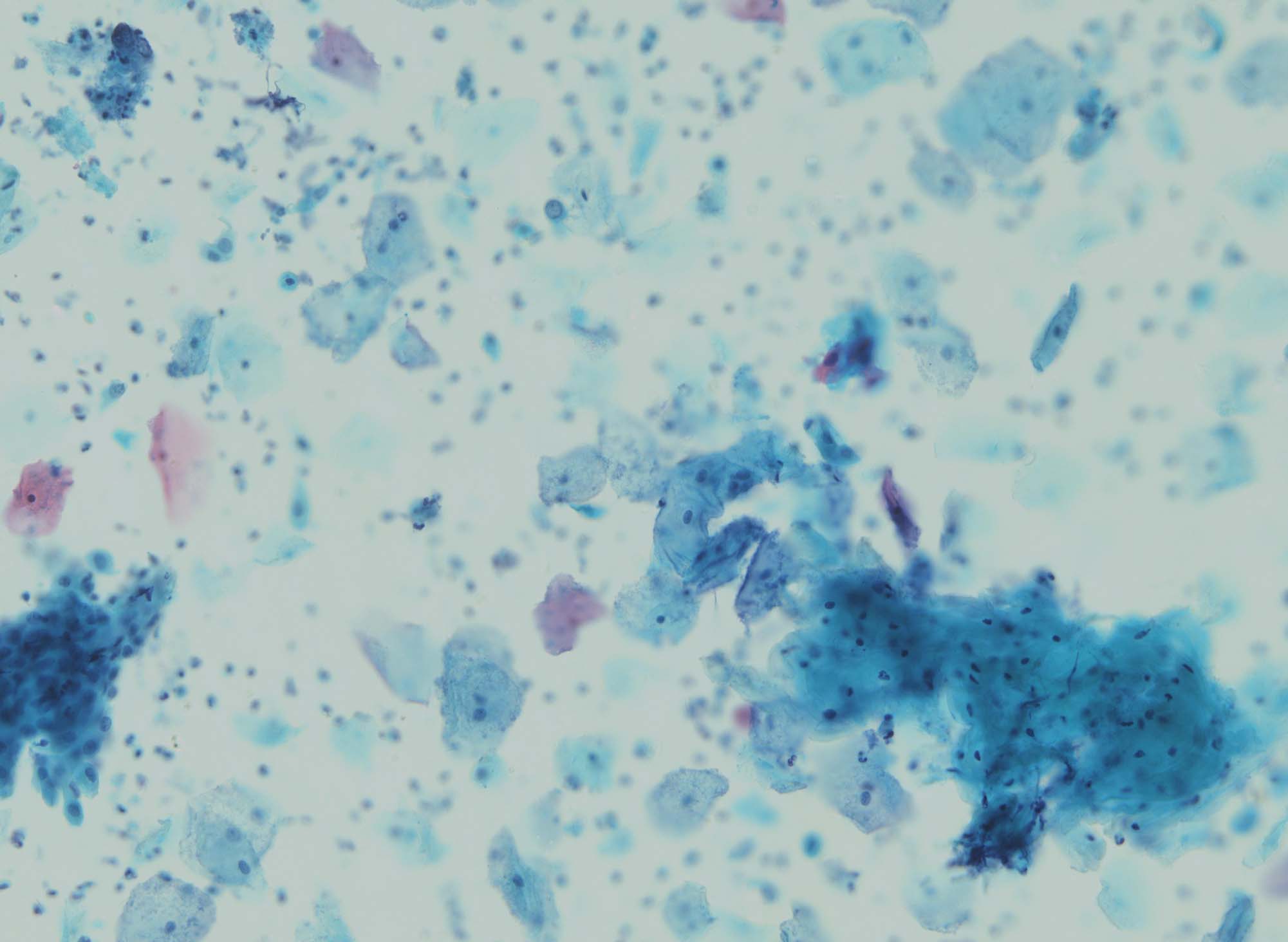}
        \caption{} \label{s5}
    \end{subfigure}
    \begin{subfigure}[b]{0.24\textwidth}
        \includegraphics[width=\textwidth]{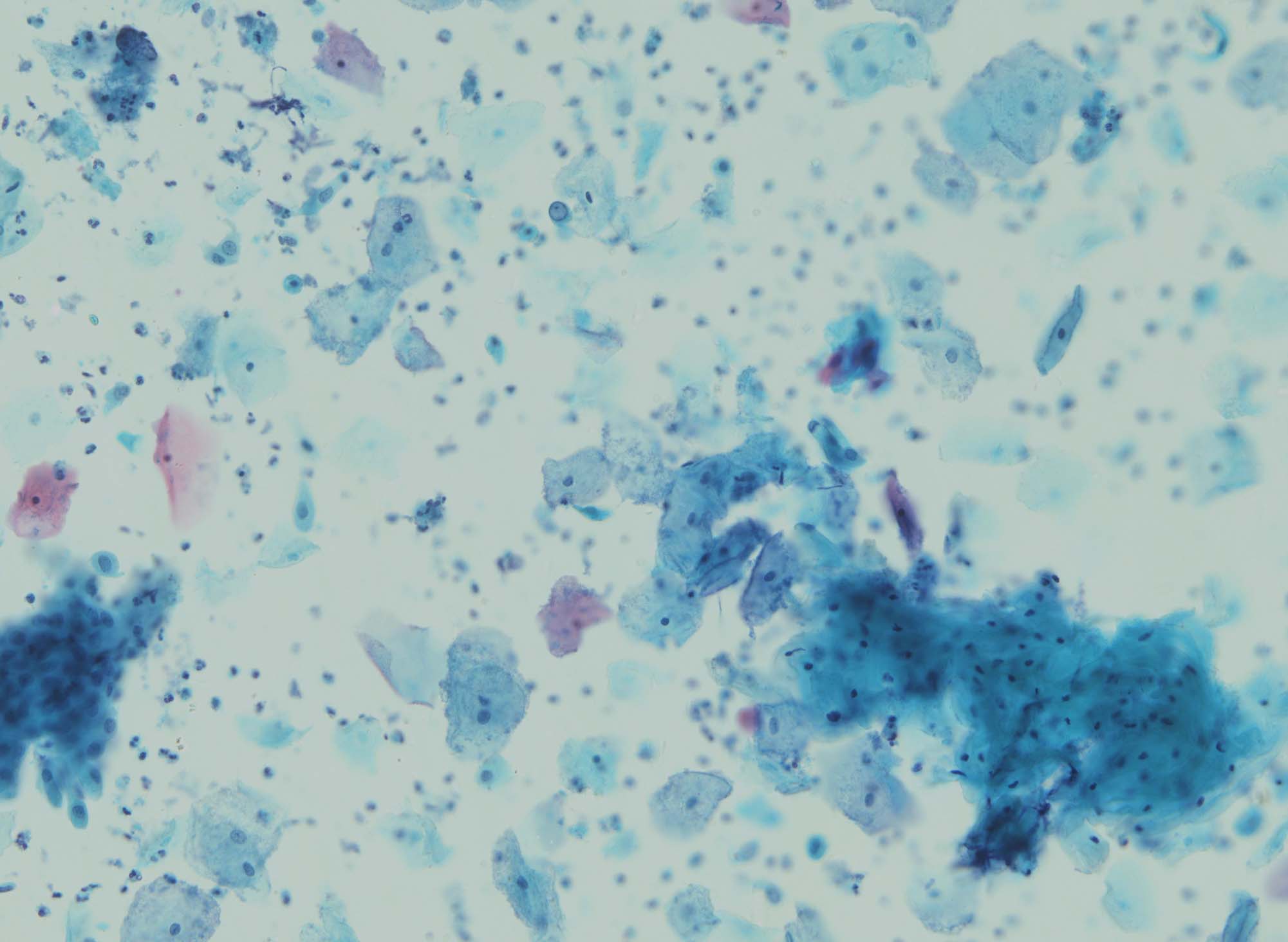}
        \caption{}  \label{s6}
    \end{subfigure}
    \begin{subfigure}[b]{0.24\textwidth}
        \includegraphics[width=\textwidth]{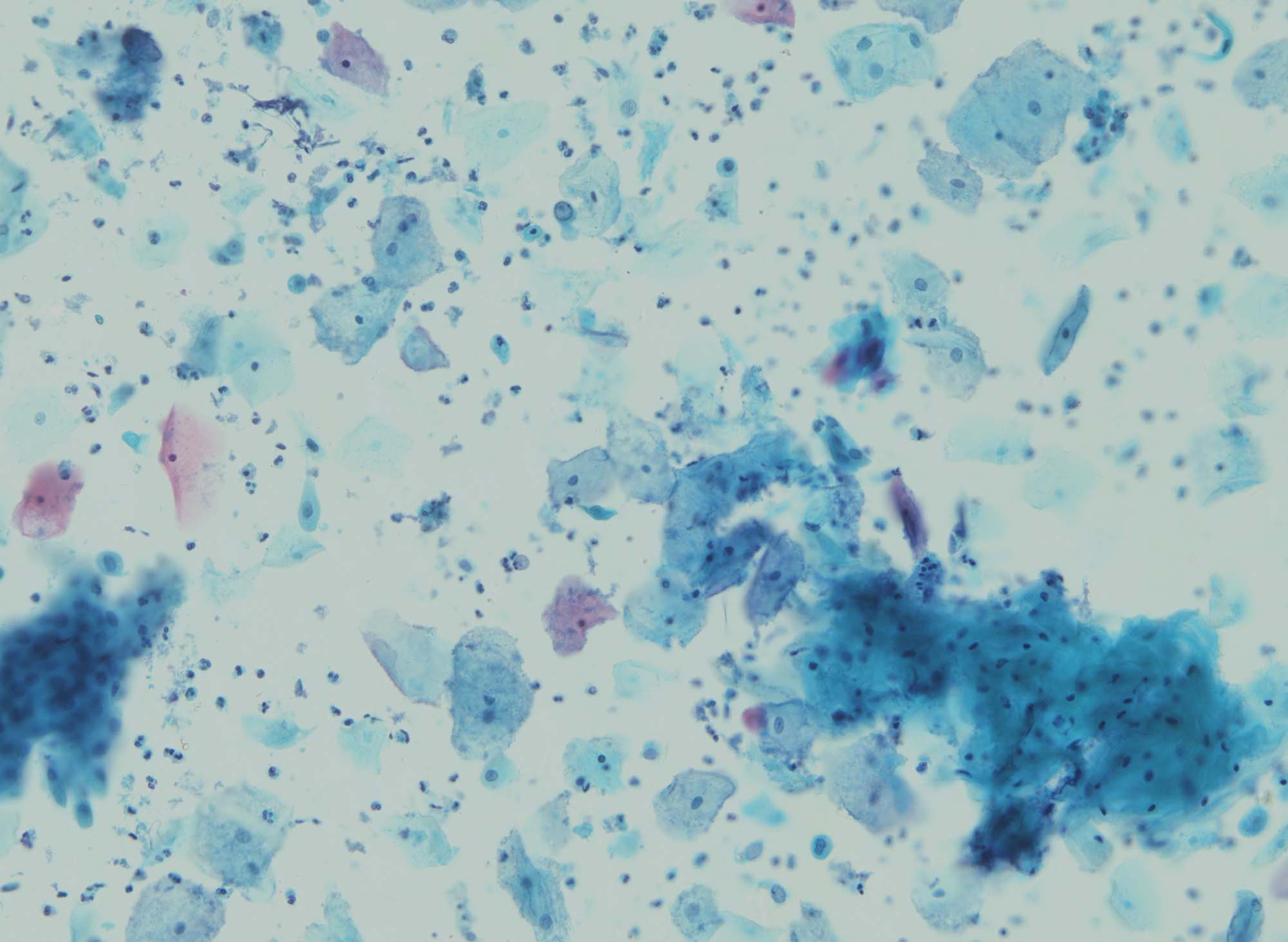}
        \caption{}    \label{s7}
    \end{subfigure}
    \begin{subfigure}[b]{0.24\textwidth}
        \includegraphics[width=\textwidth]{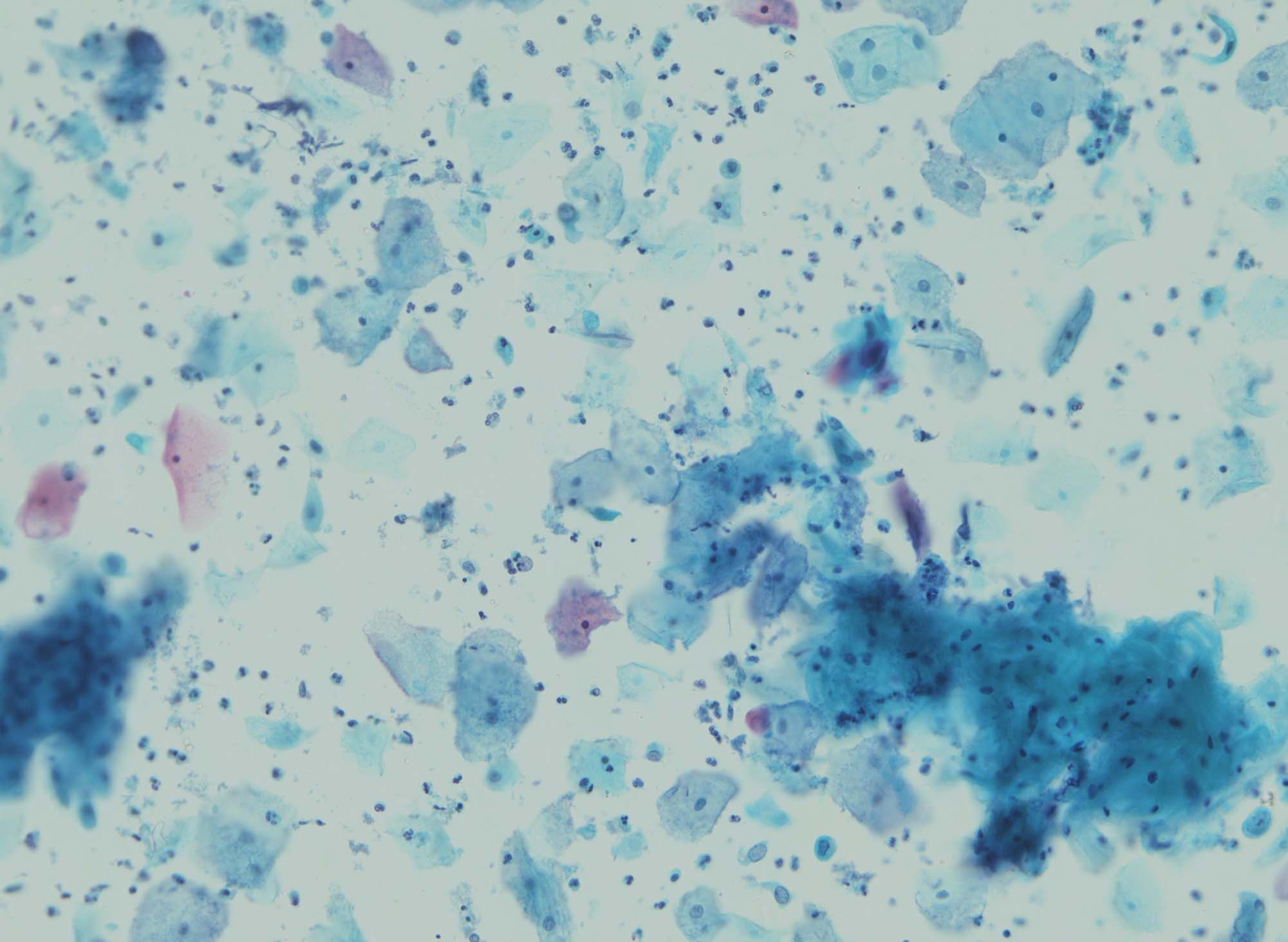}
        \caption{}   \label{s8}
    \end{subfigure}
     \caption{Samples from our unregistered multi-focus 4K ultra HD microscopic images dataset.}\label{source1}
\end{figure}
 \section{Experiments and Discussion}\label{experiment}
To demonstrate the effectiveness and efficiency of the proposed method, we conduct a series of experiments on the dataset collected by ourselves, consisting of 10 groups of unregistered multi-focus 4K ultra HD microscopic images with size of $4112 \times 3008$, which are available online along with the source code \footnote{https://github.com/yiqingmy/JointRF\label{foot}} implemented in C++. Some example images are shown in Figure \ref{source1}. We first investigate the effect of different parameter settings and then compare our method with several state-of-the-art multi-focus image fusion methods including guided filtering (GFF) \cite{li2013imageGFF}, the dense SIFT (DSIFT) \cite{liu2015multi}, and multi-scale weighted gradient (MWGF) \cite{zhou2014multi}, both in terms of the visual performance and running time. These compared methods can be downloaded online and all of them are implemented in Matlab.

The number of octaves $O$ and number of layers $L$ sampled per octave will produce an effect on both the feature points detection and saliency map construction directly, and therefore on the final fusion result. In addition, the dimensionality of U-SURF descriptor will influence on the matching accuracy and speed. We evaluate the performance of our method under varying these parameters in terms of both matching accuracy and visual quality of the fused results. Here the matching accuracy metric is determined by the ratio of the number of "inliers" ($V_{num}$) and the number of all matched pairs ($A_{num}$), i.e.
\begin{equation}\label{fused}
  Accuracy=V_{num}/A_{num}.
\end{equation}

As usual, in our implementation only the top 20 matched pairs are used for hough voting and the matching accuracy evaluation.
\begin{figure}[t]
 \centering
  \includegraphics[width=0.5\textwidth]{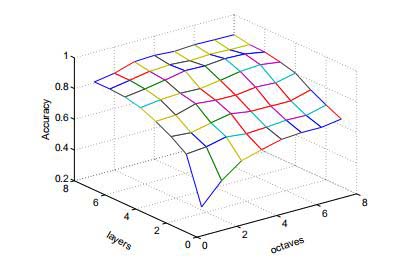}
  \caption{The matching accuracy under varying $O$ and $L$.}
  \label{acc_ol}
\end{figure}
\begin{figure}[t]
    \centering
    \begin{subfigure}[b]{0.24\textwidth}
        \includegraphics[width=\textwidth]{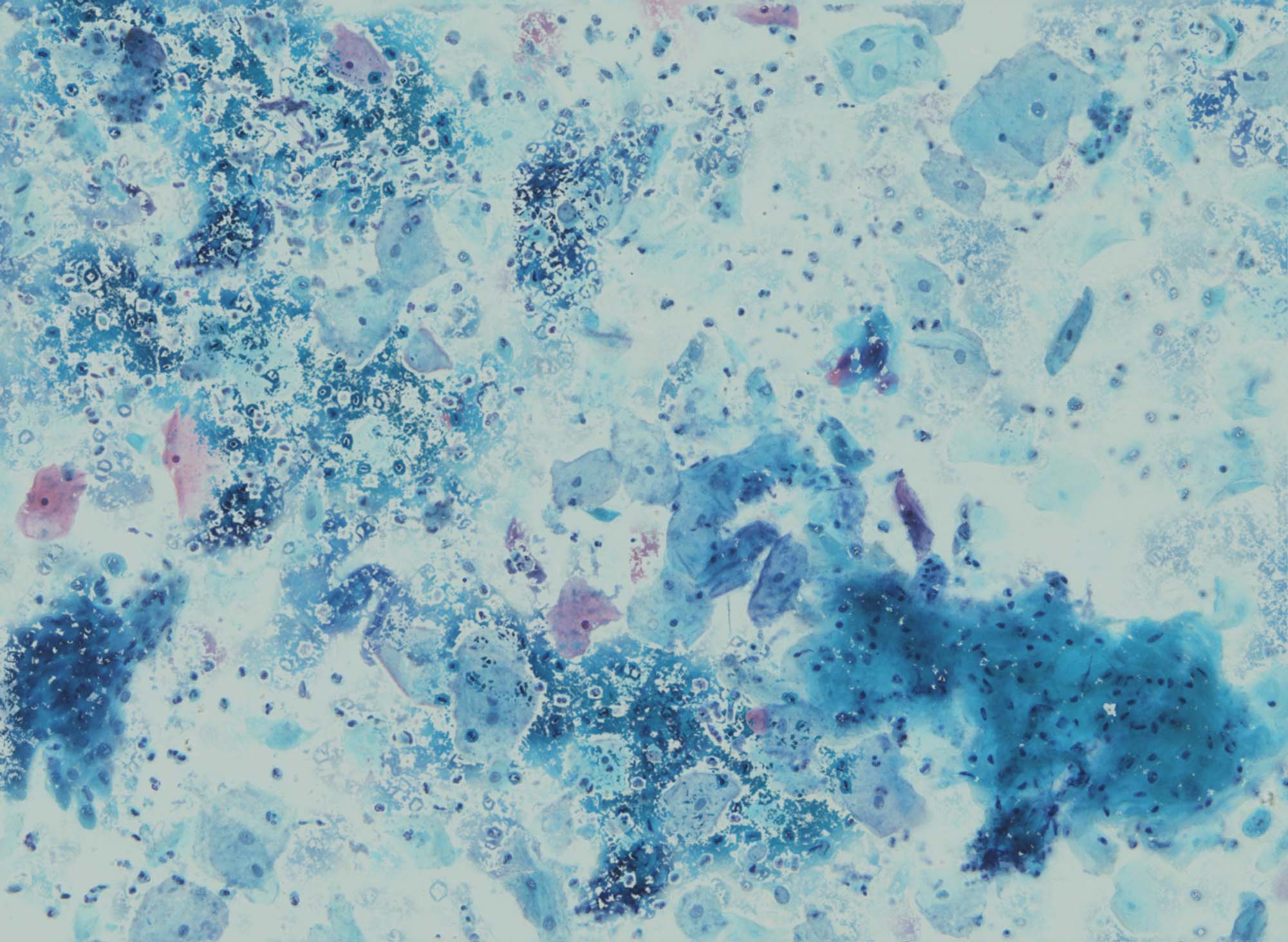}
        \caption{$O$1$L$8} \label{O1L8}
    \end{subfigure}
    \begin{subfigure}[b]{0.24\textwidth}
        \includegraphics[width=\textwidth]{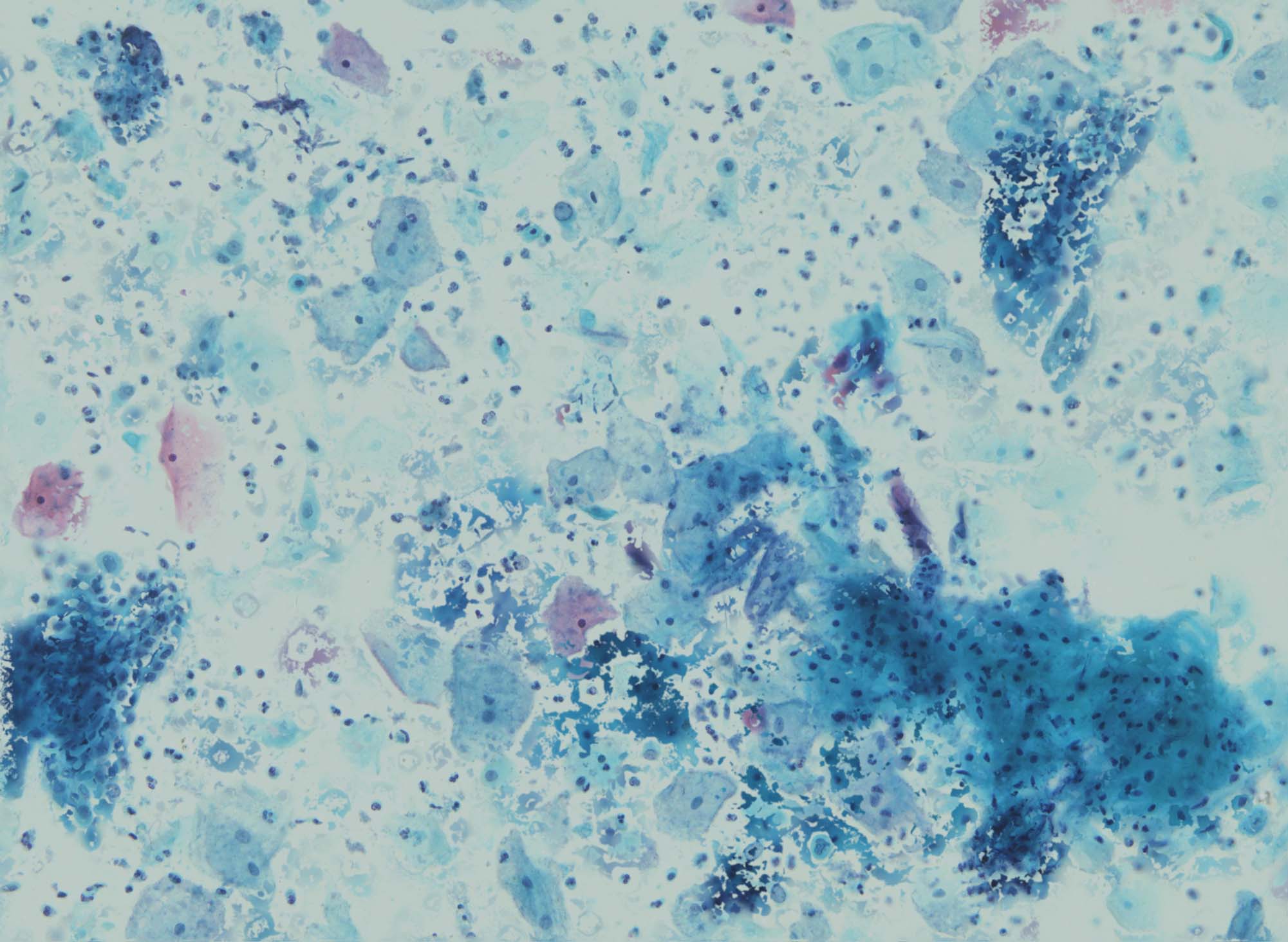}
        \caption{$O$2$L$8}  \label{O2L8}
    \end{subfigure}
    \begin{subfigure}[b]{0.24\textwidth}
        \includegraphics[width=\textwidth]{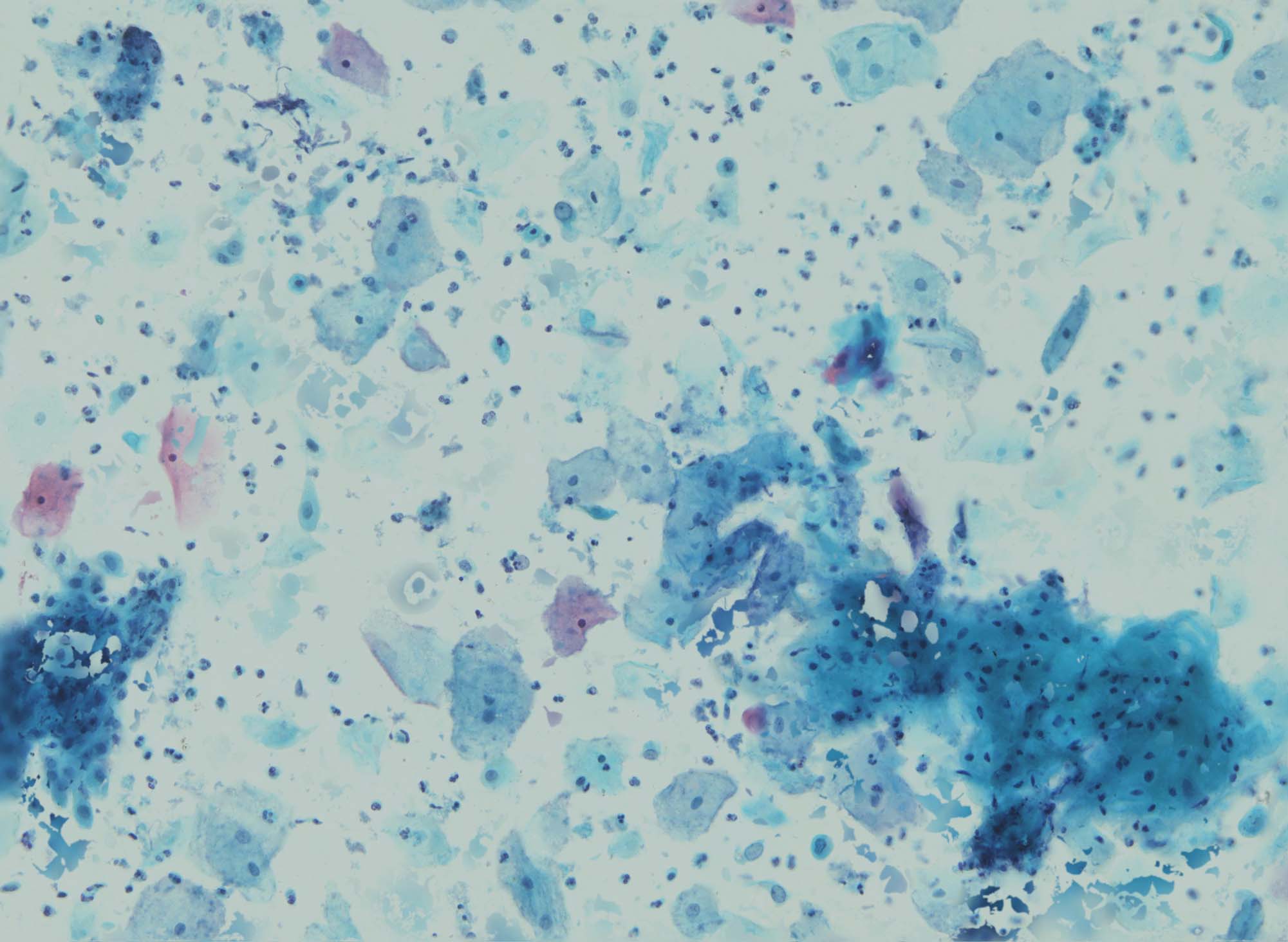}
        \caption{$O$3$L$8}    \label{O3L8}
    \end{subfigure}
     \begin{subfigure}[b]{0.24\textwidth}
        \includegraphics[width=\textwidth]{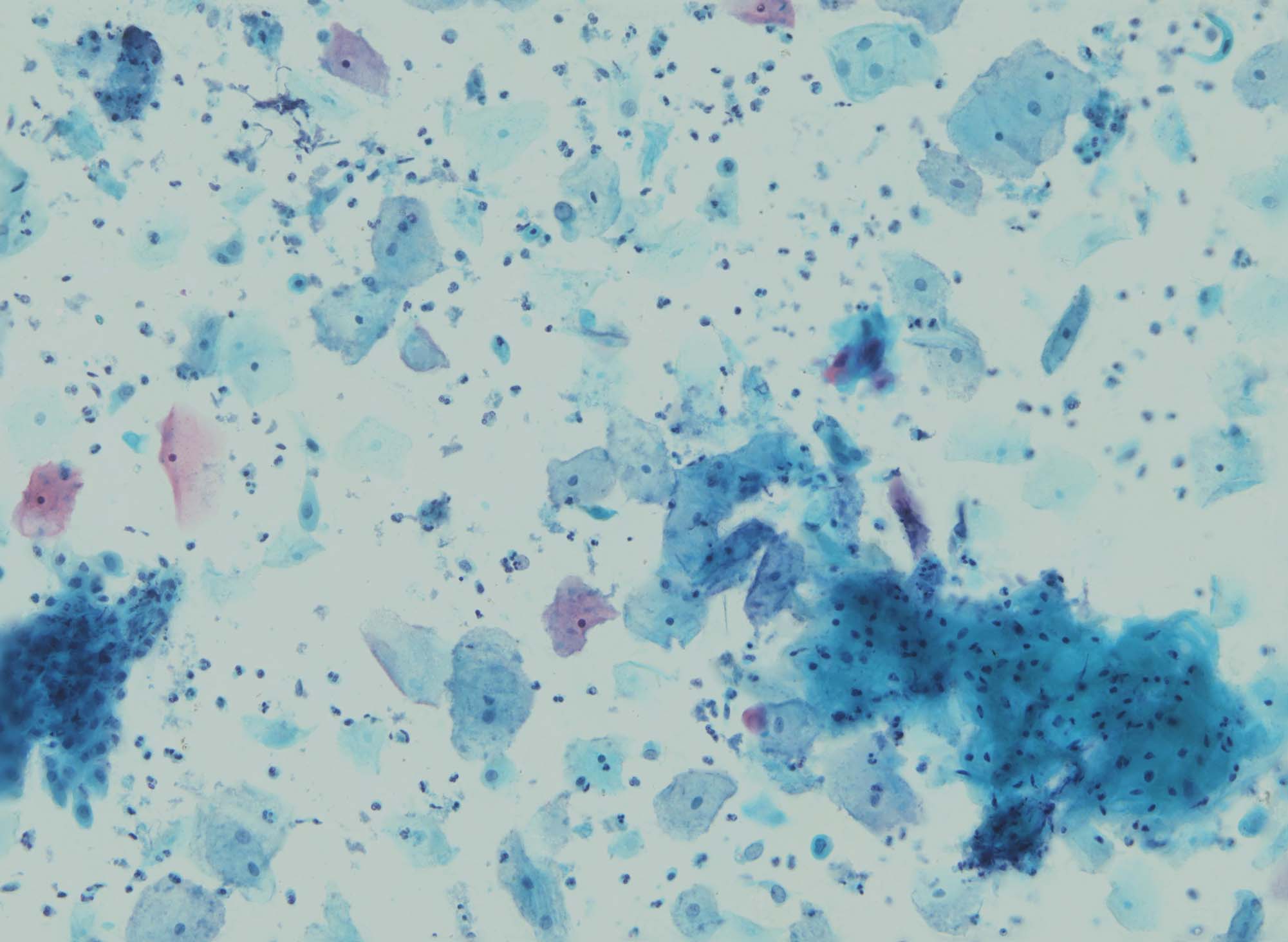}
        \caption{$O$4$L$3}   \label{O4L3}
    \end{subfigure}\\
    \begin{subfigure}[b]{0.24\textwidth}
        \includegraphics[width=\textwidth]{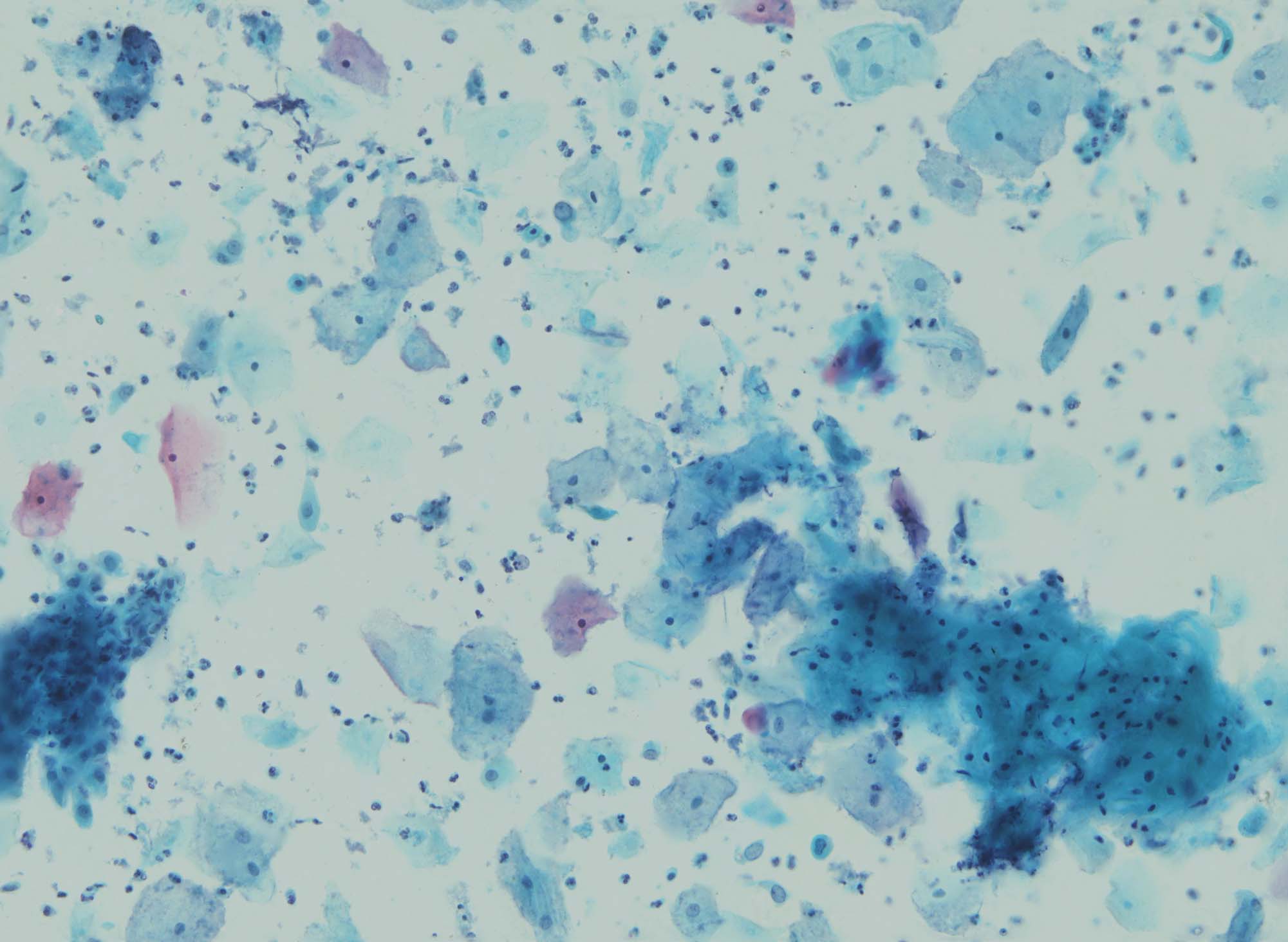}
        \caption{$O$5$L$3} \label{O5L3}
    \end{subfigure}
    \begin{subfigure}[b]{0.24\textwidth}
        \includegraphics[width=\textwidth]{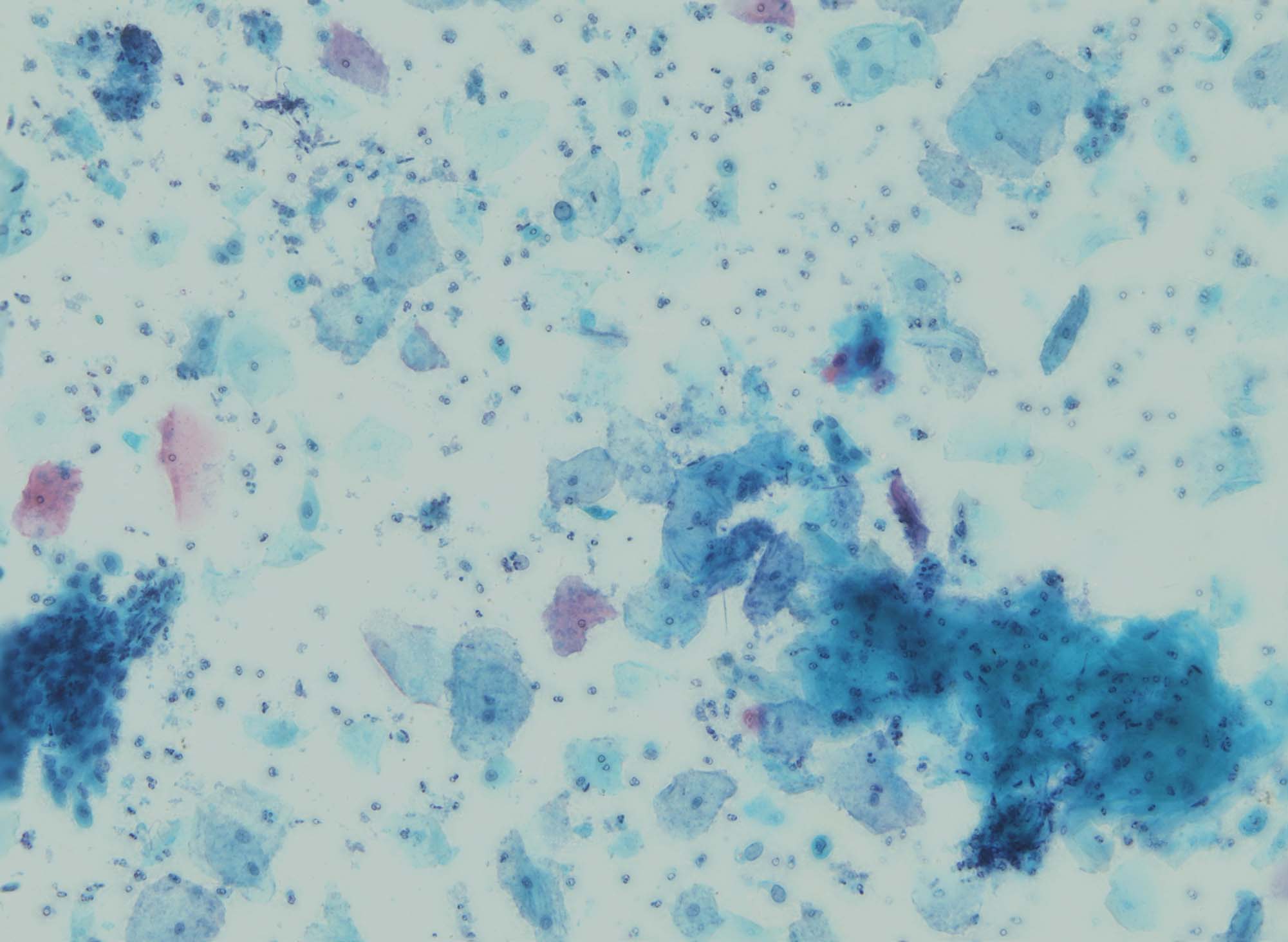}
        \caption{$O$6$L$3}  \label{O6L3}
    \end{subfigure}
    \begin{subfigure}[b]{0.24\textwidth}
        \includegraphics[width=\textwidth]{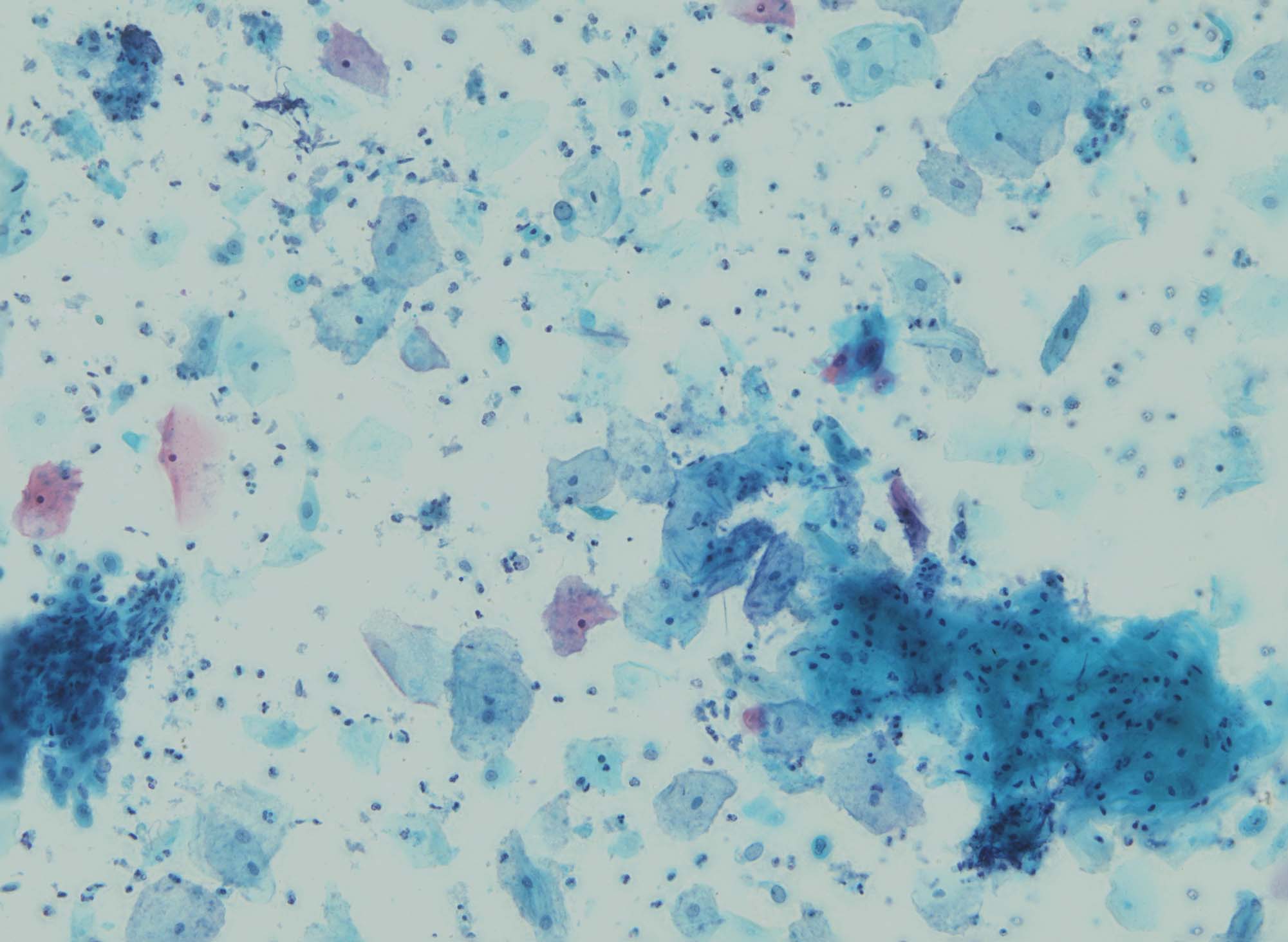}
        \caption{$O$7$L$3}    \label{O7L3}
    \end{subfigure}
    \begin{subfigure}[b]{0.24\textwidth}
        \includegraphics[width=\textwidth]{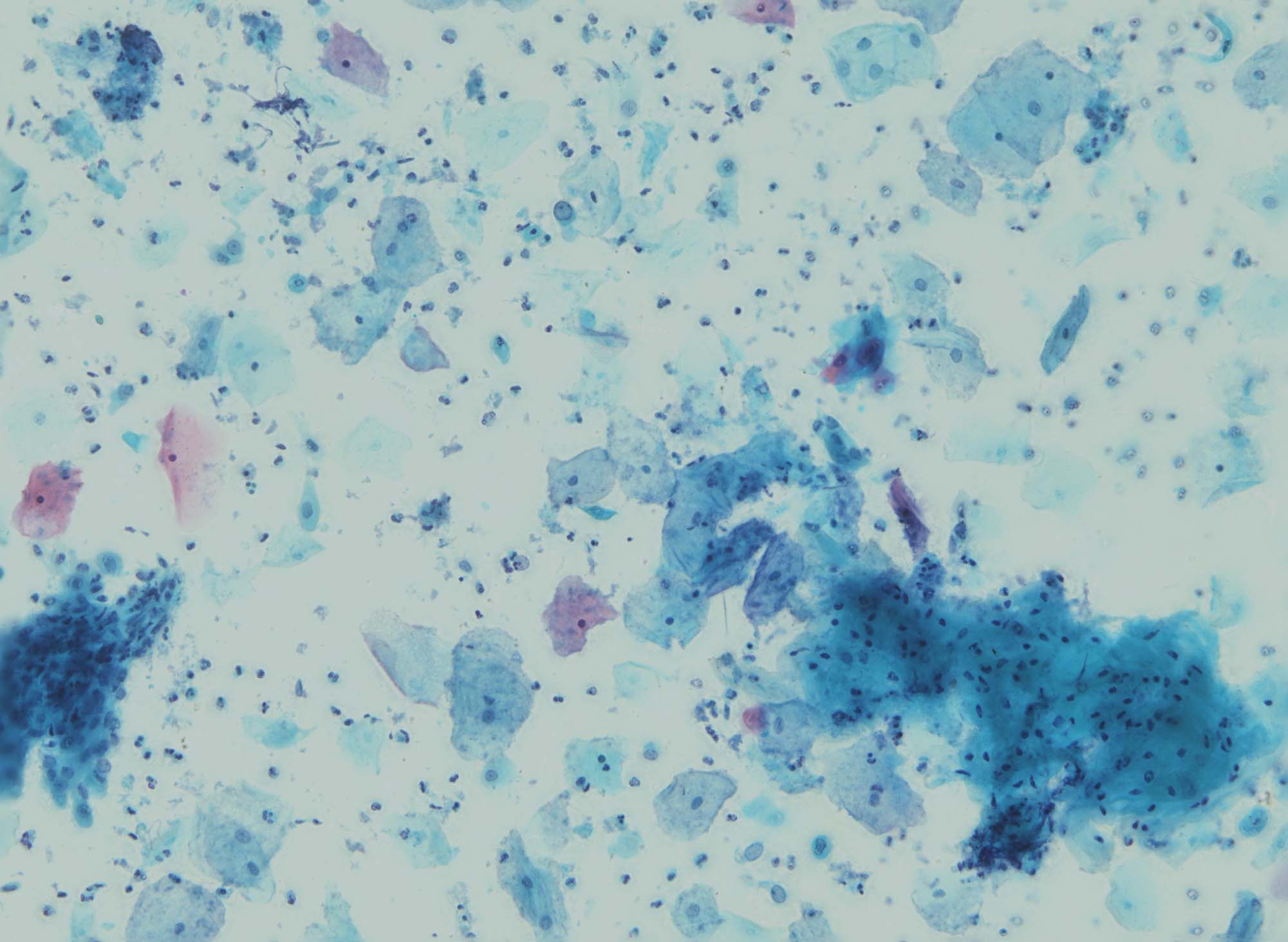}
        \caption{$O$8$L$3}   \label{O8L3}
    \end{subfigure}\\

    \begin{subfigure}[b]{0.24\textwidth}
        \includegraphics[width=\textwidth]{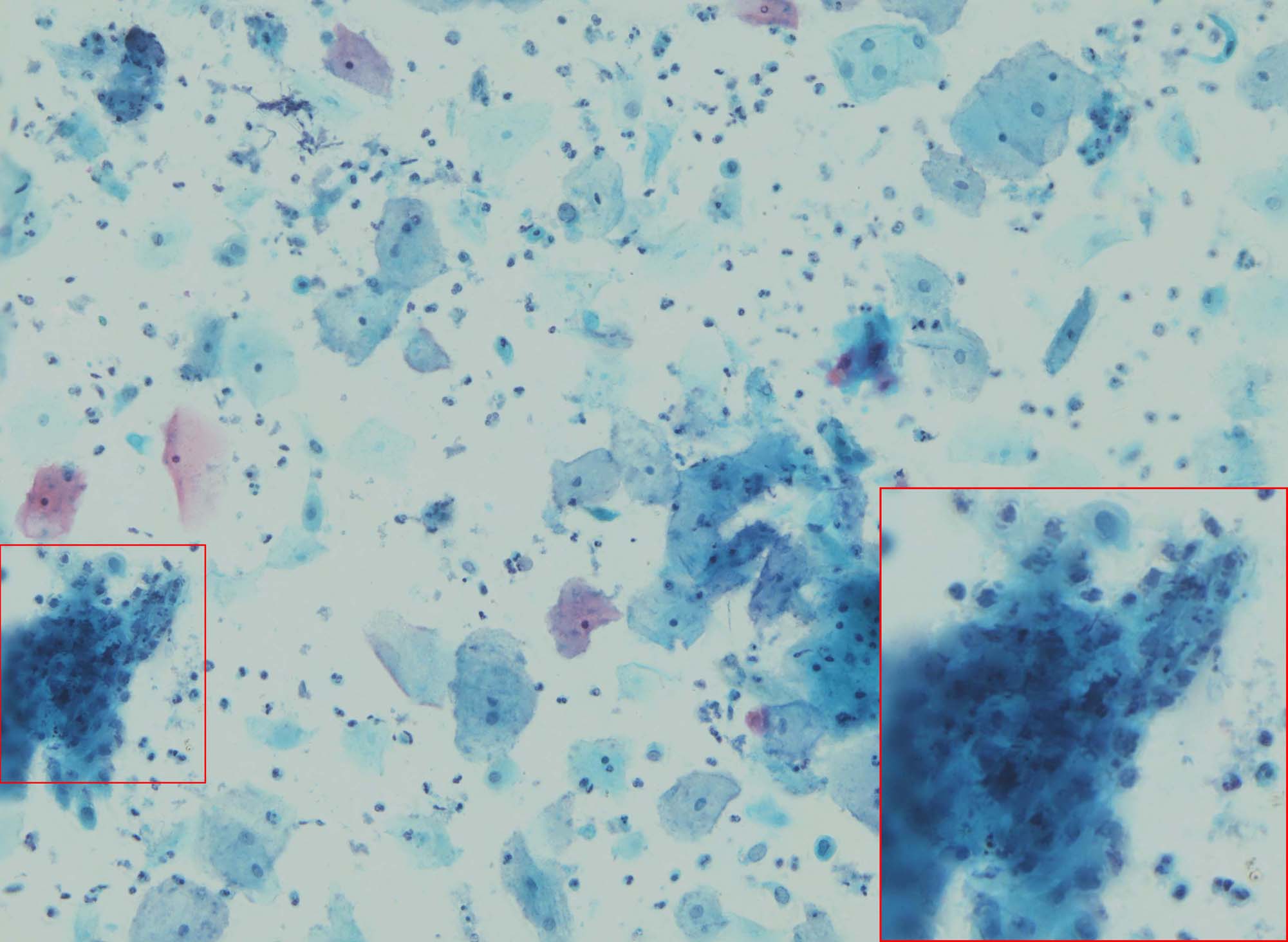}
        \caption{$O$5$L$1} \label{O5L1}
    \end{subfigure}
   \begin{subfigure}[b]{0.24\textwidth}
        \includegraphics[width=\textwidth]{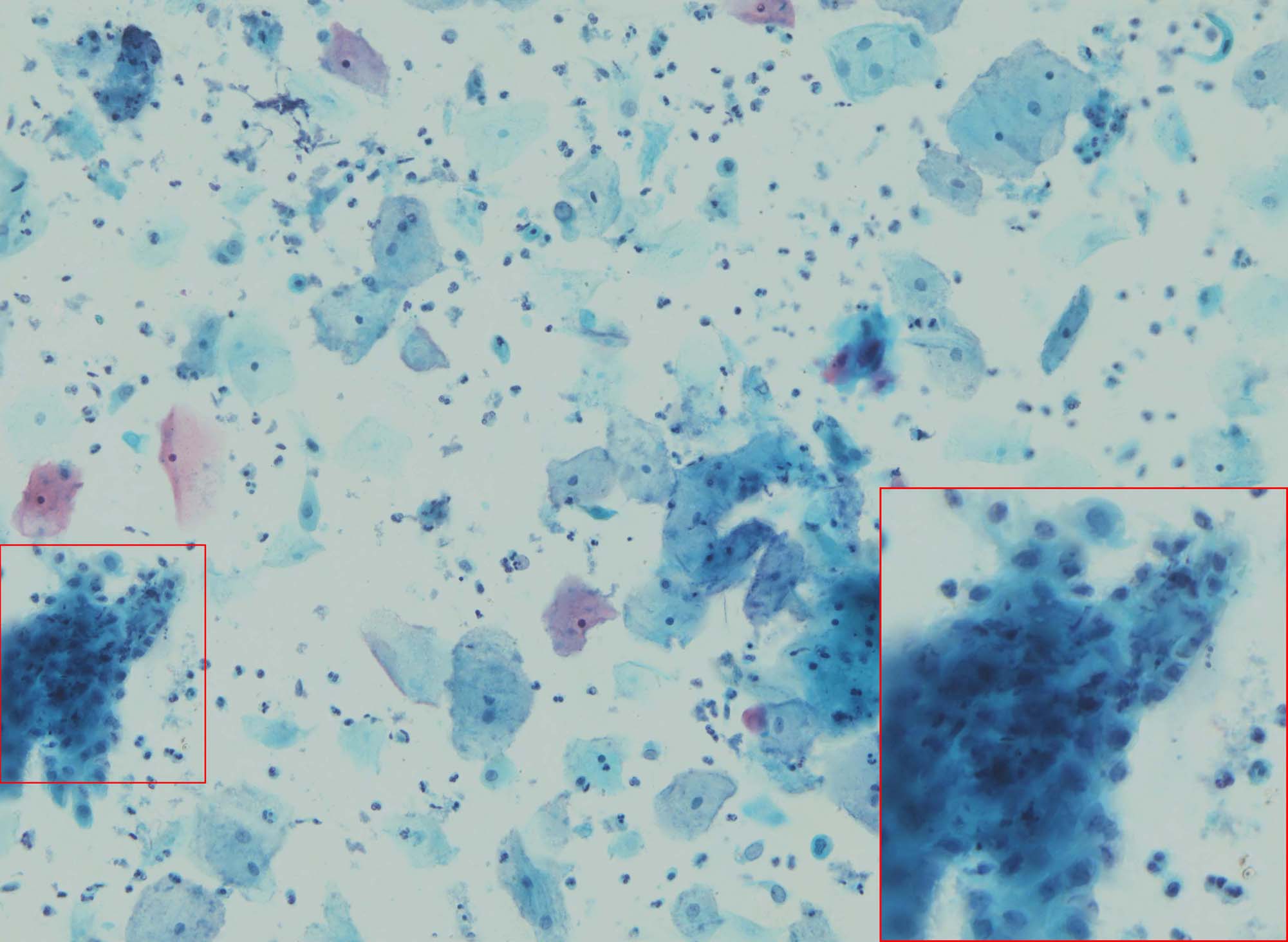}
        \caption{$O$5$L$2}   \label{O5L2}
    \end{subfigure}
       \begin{subfigure}[b]{0.24\textwidth}
        \includegraphics[width=\textwidth]{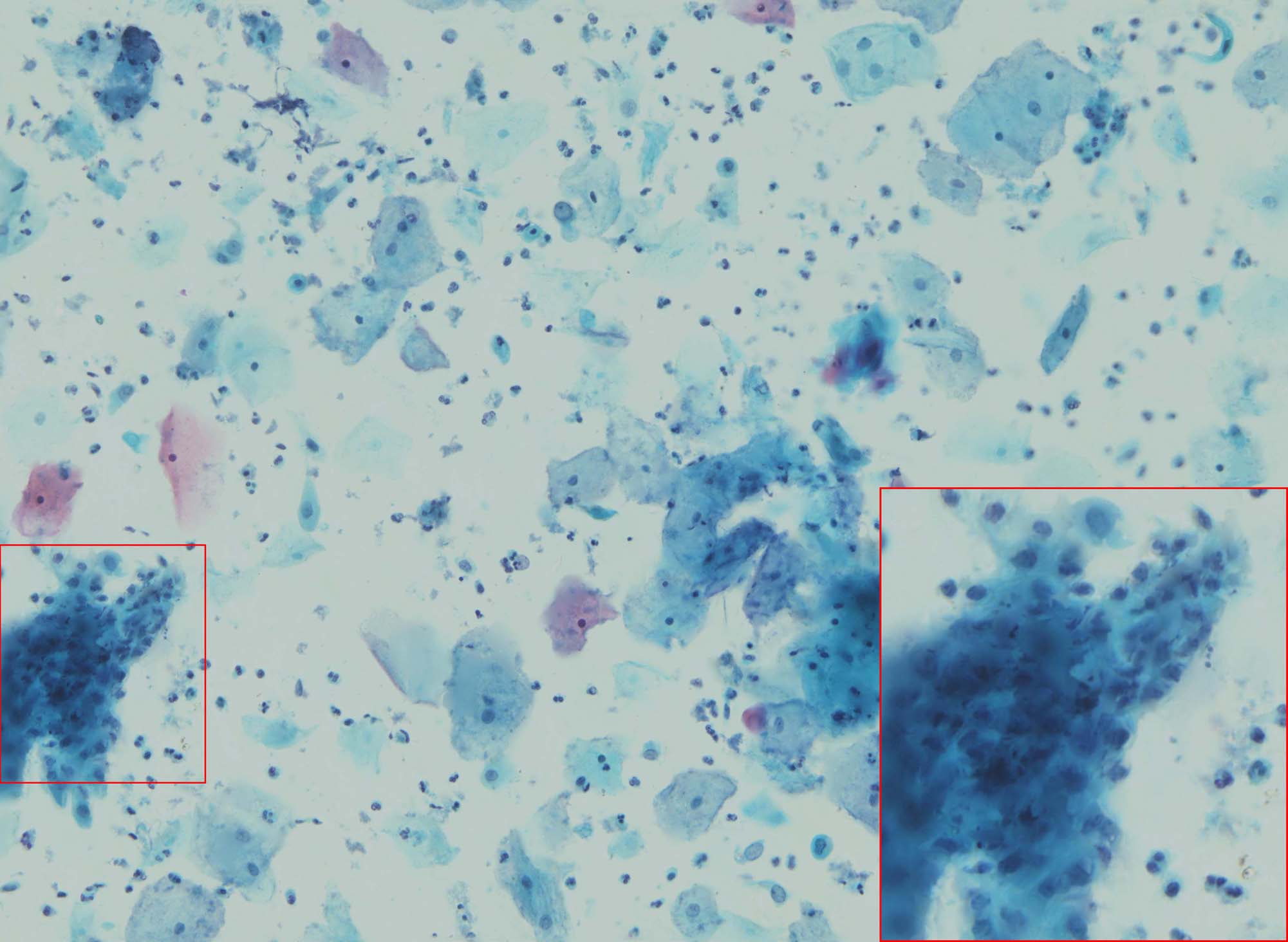}
        \caption{$O$5$L$7}  \label{O5L7}
    \end{subfigure}
    \begin{subfigure}[b]{0.24\textwidth}
        \includegraphics[width=\textwidth]{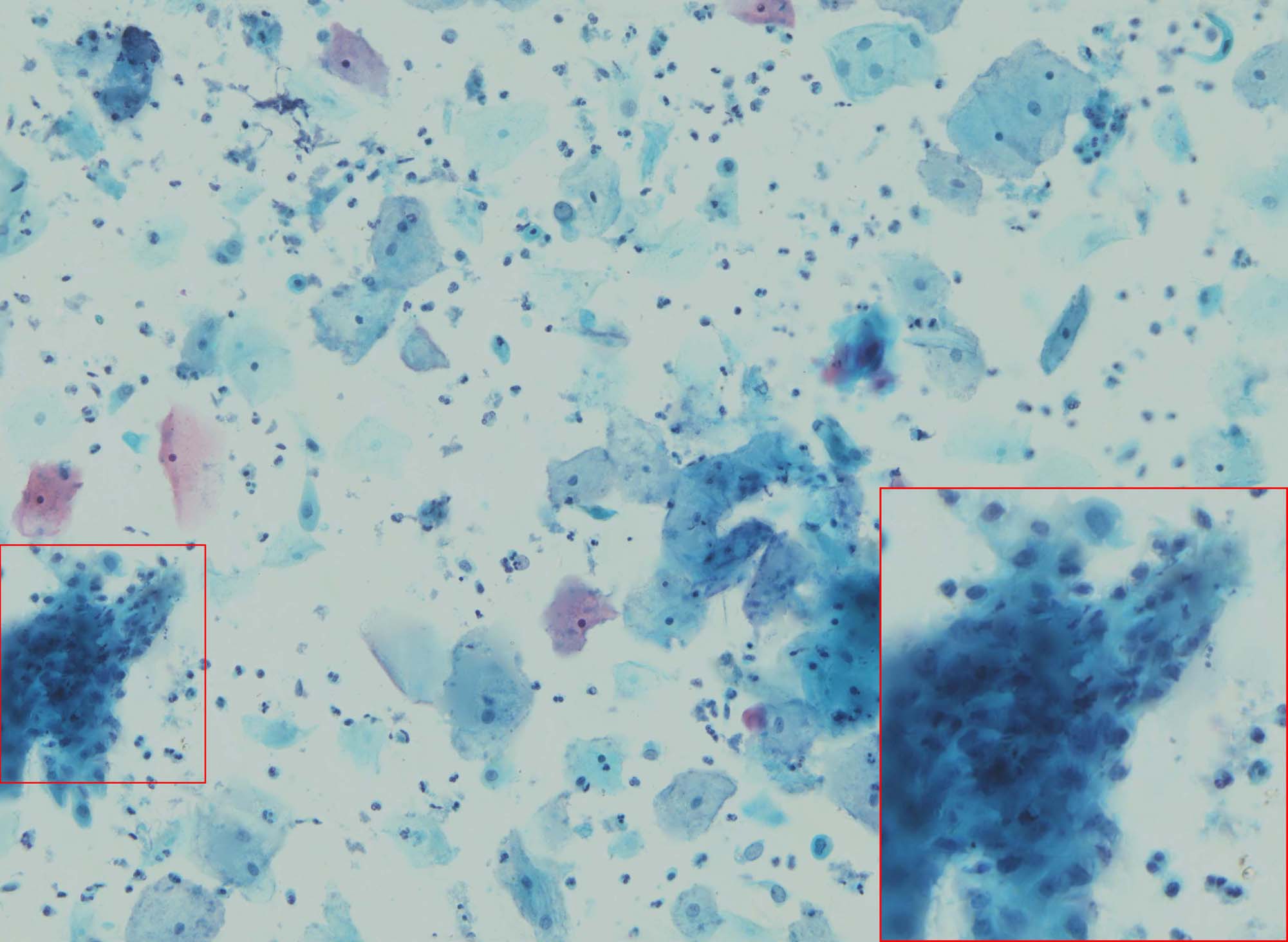}
        \caption{$O$5$L$8}    \label{O5L8}
    \end{subfigure}

    \caption{ The fused results with different octave and layer numbers. $O1L8$ means the octave number is 1 and the layer number is 8, similar to (b)-(l). For the images in the third row, the red bounding box in the bottom right corner is the magnification of the corresponding region.}\label{Fig8}
\end{figure}
The accuracy under varying total number of octaves $O$ and number of layers $L$ sampled per octave is shown in Figure \ref{acc_ol}, where the dimension of the U-SURF descriptors is 64. We can see that the accuracy is improved globally when the number of octaves and layers are increased, and finally saturated. Meanwhile, Figure \ref{Fig8} shows the final fused results based on the source images shown in Figure \ref{source1} with different $O$ and $L$. It can be shown that when the number of octaves $O$ is less than 4, although the value of $L$ is large, there are lots of obvious artifacts in the fused images (See Figure \ref{O1L8}-\ref{O3L8}), whereas with the increasing of the value of $O$ with a few layers in each octave, the visual quality of the fused results is better and there are not obvious distortion and artifacts, as shown in Figure \ref{O4L3}-\ref{O8L3}. Since there are no obvious distortion and artifacts when $O \geq 5$ covering all the number of layers, we fix the value of $O$ to 5 when evaluating the effect of the number of layers $L$ per octave. As shown in the last row of Figure \ref{Fig8}, the visual quality is improved obviously by vary the value of $L$ from 1 to 2, where the distortion and artifacts are prominently decreased (refer to the magnified region in Figure \ref{O5L1} and Figure \ref{O5L2}). The performance can be further improved slightly by set the $L=7$ or $L=8$, as shown in the magnified region in Figure \ref{O5L7} and Figure \ref{O5L8}. In the following experiments, we choose to set $O=5, L=2$.
\begin{table}[h]
  \centering
  \caption{The average accuracy of image matching and matching time with different dimensions of descriptors.}\label{dimension}
  \begin{tabular}{rrrrrr}
  \hline
  Dimension(D) &    16 & 36 &    64 &   100 &   144  \\
  \hline
  Accuracy &     0.293  & 0.768  &     0.813  &     0.057  &     0.057    \\
  Matching Time(s) &   0.002  &  0.003  &     0.006  &     0.008  &     0.014    \\
  \hline
  \end{tabular}
\end{table}

\begin{figure}[t]
    \centering
    \begin{subfigure}[b]{0.24\textwidth}
        \includegraphics[width=\textwidth]{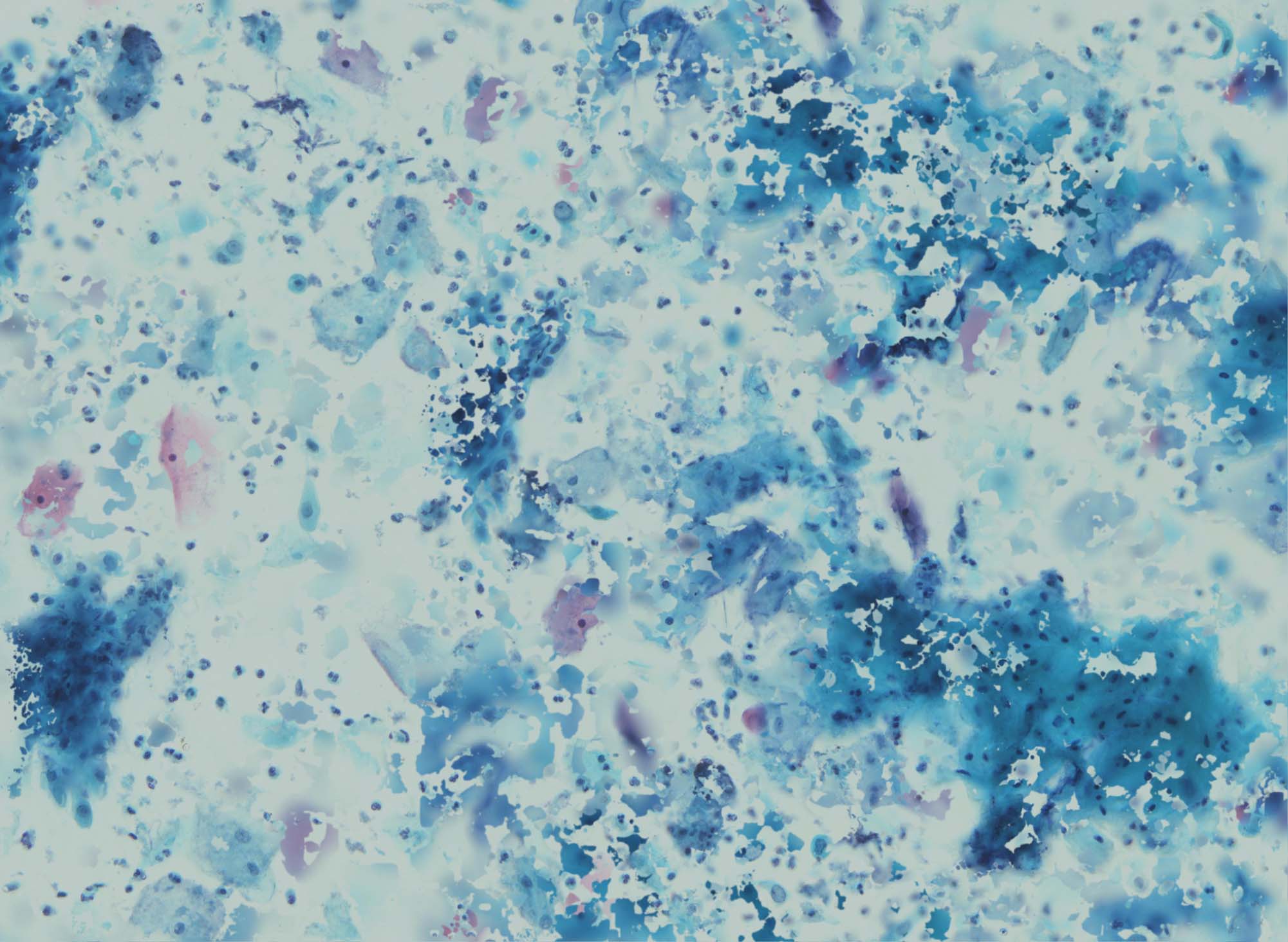}
        \caption{$16D$} \label{16O5S2}
    \end{subfigure}
    \begin{subfigure}[b]{0.24\textwidth}
        \includegraphics[width=\textwidth]{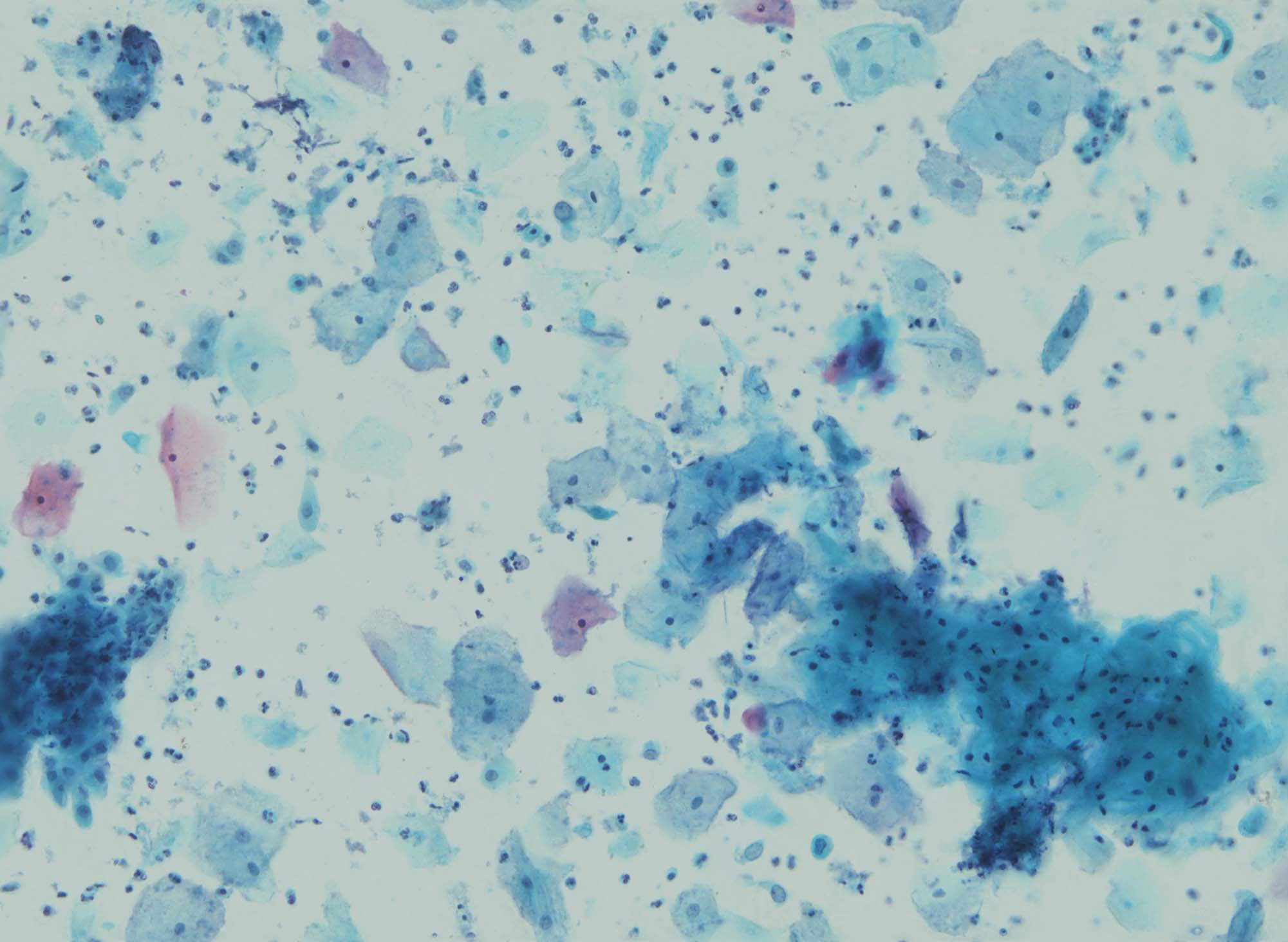}
        \caption{$36D$}  \label{36O5S2}
    \end{subfigure}
    \begin{subfigure}[b]{0.24\textwidth}
        \includegraphics[width=\textwidth]{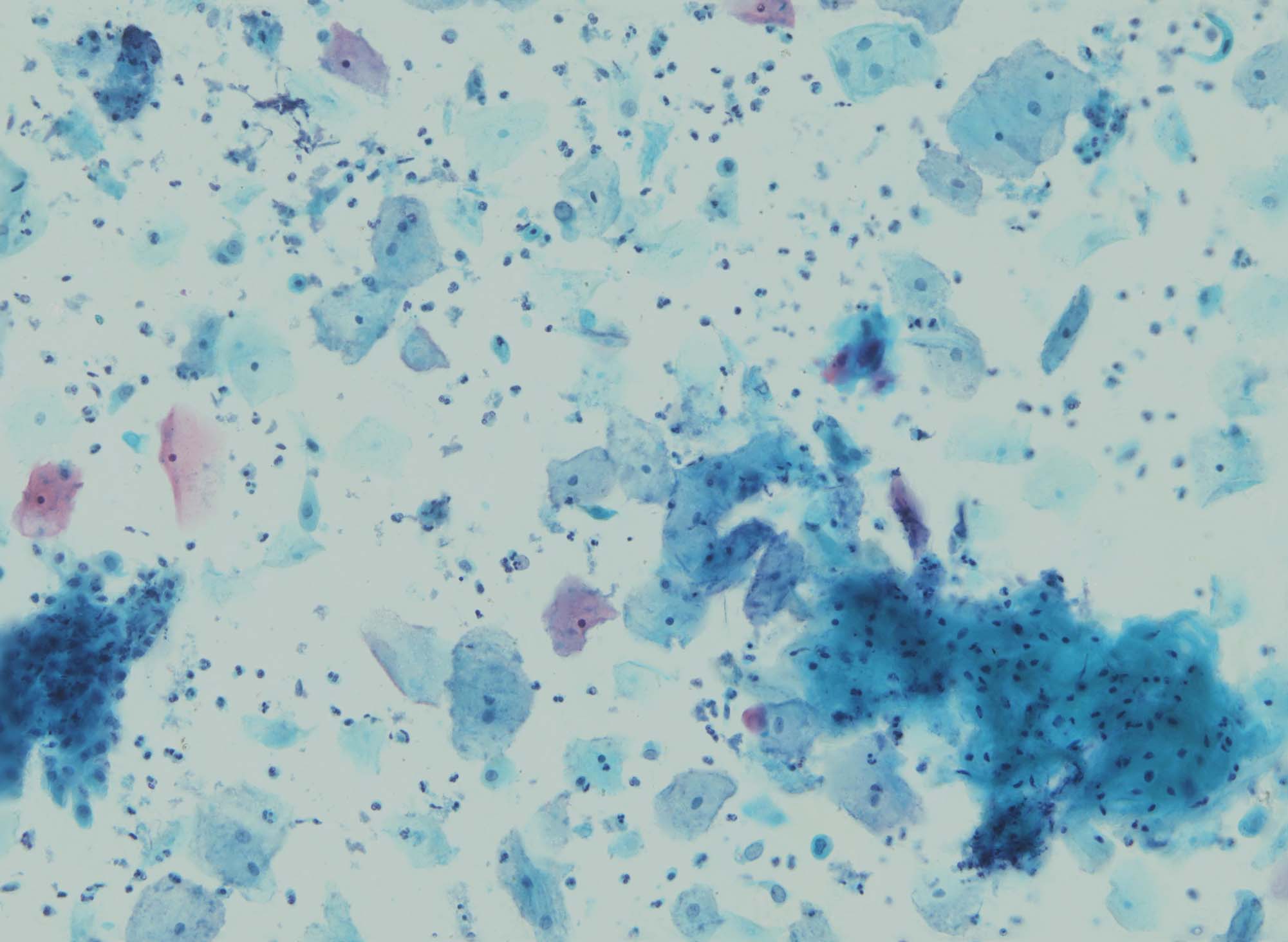}
        \caption{$64D$}    \label{64O5S2}
    \end{subfigure}
    \begin{subfigure}[b]{0.24\textwidth}
        \includegraphics[width=\textwidth]{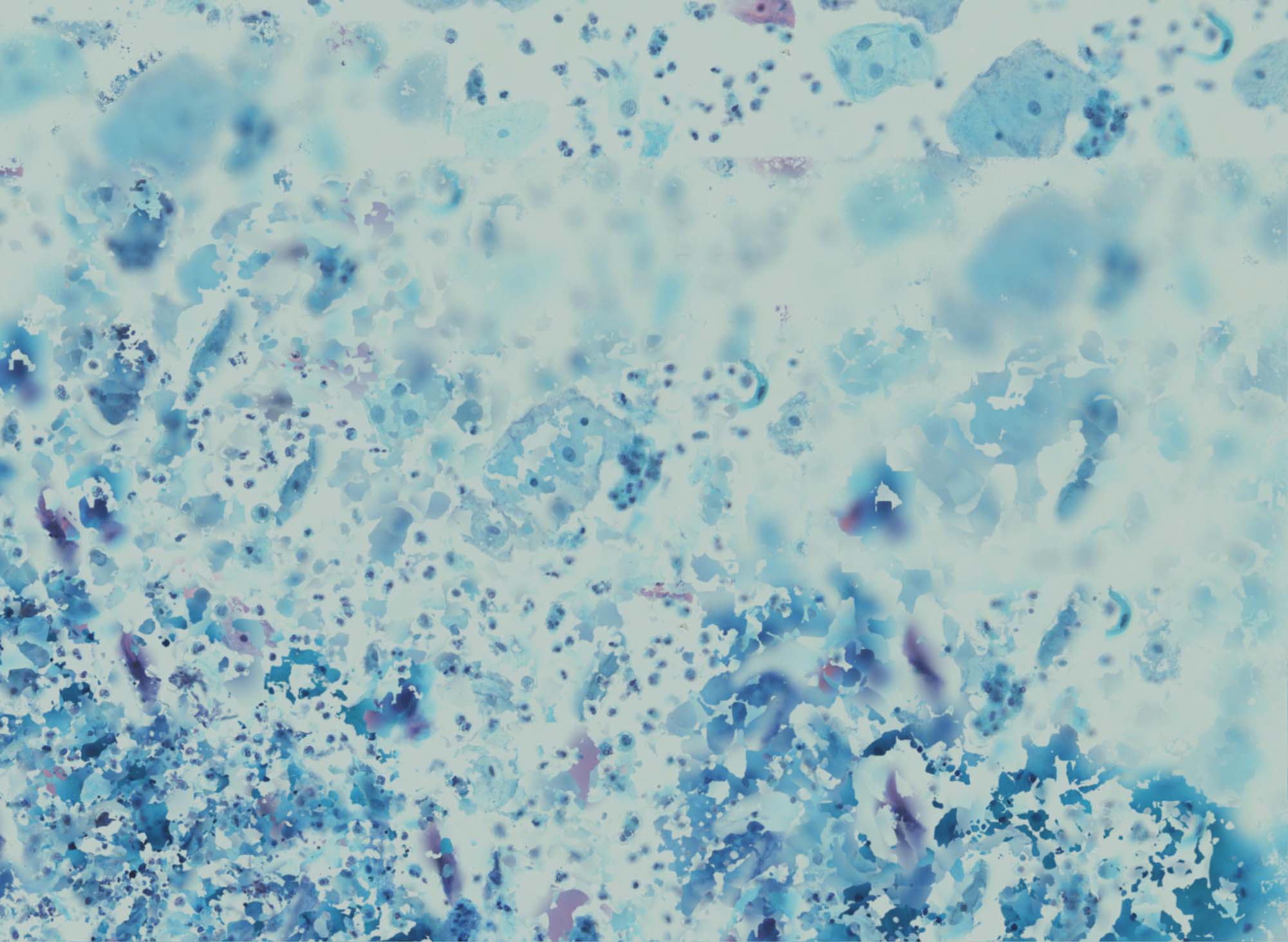}
        \caption{$100D$}   \label{100O5S2}
    \end{subfigure}
    \caption{ The fused results with different dimensions of the feature points descriptors. The number of octaves and layers is set as 5 and 2 respectively.}\label{results_dim}
\end{figure}

Then, we change the dimension of the feature descriptors and Table \ref{dimension} lists the accuracy and the time-cost of the feature matching. As the dimension increases, the matching time is increased and the accuracy can be increased gradually until the dimension is 64. Surprisedly, the accuracy would plummet when higher dimension of the U-SURF is used. The corresponding fused images are shown in Figure \ref{results_dim}, and as it shown that the visual quality is very poor when the dimension of the descriptors is too low (i.e., $D$=16) or too high ($D=100$). Therefore, we set the dimension to 64 for the following comparative experiments.

\begin{figure}[t]
    \centering
    \begin{subfigure}[b]{0.24\textwidth}
        \includegraphics[width=\textwidth]{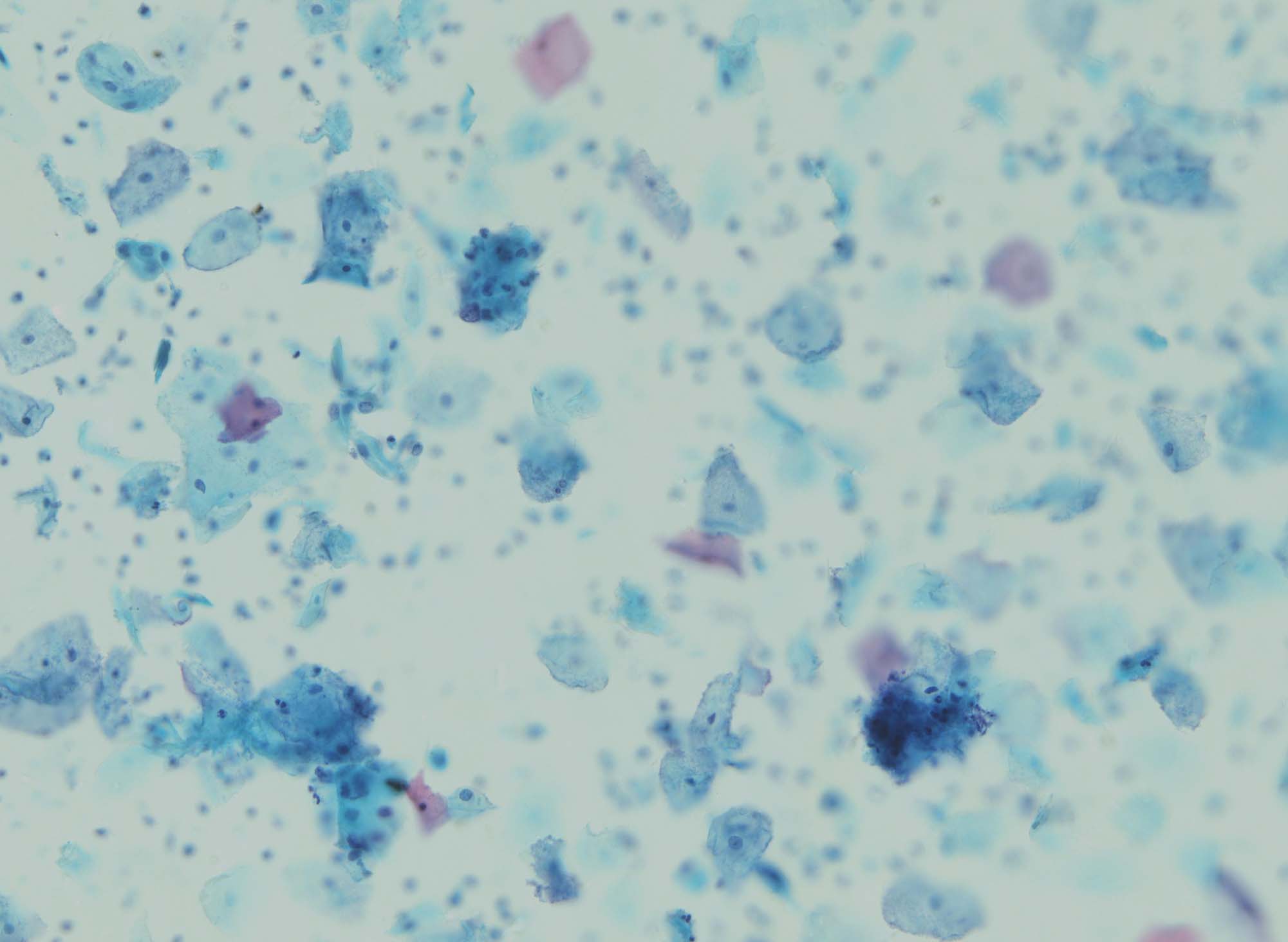}
        \caption{}    \label{02}
    \end{subfigure}
        \begin{subfigure}[b]{0.24\textwidth}
        \includegraphics[width=\textwidth]{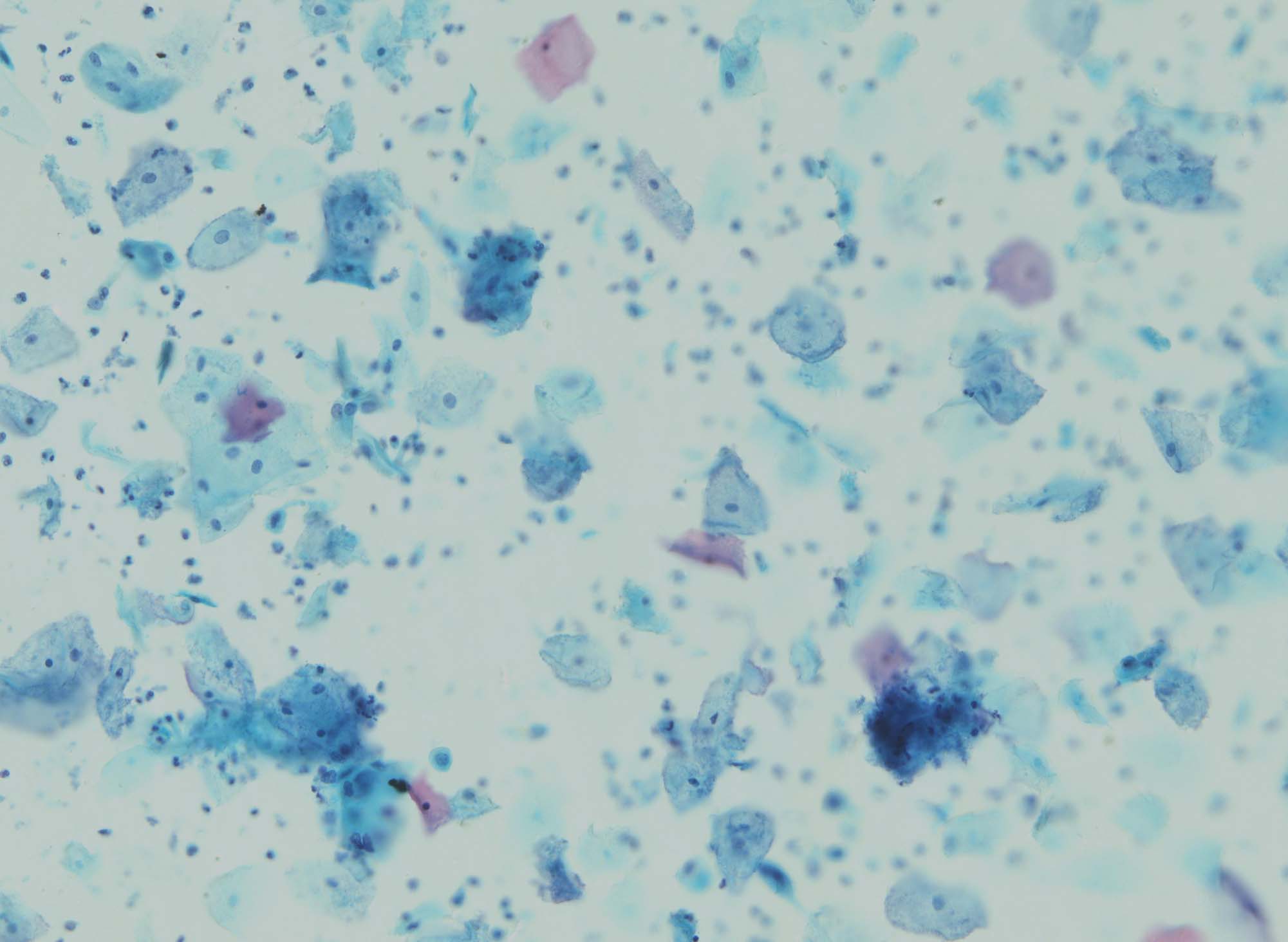}
        \caption{}   \label{03}
    \end{subfigure}
    \begin{subfigure}[b]{0.24\textwidth}
        \includegraphics[width=\textwidth]{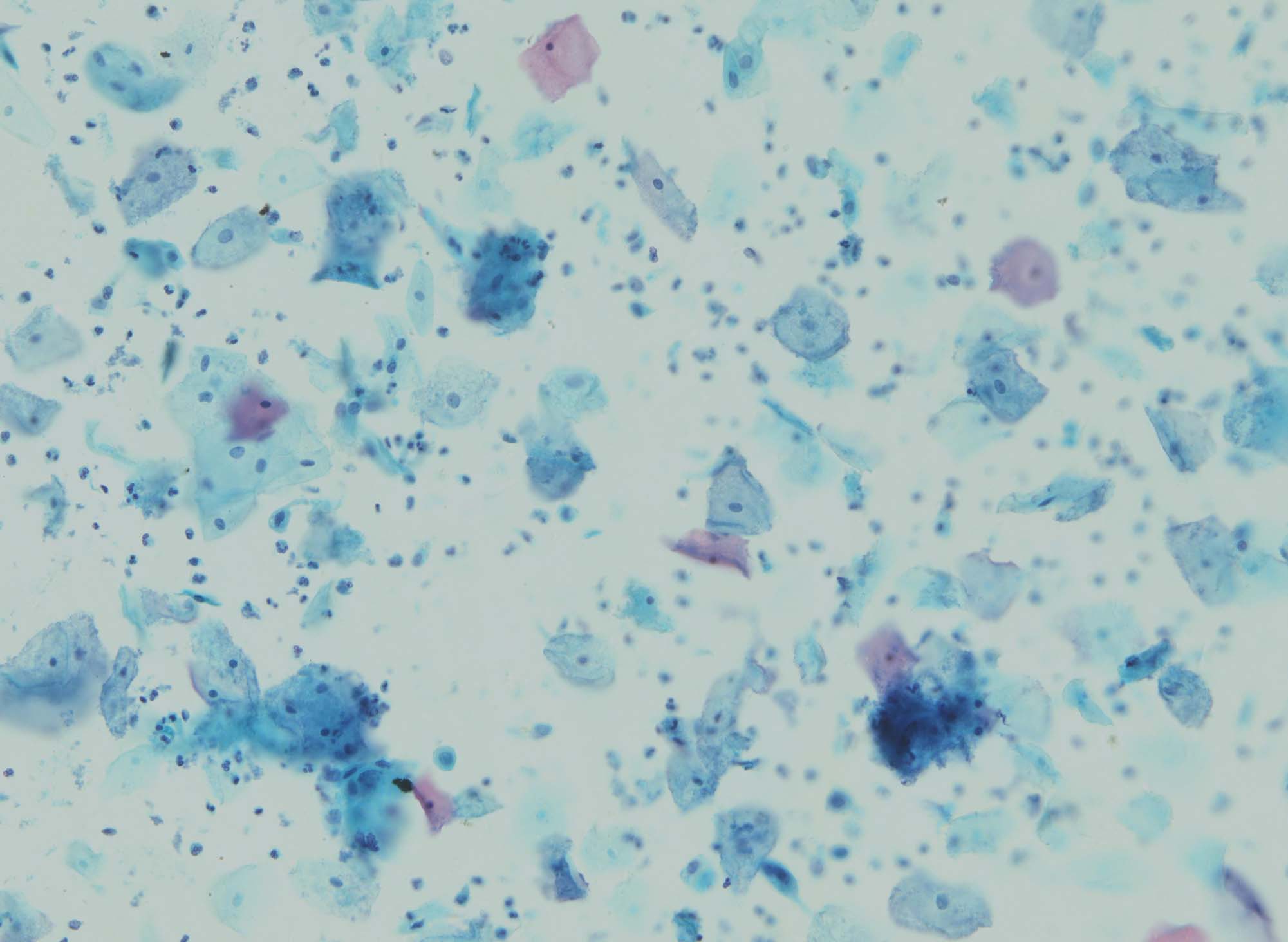}
        \caption{}   \label{04}
    \end{subfigure}
    \begin{subfigure}[b]{0.24\textwidth}
        \includegraphics[width=\textwidth]{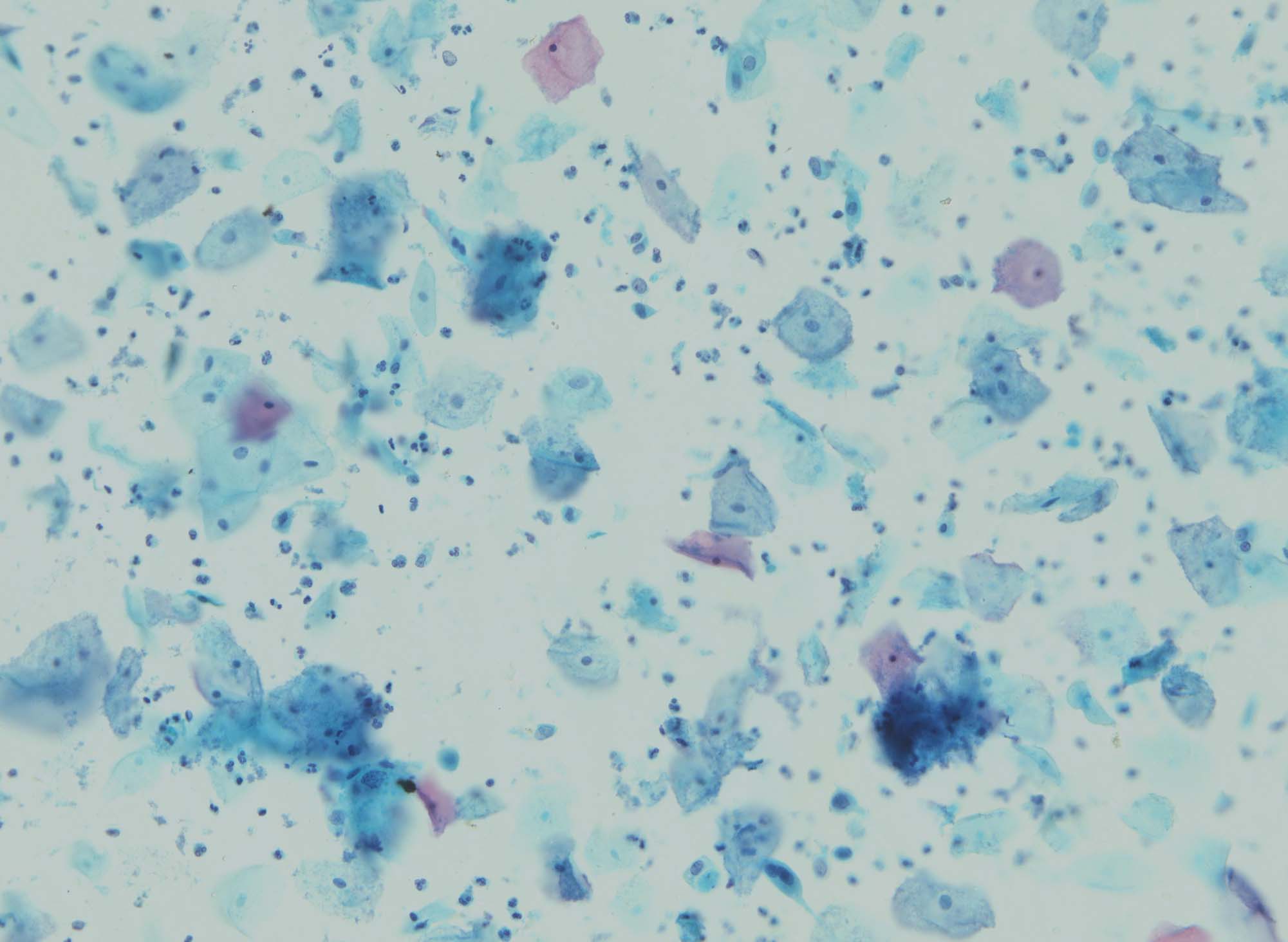}
        \caption{}   \label{05}
    \end{subfigure}\\
    \begin{subfigure}[b]{0.24\textwidth}
        \includegraphics[width=\textwidth]{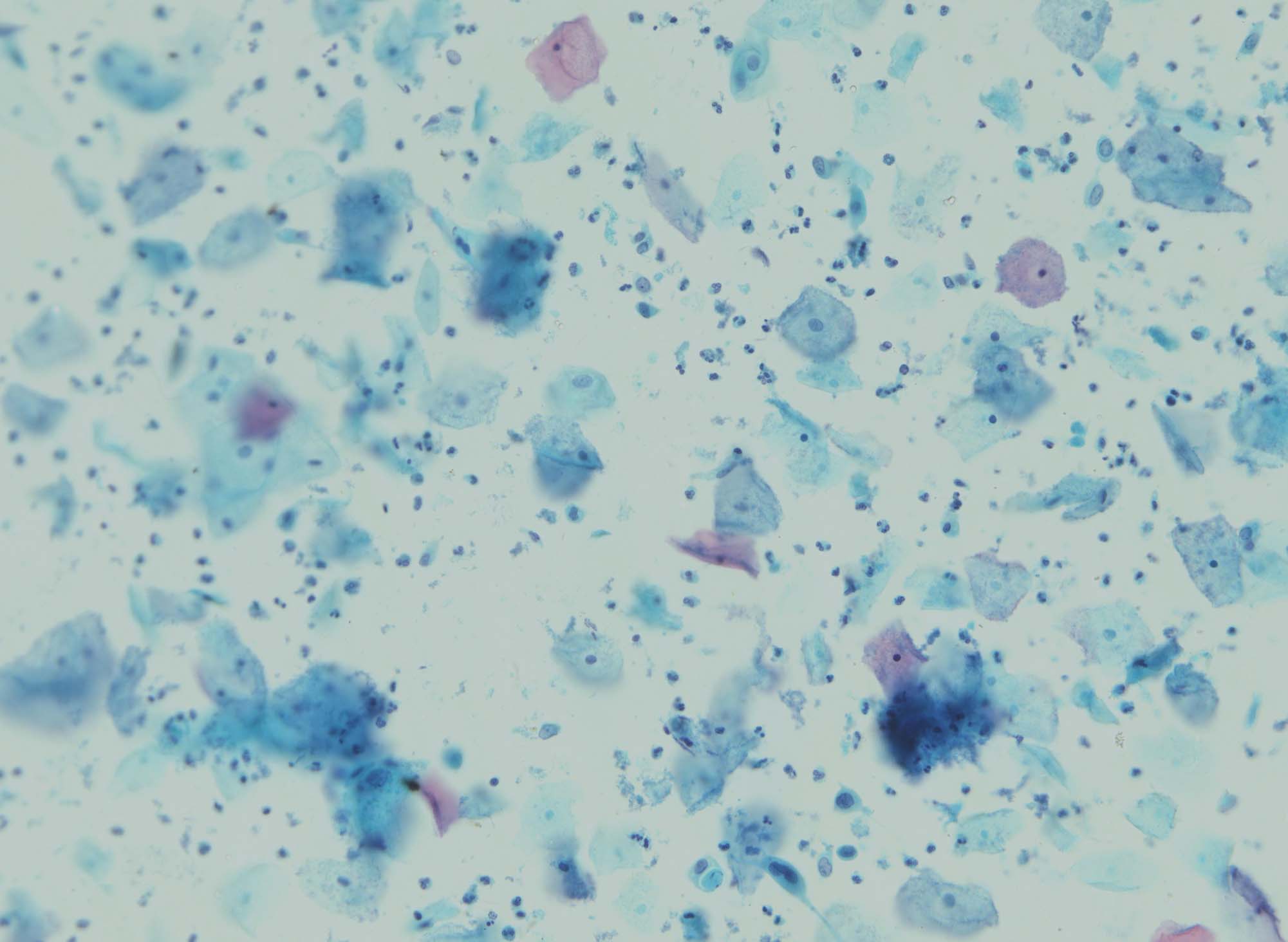}
        \caption{} \label{06}
    \end{subfigure}
    \begin{subfigure}[b]{0.24\textwidth}
        \includegraphics[width=\textwidth]{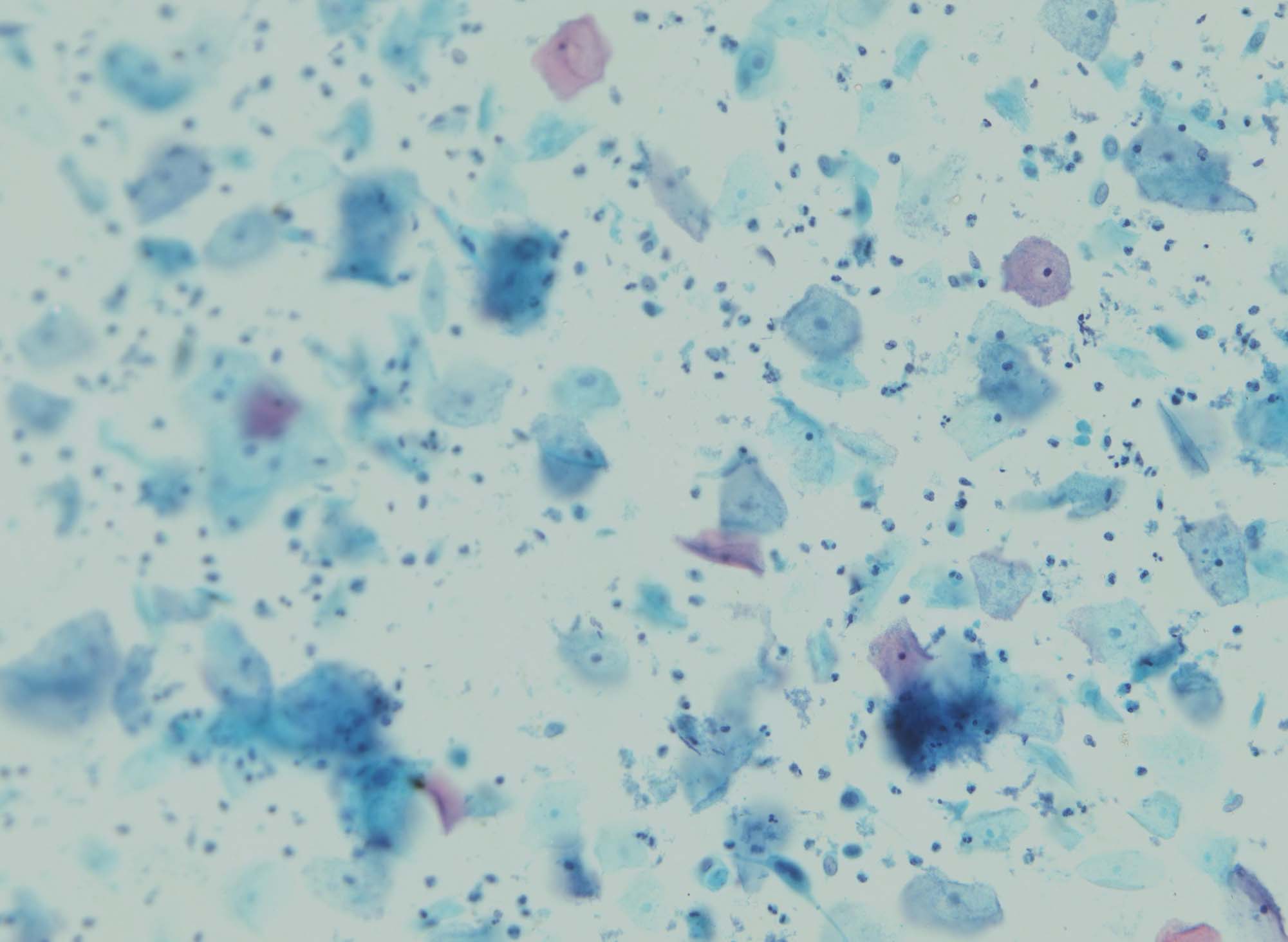}
        \caption{}  \label{07}
    \end{subfigure}
    \begin{subfigure}[b]{0.24\textwidth}
        \includegraphics[width=\textwidth]{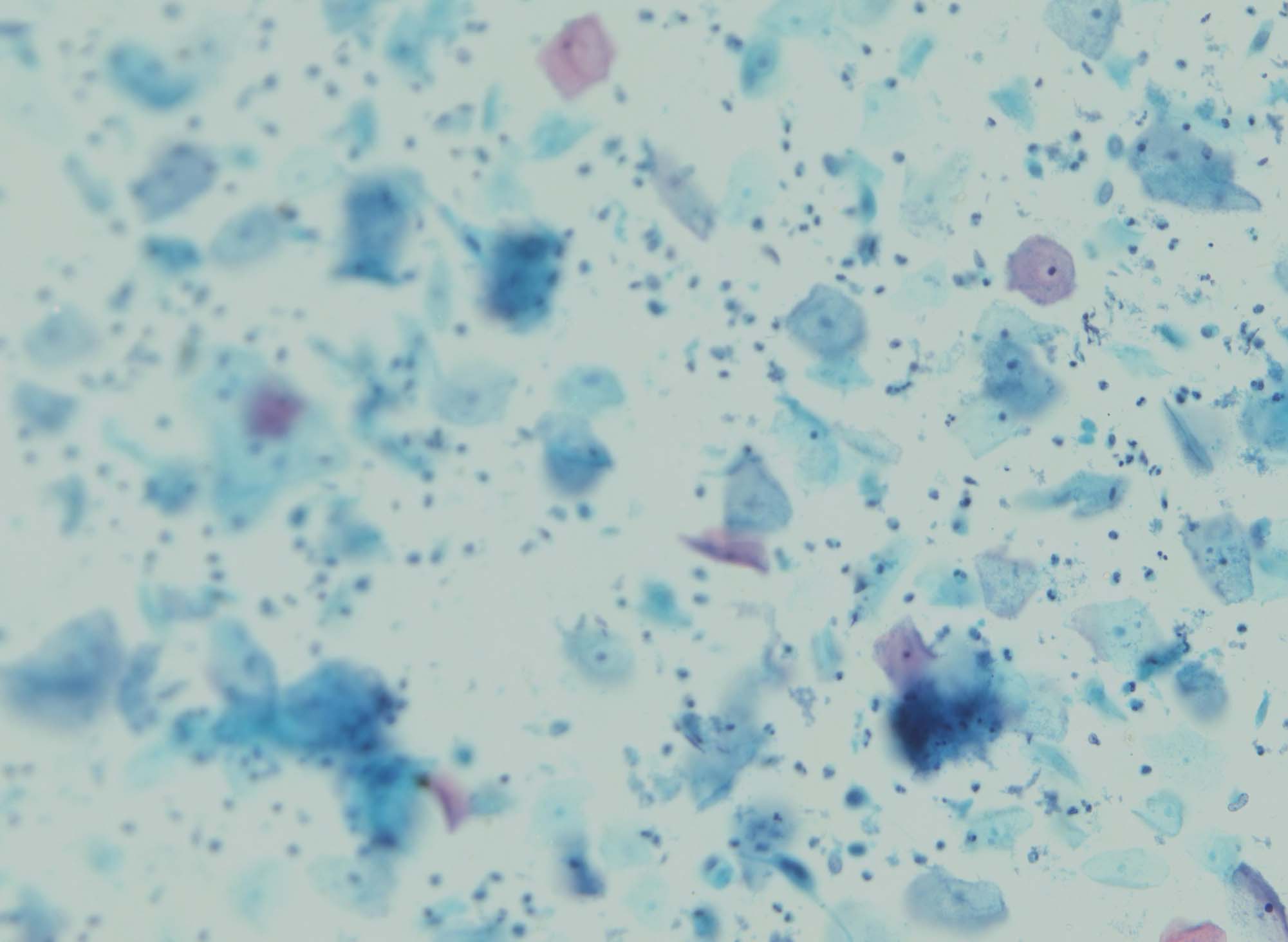}
        \caption{}    \label{08}
    \end{subfigure}
    \begin{subfigure}[b]{0.24\textwidth}
        \includegraphics[width=\textwidth]{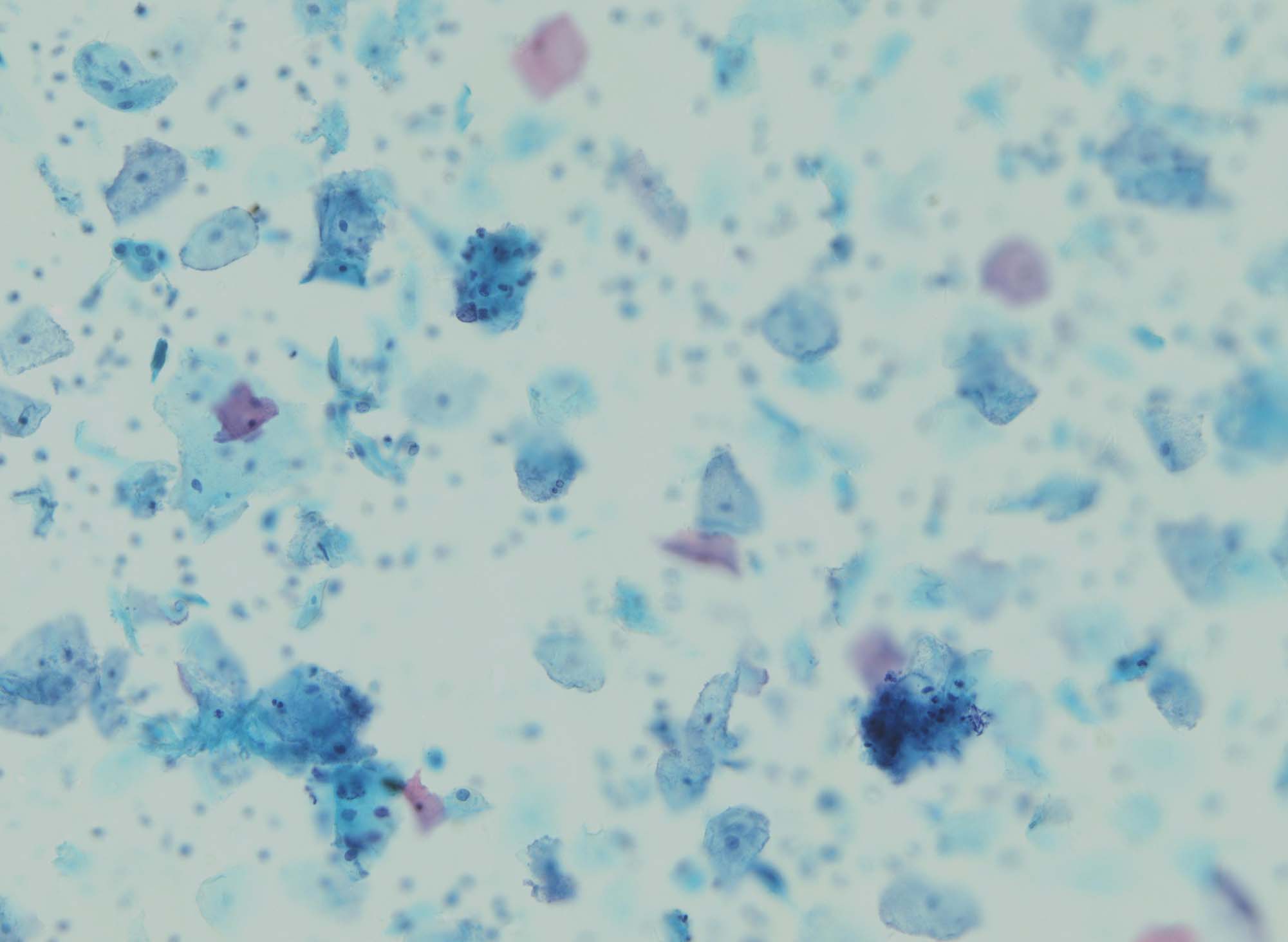}
        \caption{}   \label{09}
    \end{subfigure}\\
    \begin{subfigure}[b]{0.24\textwidth}
        \includegraphics[width=\textwidth]{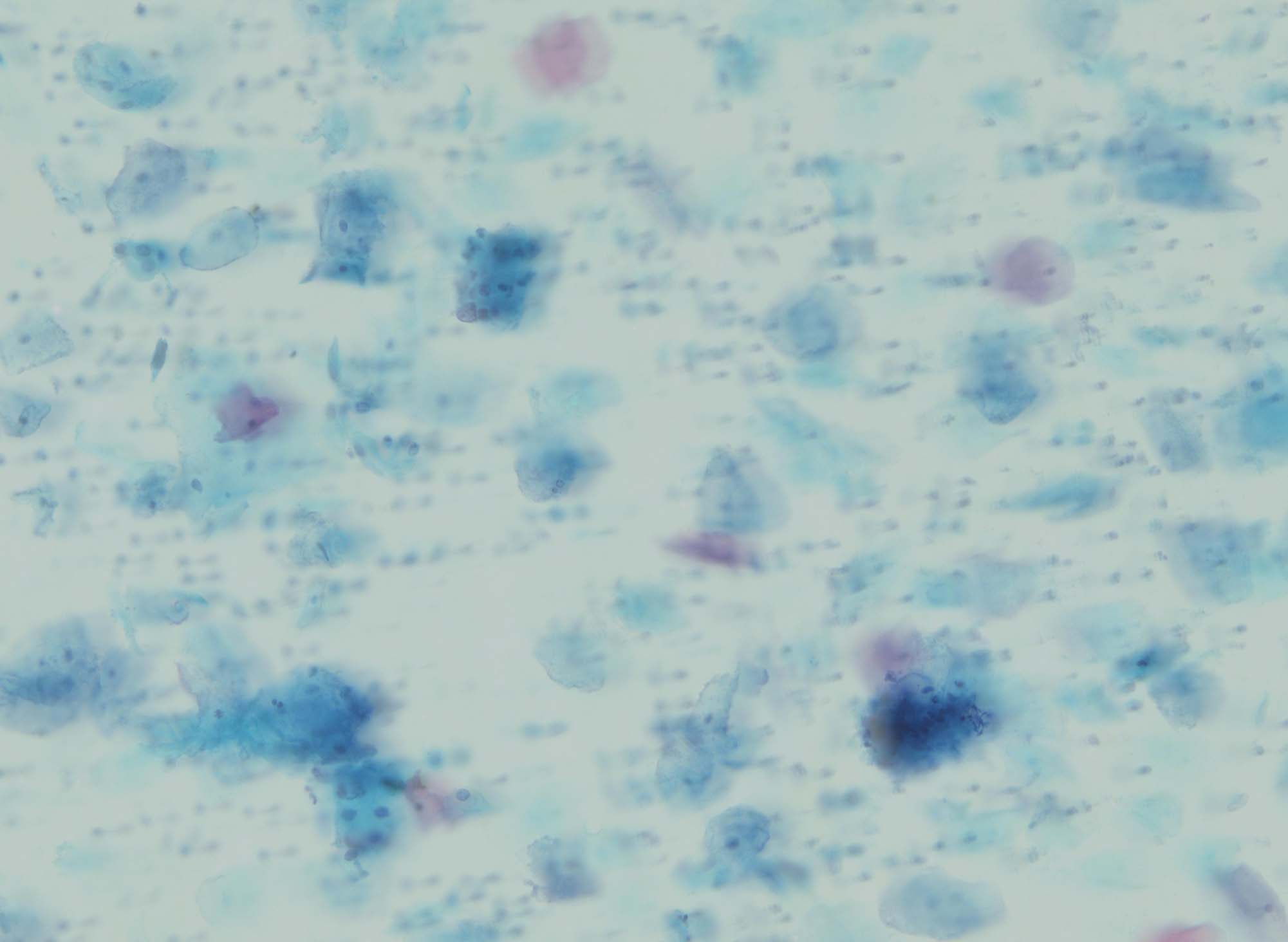}
        \caption{}   \label{DSIFT}
    \end{subfigure}
        \begin{subfigure}[b]{0.24\textwidth}
        \includegraphics[width=\textwidth]{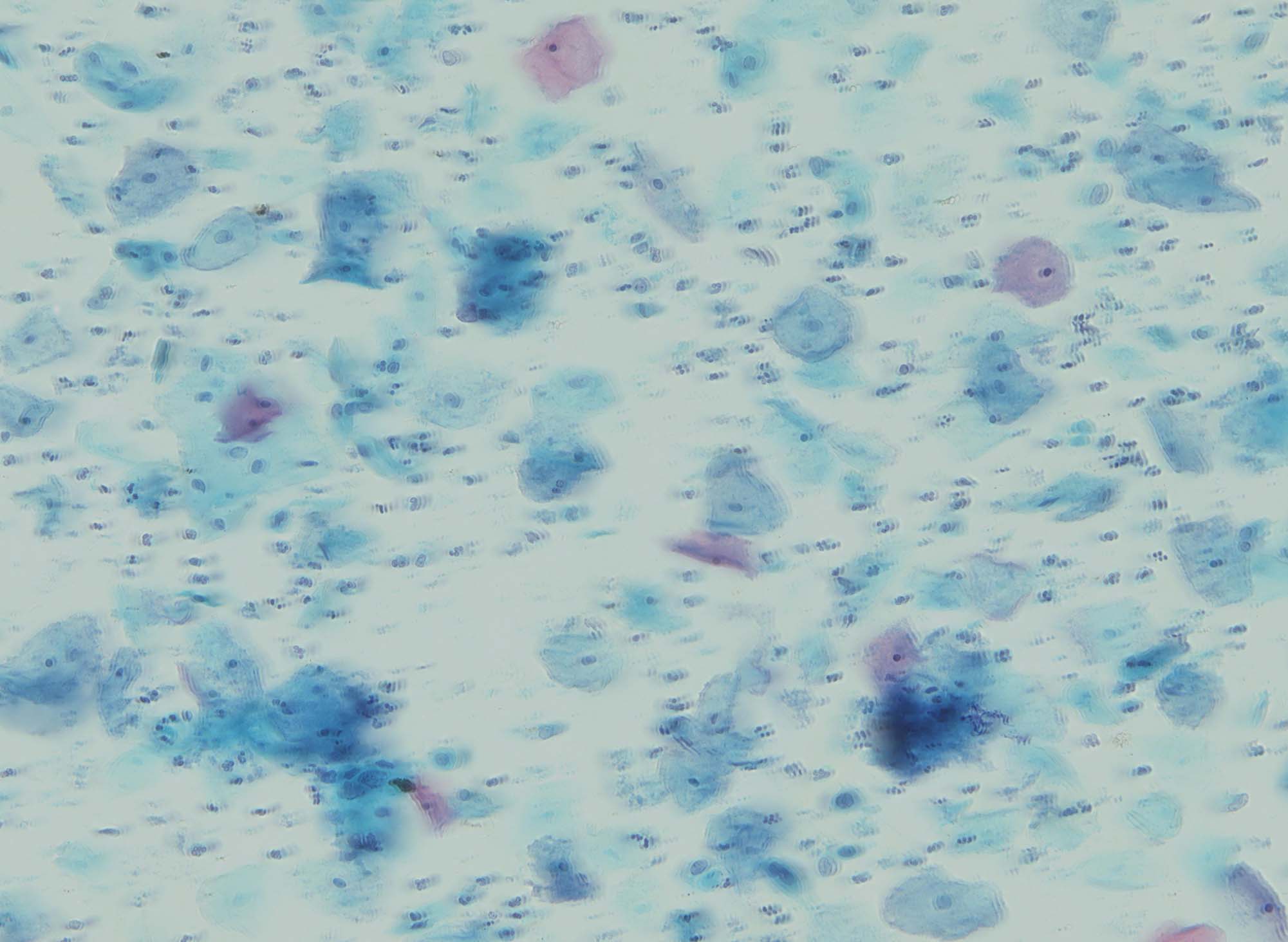}
        \caption{}  \label{GFF}
    \end{subfigure}
    \begin{subfigure}[b]{0.24\textwidth}
        \includegraphics[width=\textwidth]{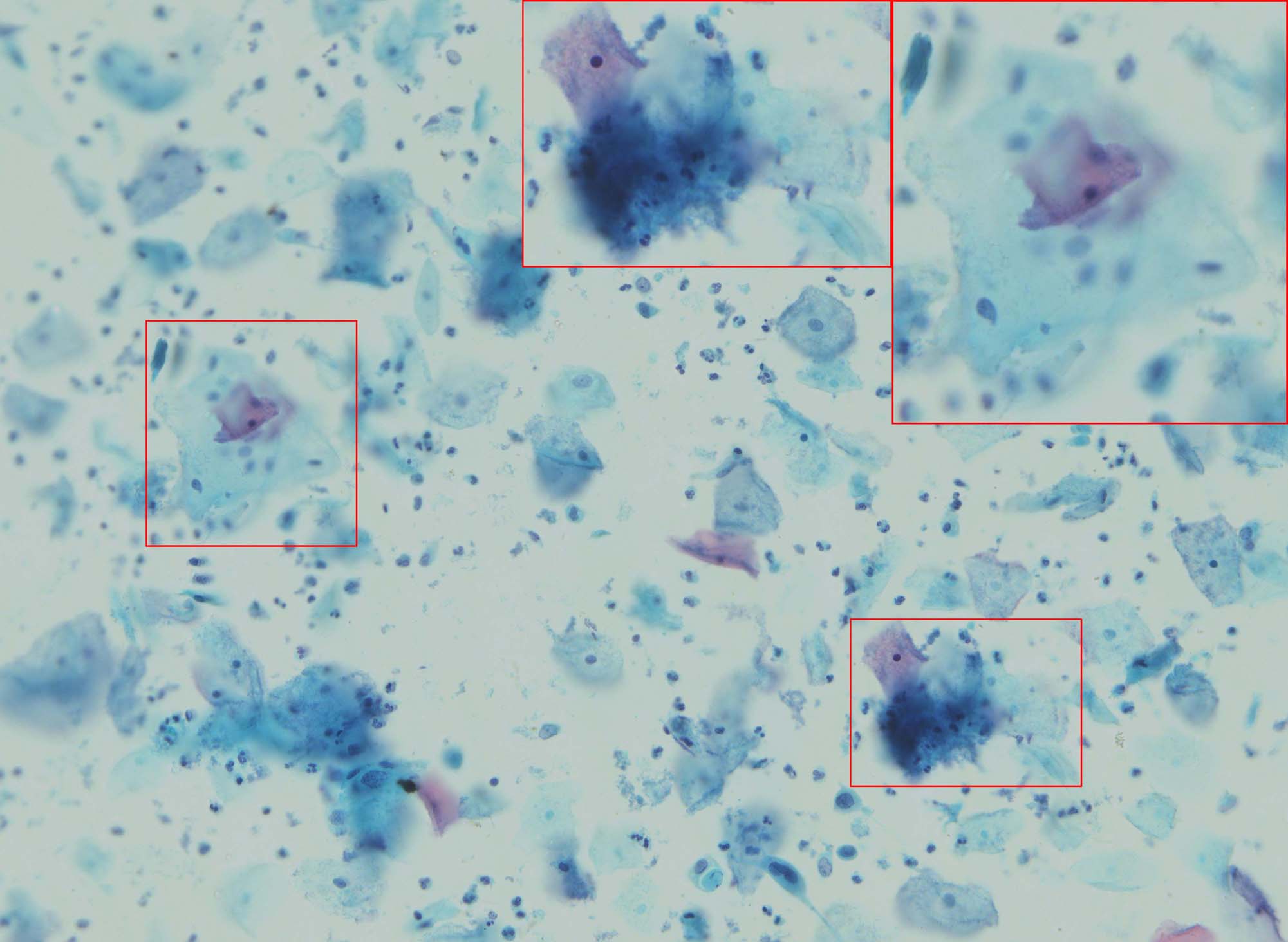}
        \caption{} \label{MWGF}
    \end{subfigure}
    \begin{subfigure}[b]{0.24\textwidth}
        \includegraphics[width=\textwidth]{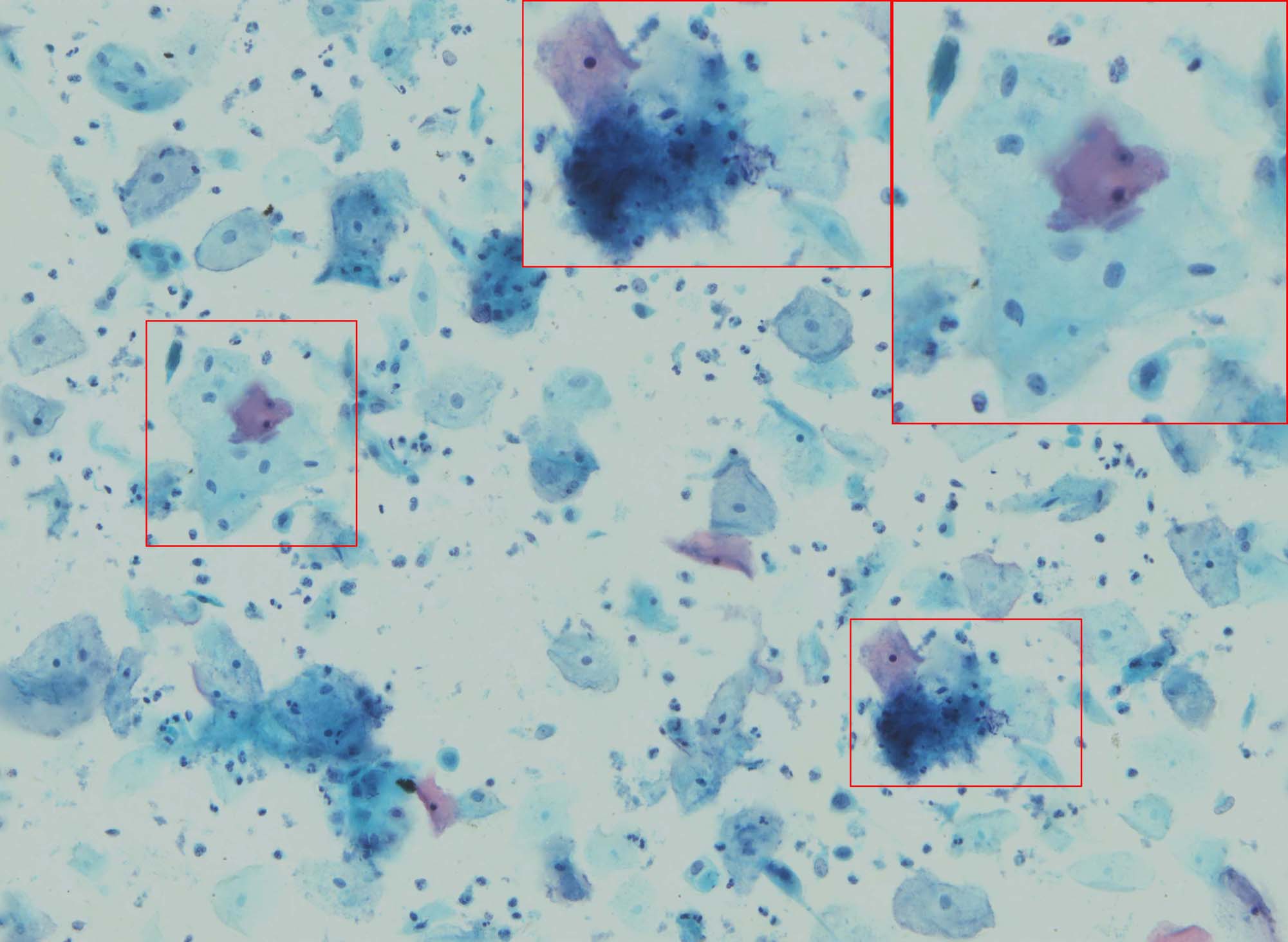}
        \caption{}   \label{PROPOSED}
    \end{subfigure}
    \caption{ The unregistered multi-focus source images and fusion results: (a)-(h) source images; (i) DSIFT; (j) GFF; (k) MWGF; (l) Ours. The regions in the upper right corner are the magnified view for comparison in (k) and (l).}\label{compare1}
\end{figure}

 \begin{figure}[t]
    \centering
    \begin{subfigure}[b]{0.24\textwidth}
        \includegraphics[width=\textwidth]{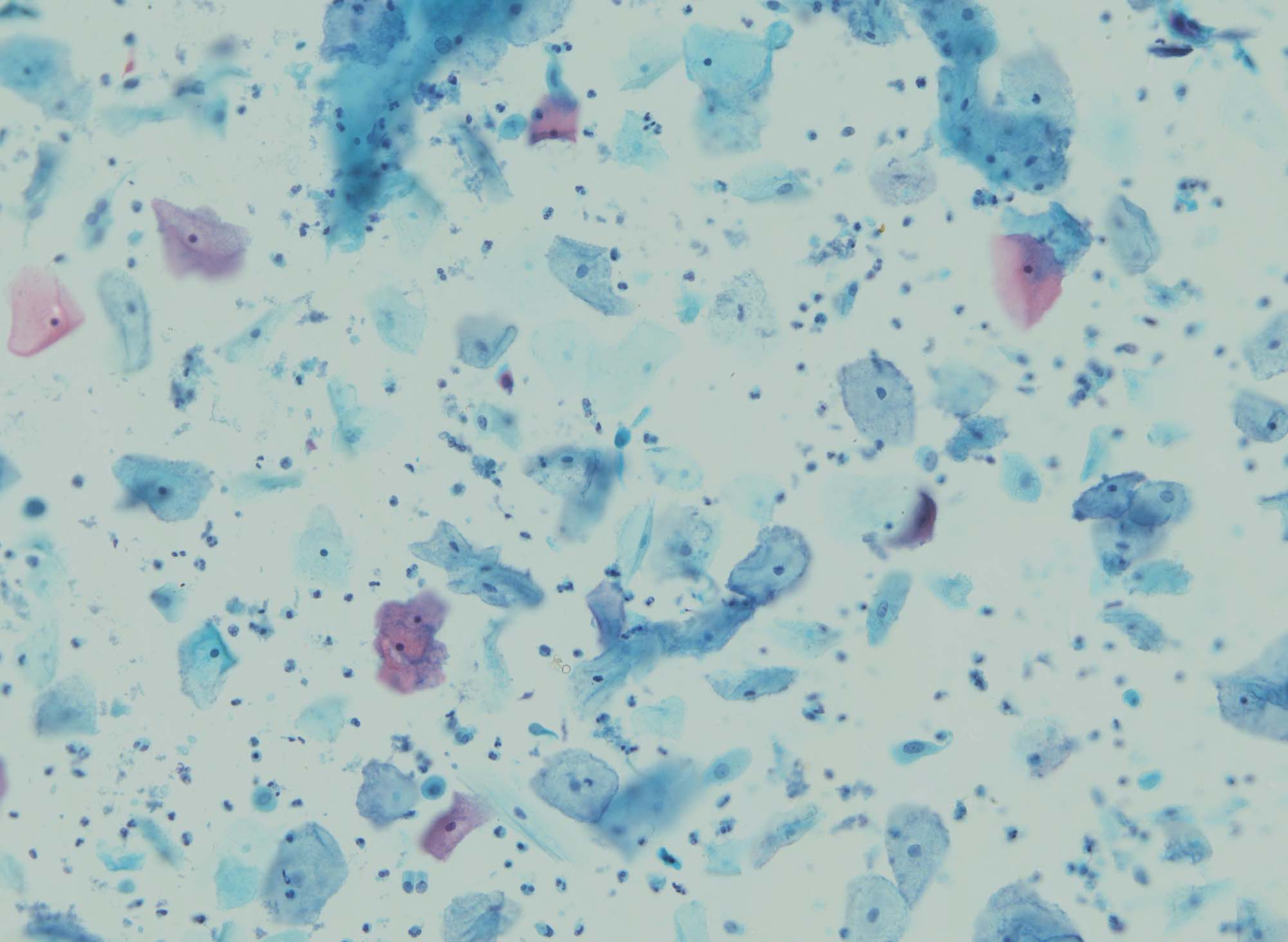}
        \caption{}    \label{11}
    \end{subfigure}
    \begin{subfigure}[b]{0.24\textwidth}
        \includegraphics[width=\textwidth]{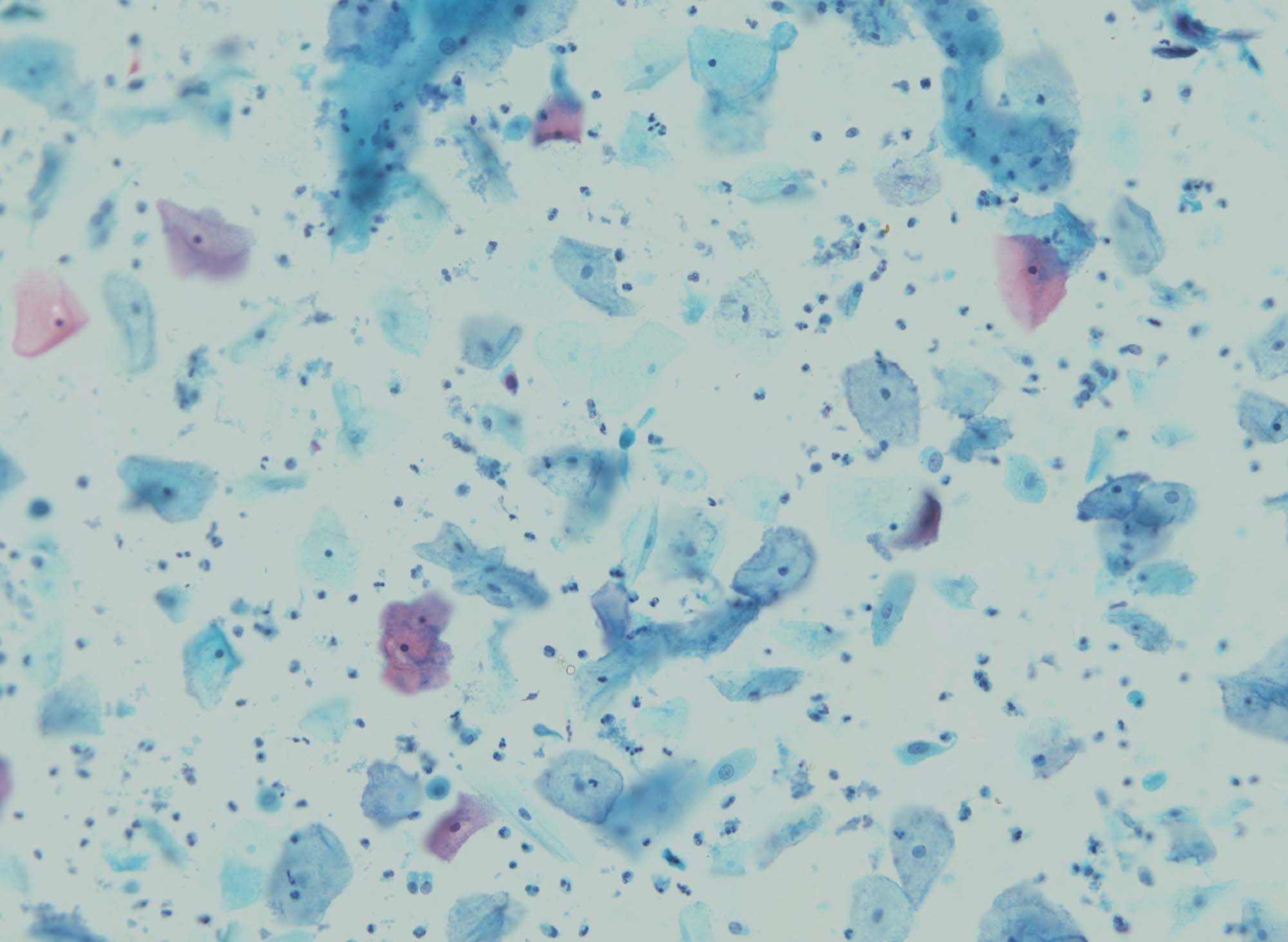}
        \caption{}   \label{12}
    \end{subfigure}
    \begin{subfigure}[b]{0.24\textwidth}
        \includegraphics[width=\textwidth]{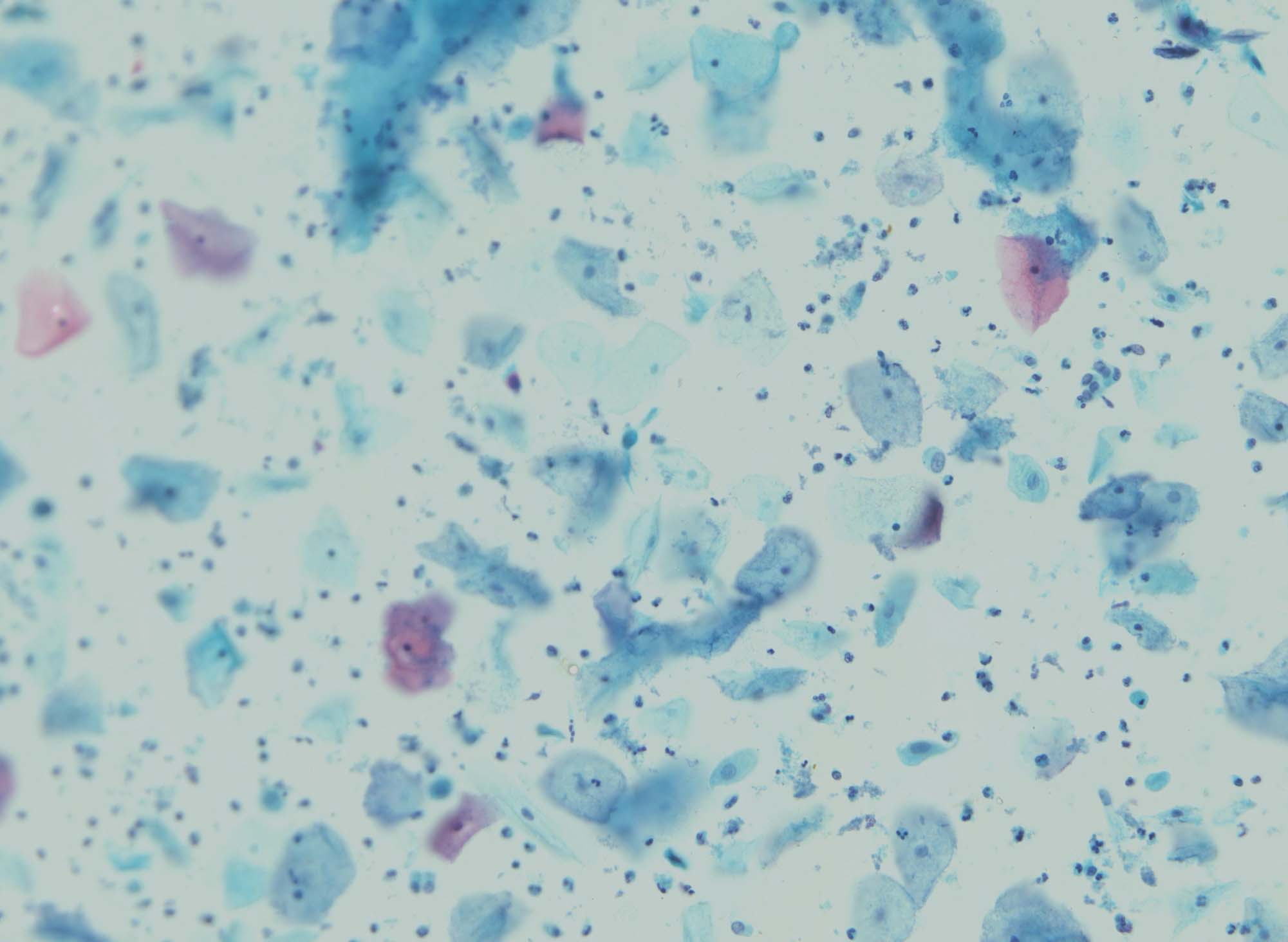}
        \caption{}   \label{13}
    \end{subfigure}
    \begin{subfigure}[b]{0.24\textwidth}
        \includegraphics[width=\textwidth]{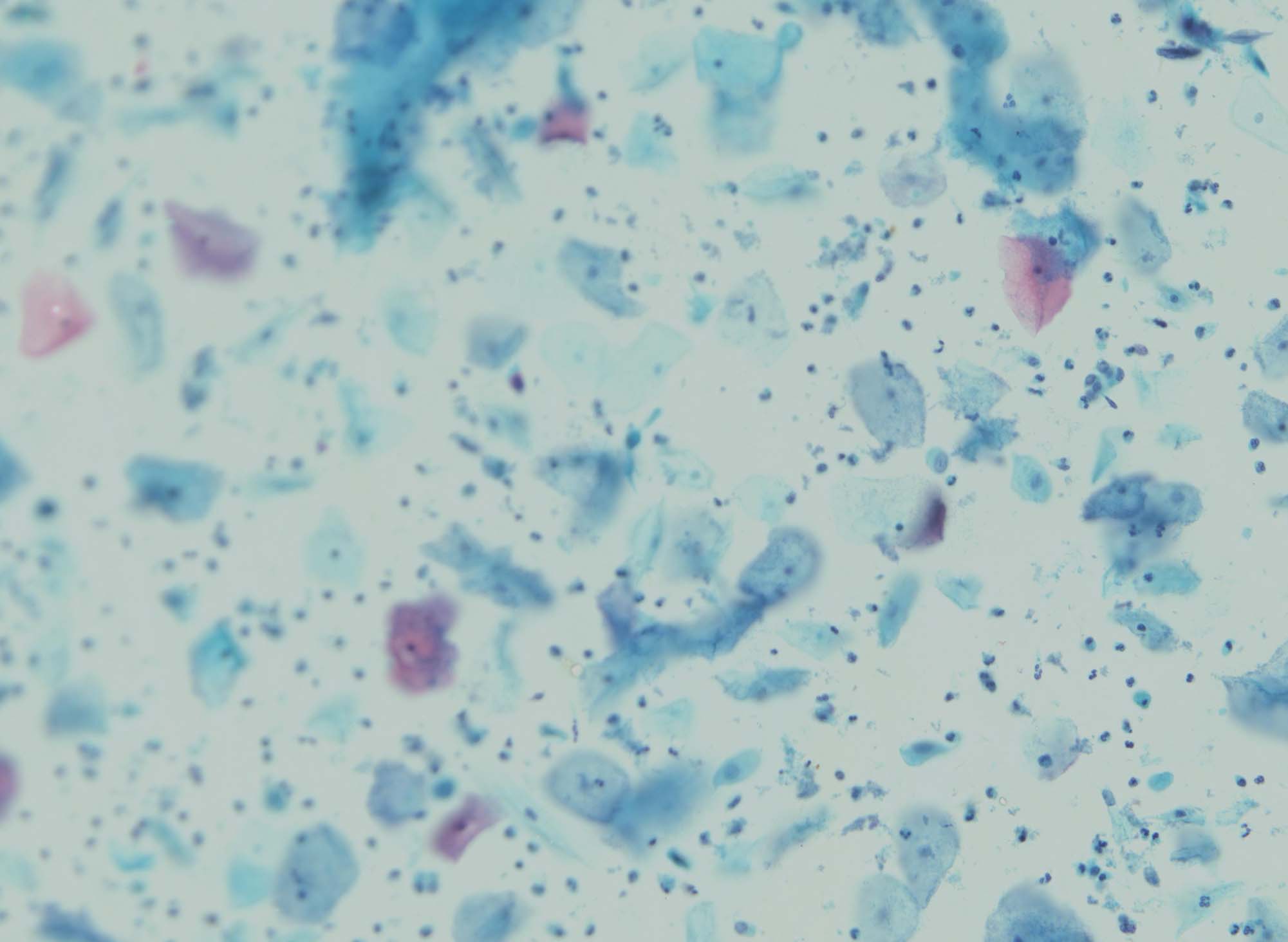}
        \caption{}   \label{14}
    \end{subfigure}\\
    \begin{subfigure}[b]{0.24\textwidth}
        \includegraphics[width=\textwidth]{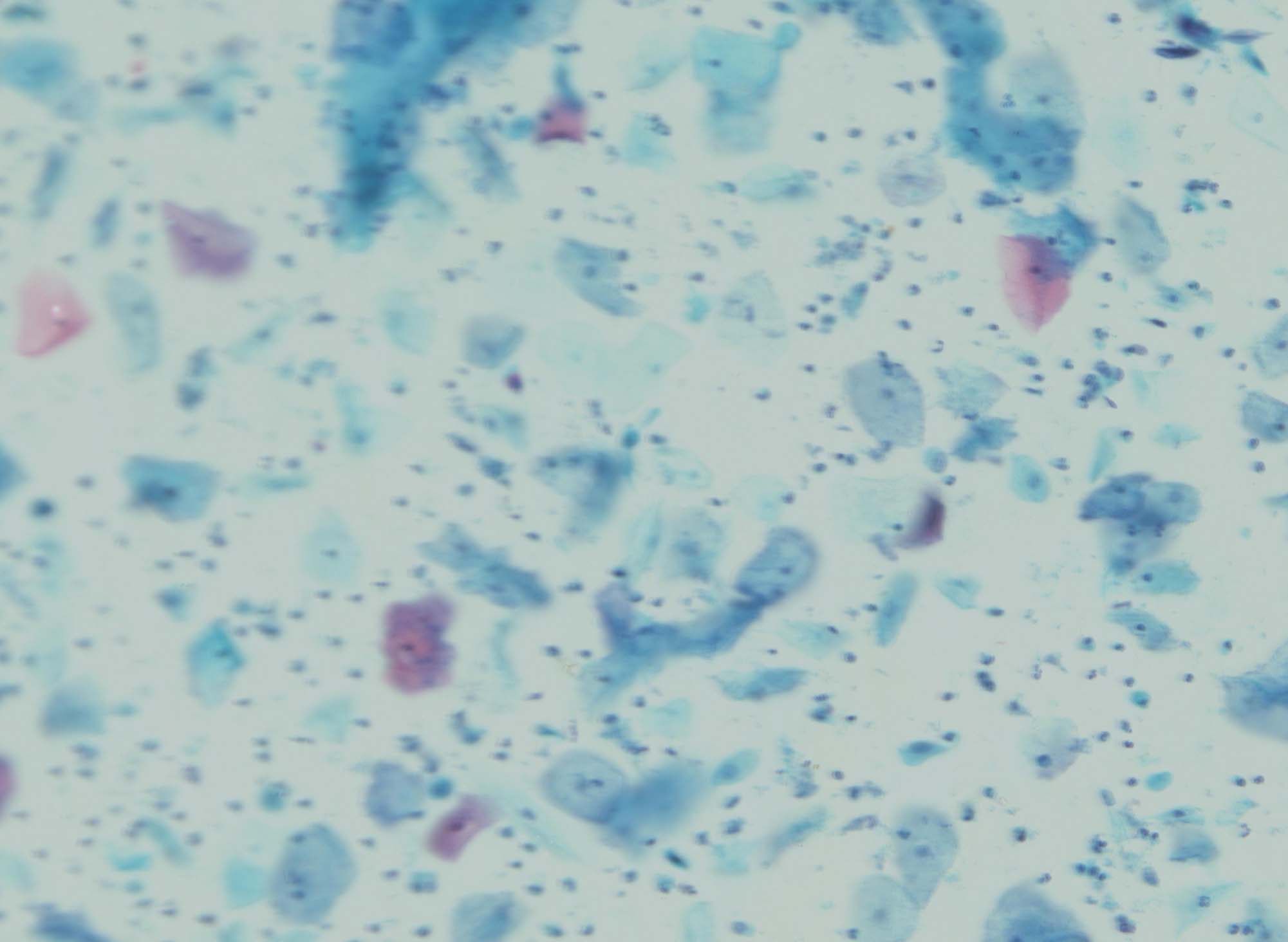}
        \caption{}   \label{1DSIFT}
    \end{subfigure}
        \begin{subfigure}[b]{0.24\textwidth}
        \includegraphics[width=\textwidth]{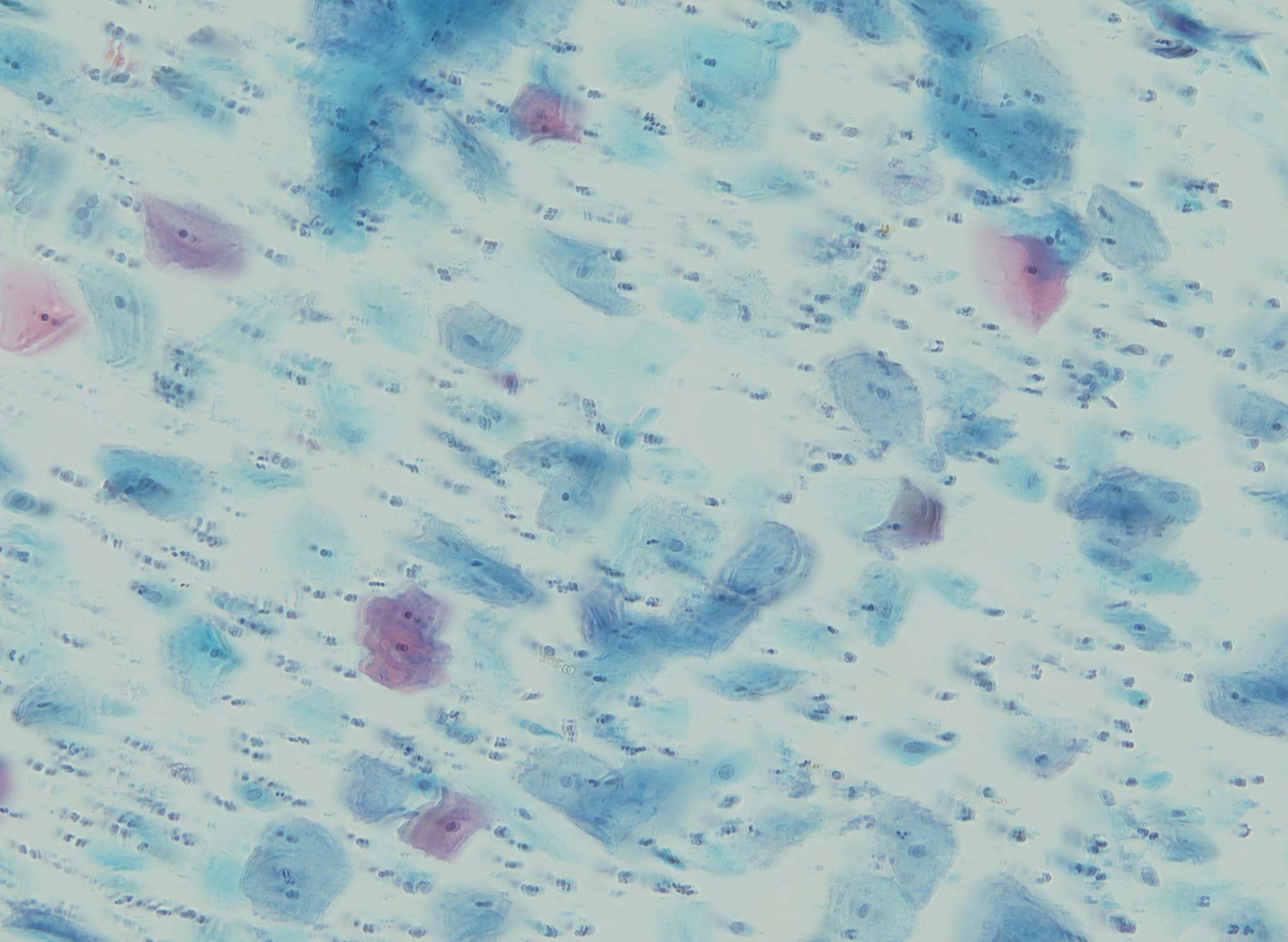}
        \caption{}  \label{1GFF}
    \end{subfigure}
    \begin{subfigure}[b]{0.24\textwidth}
        \includegraphics[width=\textwidth]{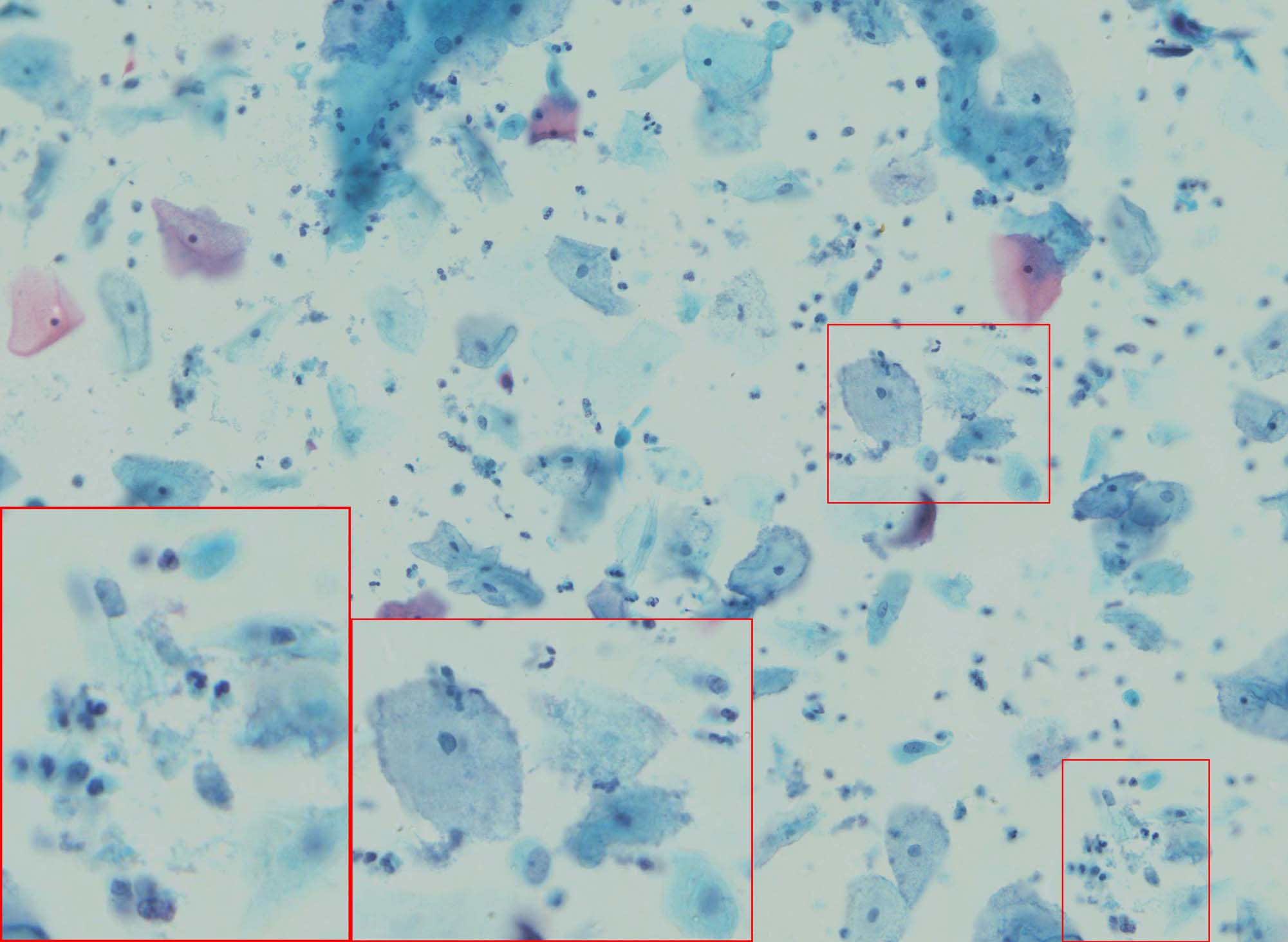}
        \caption{} \label{1MWGF}
    \end{subfigure}
    \begin{subfigure}[b]{0.24\textwidth}
        \includegraphics[width=\textwidth]{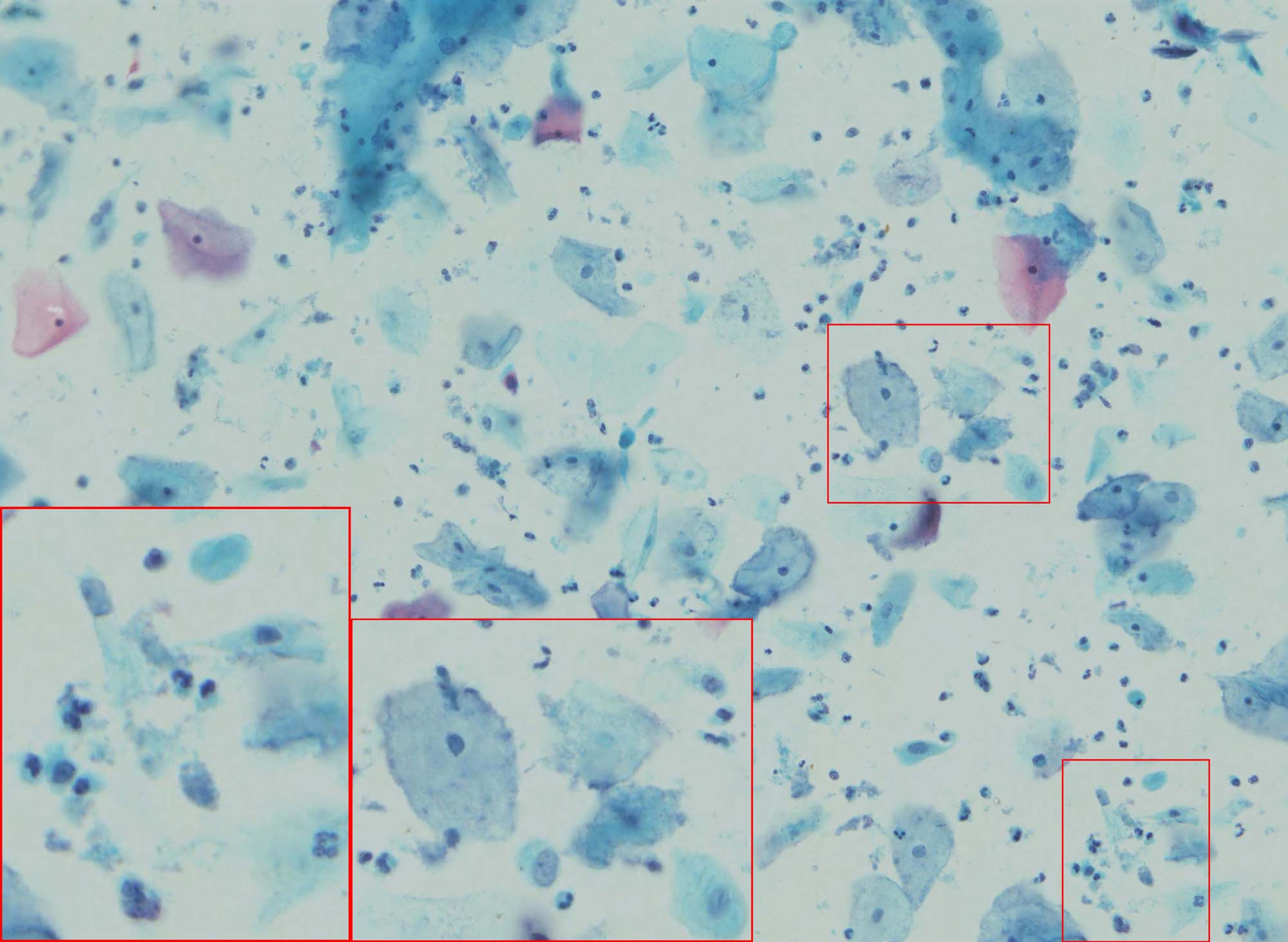}
        \caption{}   \label{1PROPOSED}
    \end{subfigure}

    \caption{ The unregistered multi-focus source images and fusion results: (a)-(d) source images; (e) DSIFT; (f) GFF; (g) MWGF and (h) ours. The regions in the bottom left corner are the magnified view for comparison in (g) and (h).}\label{compare2}
\end{figure}

\begin{figure}[t]
    \centering
    \begin{subfigure}[b]{0.24\textwidth}
        \includegraphics[width=\textwidth]{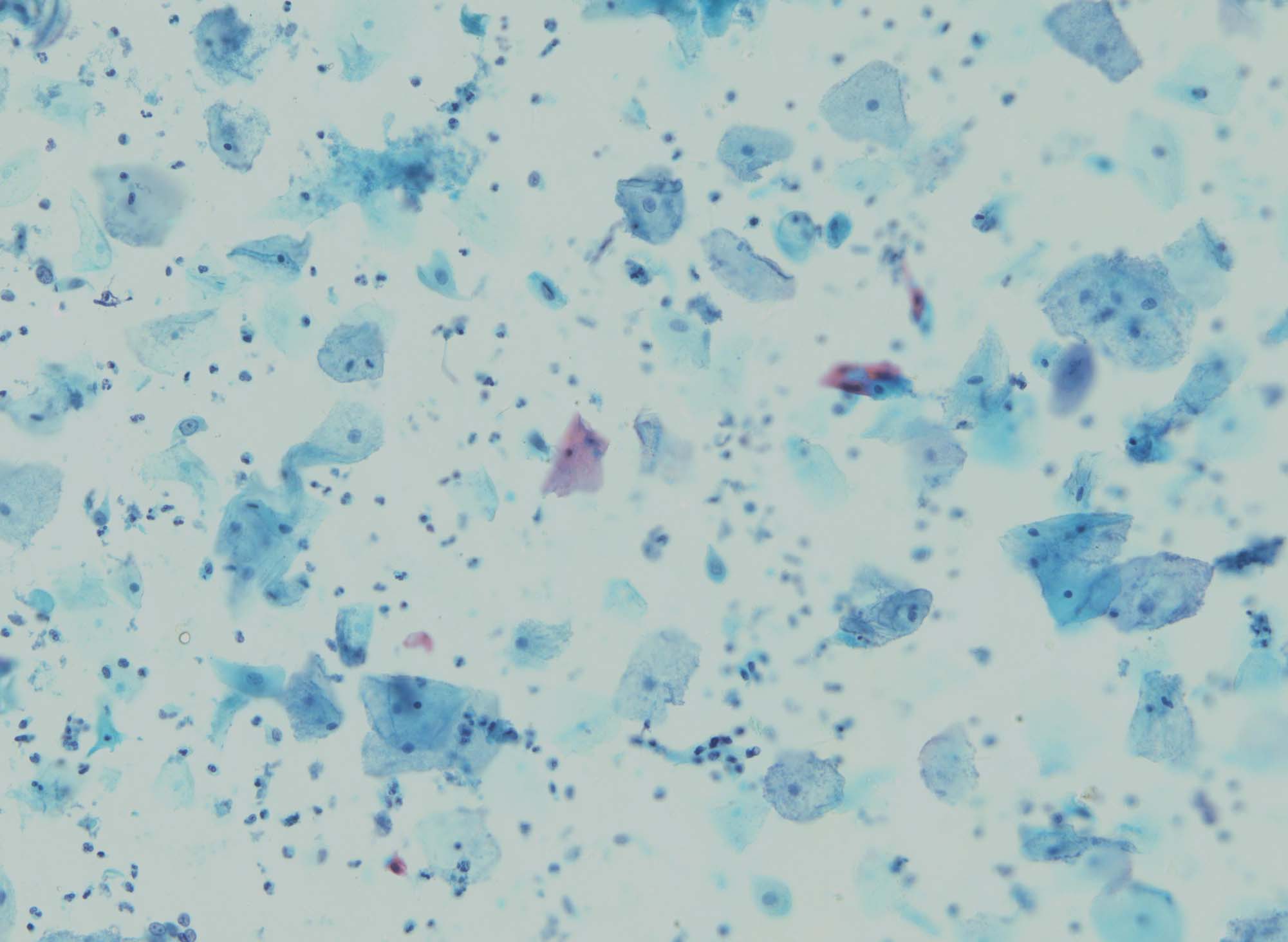}
        \caption{}    \label{21}
    \end{subfigure}
        \begin{subfigure}[b]{0.24\textwidth}
        \includegraphics[width=\textwidth]{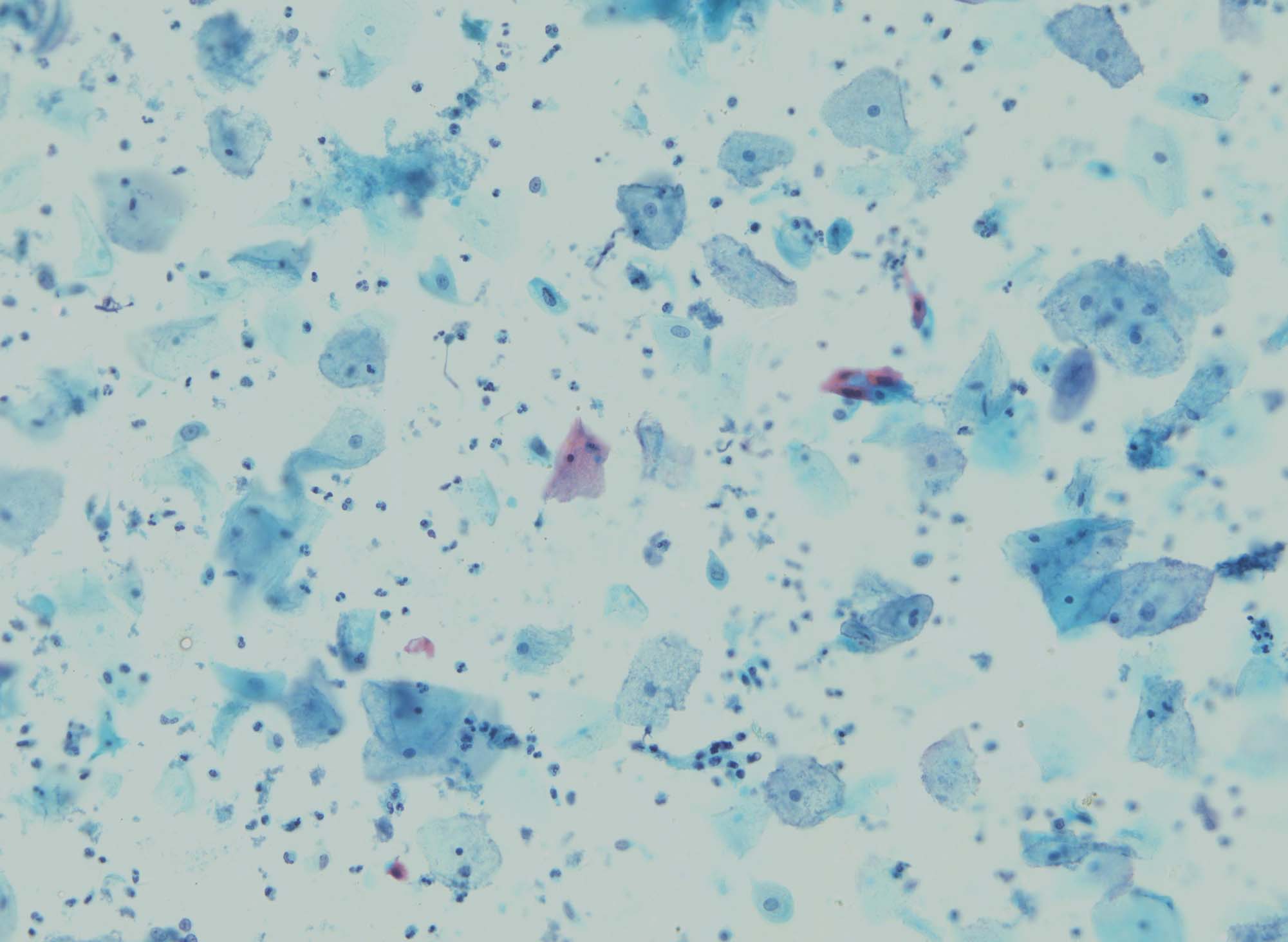}
        \caption{}   \label{22}
    \end{subfigure}
    \begin{subfigure}[b]{0.24\textwidth}
        \includegraphics[width=\textwidth]{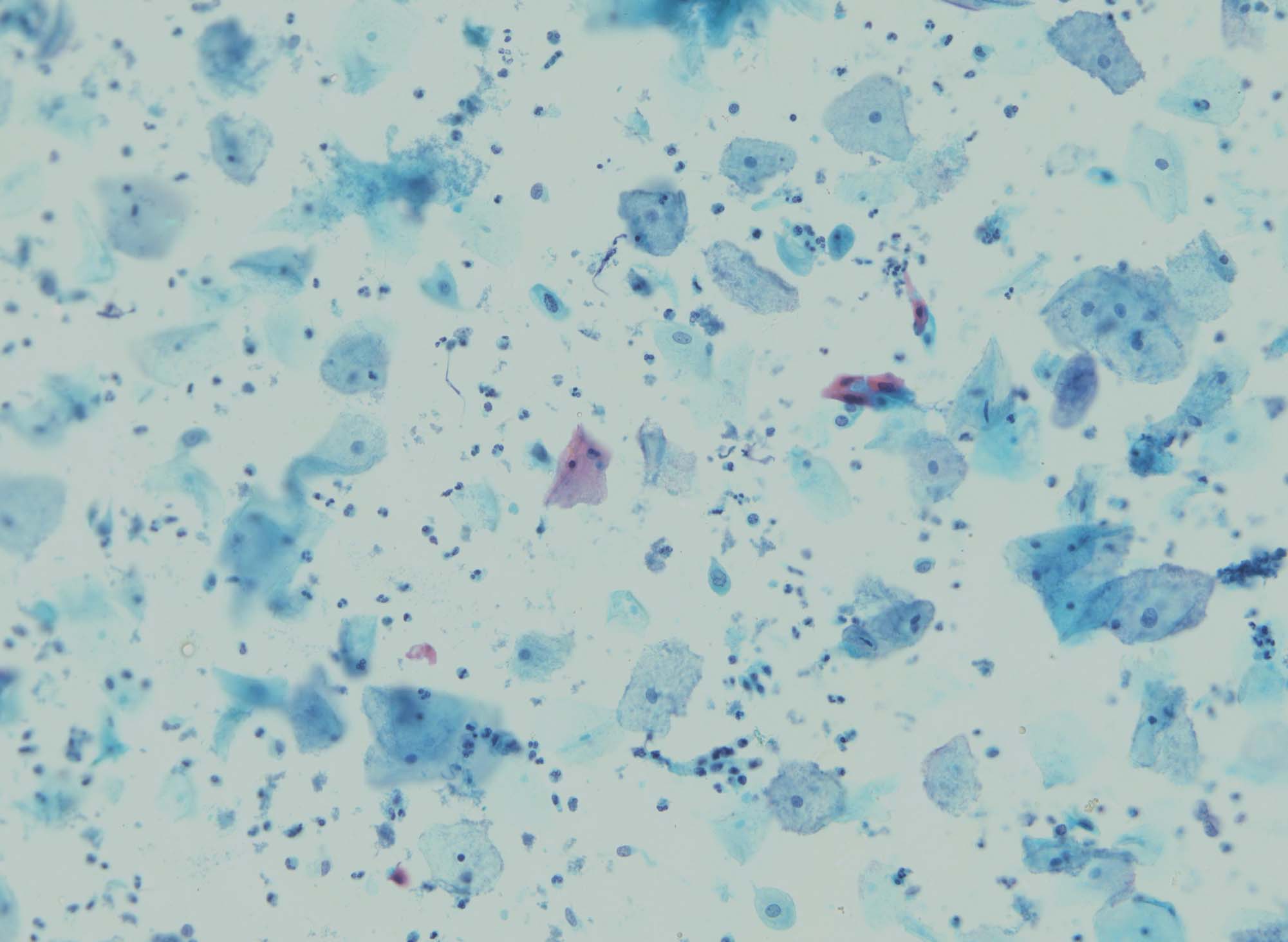}
        \caption{}   \label{23}
    \end{subfigure}
    \begin{subfigure}[b]{0.24\textwidth}
        \includegraphics[width=\textwidth]{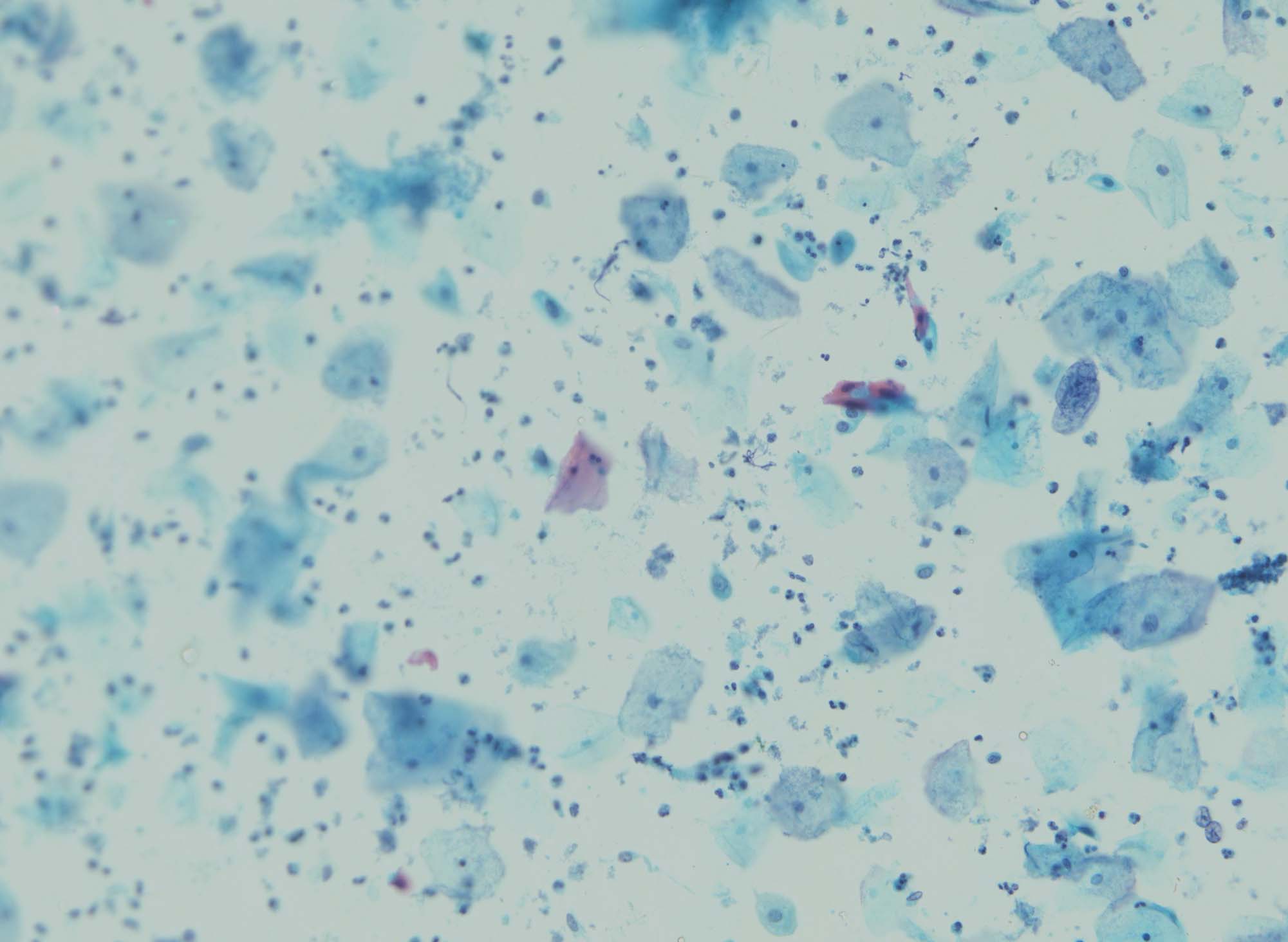}
        \caption{}   \label{24}
    \end{subfigure}\\
    \begin{subfigure}[b]{0.24\textwidth}
        \includegraphics[width=\textwidth]{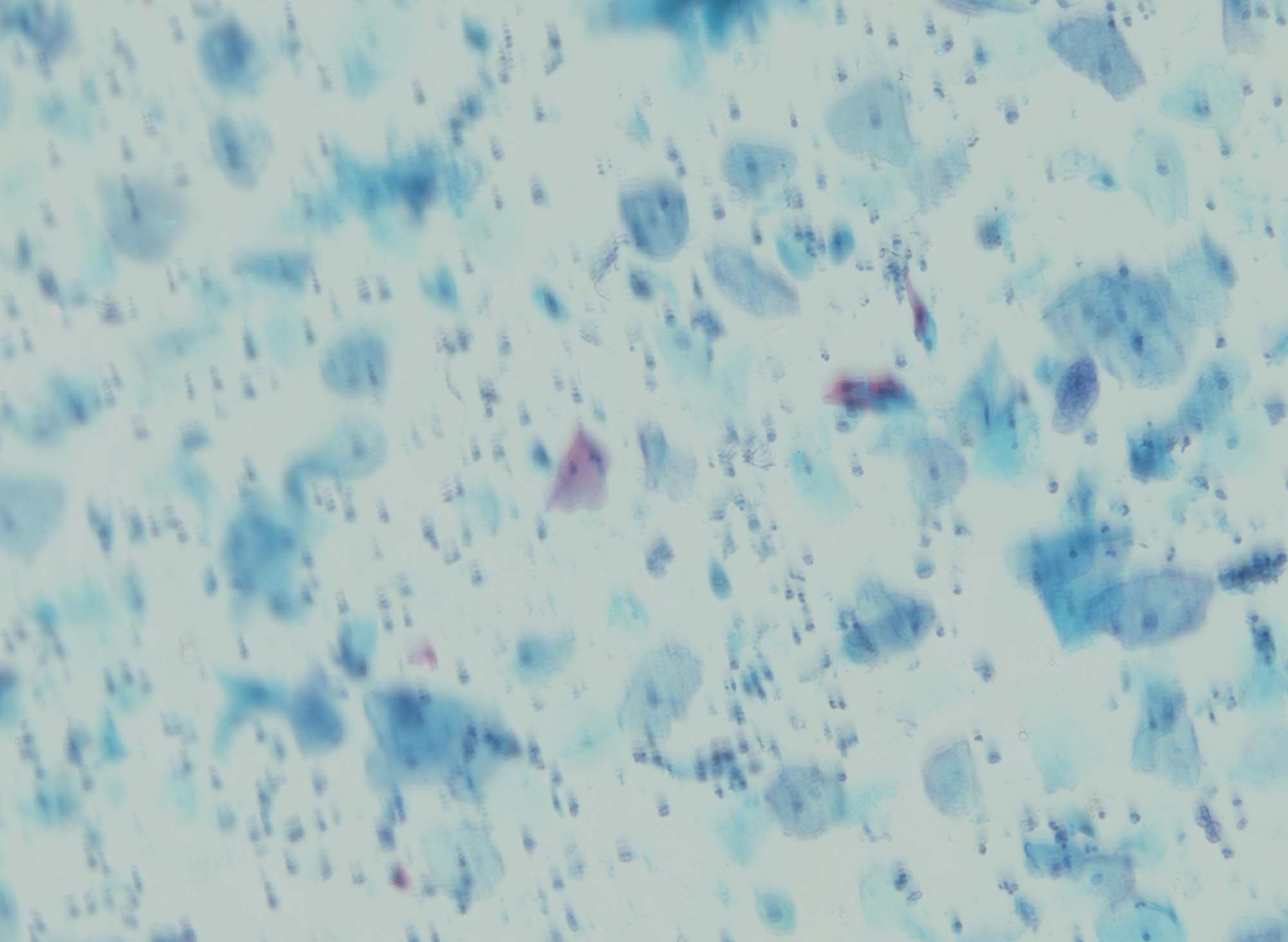}
        \caption{}   \label{2DSIFT}
    \end{subfigure}
    \begin{subfigure}[b]{0.24\textwidth}
        \includegraphics[width=\textwidth]{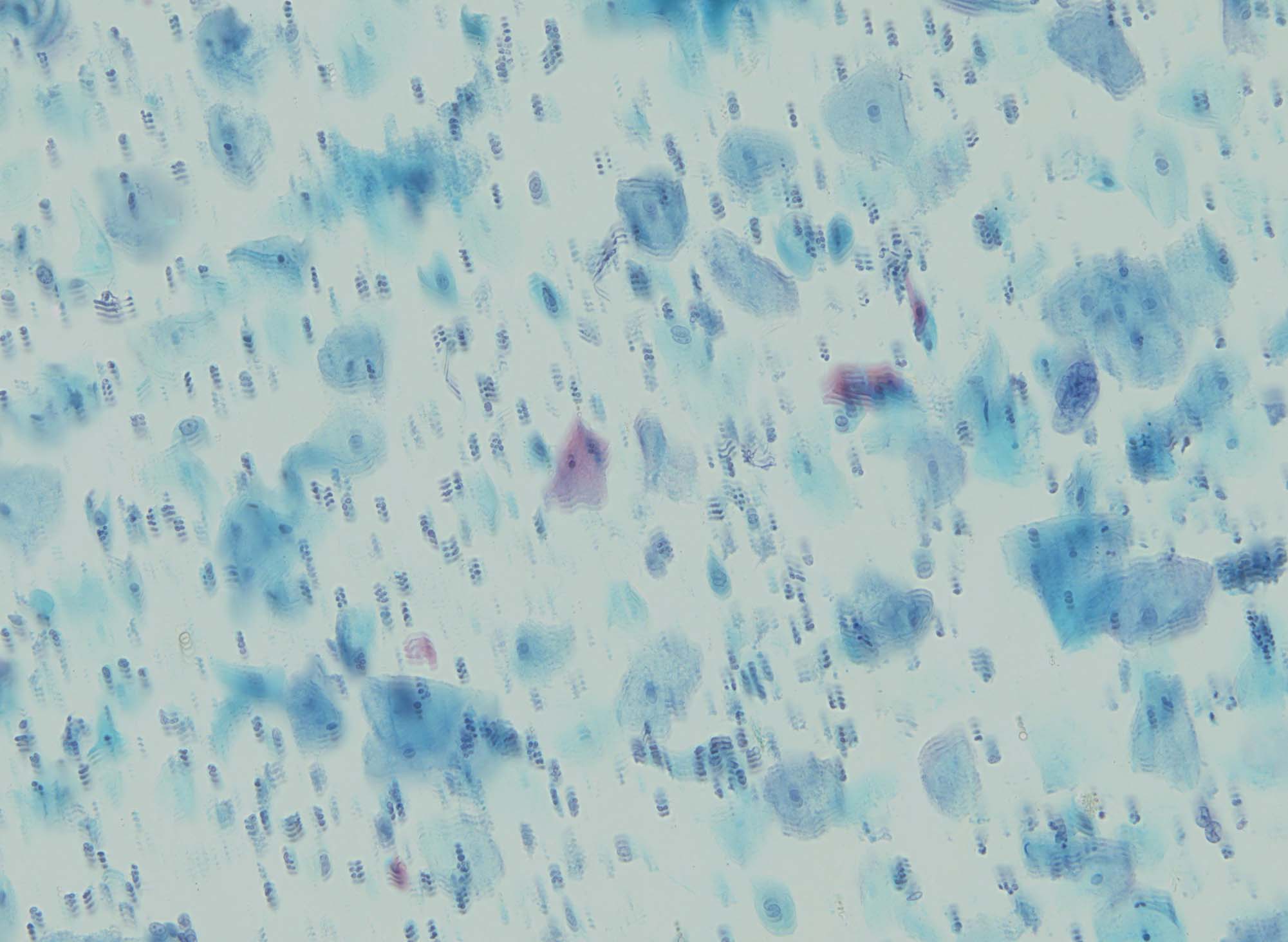}
        \caption{}  \label{2GFF}
    \end{subfigure}
    \begin{subfigure}[b]{0.24\textwidth}
        \includegraphics[width=\textwidth]{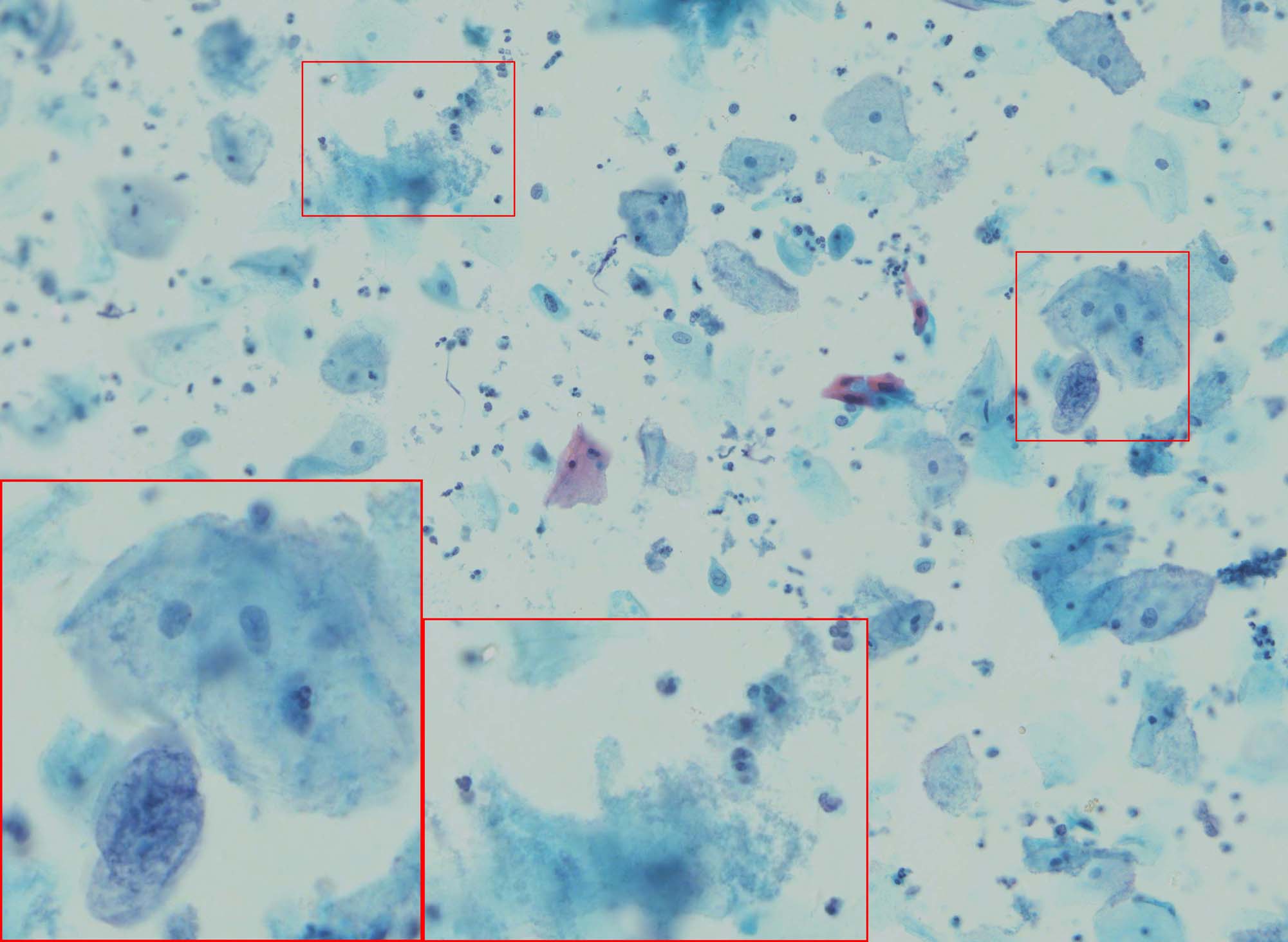}
        \caption{} \label{2MWGF}
    \end{subfigure}
    \begin{subfigure}[b]{0.24\textwidth}
        \includegraphics[width=\textwidth]{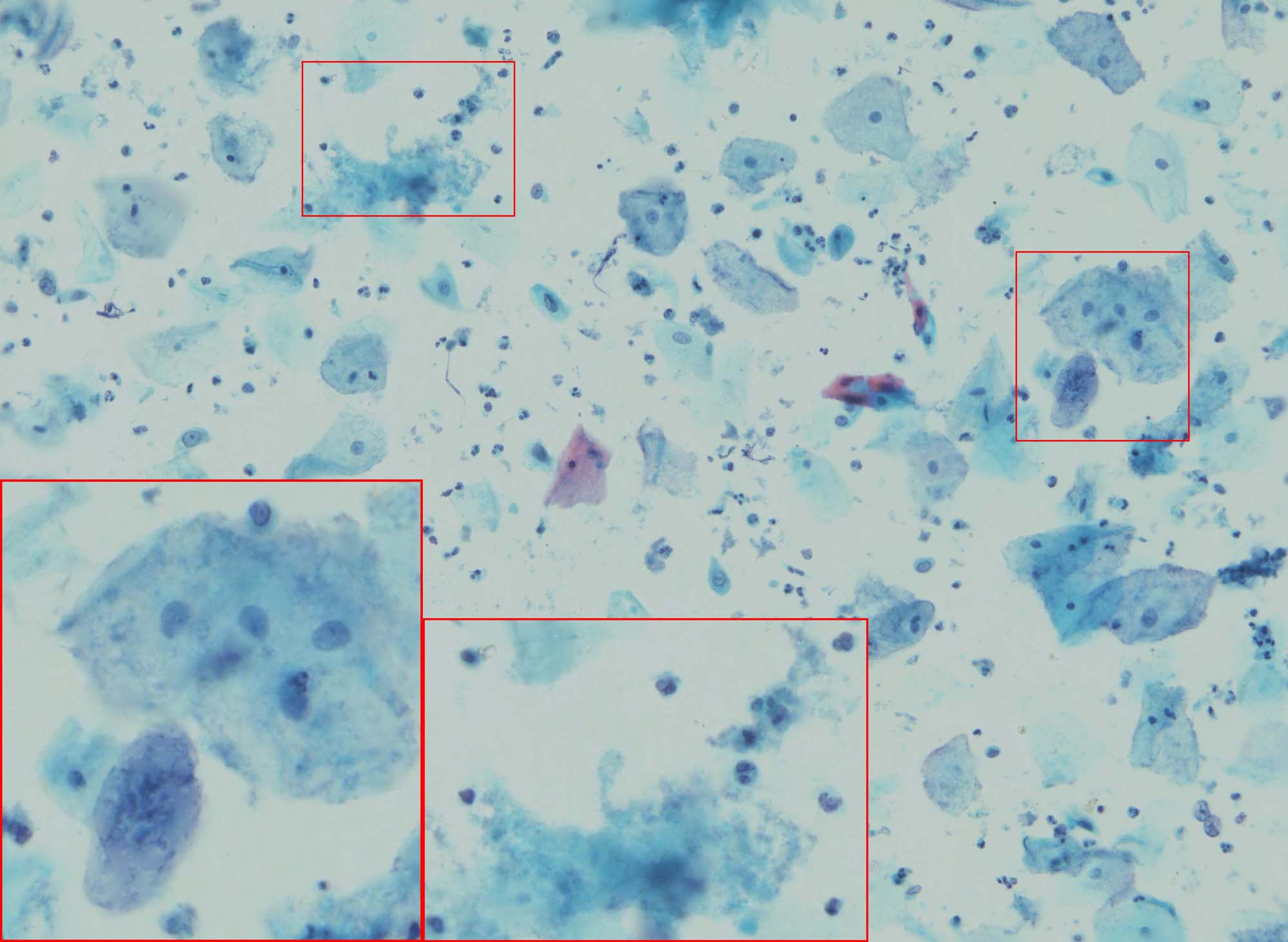}
        \caption{}   \label{2PROPOSED}
    \end{subfigure}
    \caption{ The multi-focus source images and fusion results: (a)-(d) source images; (e) is the result of DSIFT; (f) is the result GFF; (g) is the result of MWGF and (h) is the result of our method. The regions in the bottom left corner are the magnified view for comparison in (g) and (h).}\label{compare3}
\end{figure}

Figure \ref{compare1}, Figure \ref{compare2} and Figure \ref{compare3} show three groups of source images and the comparative results produced by different methods respectively. Each of them displays the sequence of unregistered source images and corresponding results by different methods. The fused results based on DSIFT are shown in Figure \ref{DSIFT}, \ref{1DSIFT} and \ref{2DSIFT}, which are extremely blurred obviously even worse than the input images. For the fused images by GFF, there are serious ghost effect that the information is destroyed from the input images, as shown in Figure \ref{GFF}, \ref{1GFF} and \ref{2GFF}. Figure \ref{MWGF}, \ref{1MWGF} and \ref{2MWGF} illustrate the results of MWGF, which show that the performance of them are better at a glimpse, but the performance is degraded by the ghost effect, where there are some geometrical distortion. And for clarity, we magnify several regions. Fortunately, as shown in Figure \ref{PROPOSED},\ref{1PROPOSED} and \ref{2PROPOSED}, the fused images are very salient by the proposed method. And the information of objects with very different size is reserved well that there are no artifacts and distortion.

We also compare the average running time by fusing the source images in Figure \ref{compare1}, Figure \ref{compare2} and Figure \ref{compare3} with these methods. We re-implement the GFF-based method with C++ to make a reference for different implementation. In addition, we also implement an intuitive approach which first performs registration and then fusion (RF) to validate the efficiency of our method \footnote{The RF method produces similar visual fused images as ours and therefore are not included due to the limited space.}. The experiments are conducted on a PC equipped with a 4.20GHz CPU and 8GB memory and the results are shown in Table \ref{compareTime}. It can be seen that our method is extremely faster than others, though they can be speeded up by almost 66$\%$ implemented in C++. And it can be furthermore accelerated through GPU programming to make it applicable for real-time applications such as digital cytopathology \cite{Pantanowitz2009The}.
\begin{table}[h]
  \centering
  \caption{The average running time of different fusion methods.}\label{compareTime}
 \begin{tabular}{rrrrrrr}
  \hline
  Methods &   DSIFT     & MWGF     &   GFF   & GFF(C++) & RF & Proposed\\
  \hline
  Time(s) &   2.27e+03  & 1.39e+03 & 1.05e+02& 3.57e+01 &2.17e+01&1.54e+01
  \\
  \hline
  \end{tabular}
\end{table}

\section{Conclusion}

In this paper, we first propose a novel method to fuse the unregistered microscopical multi-focus images in a very efficient way. Different from the existing methods in literature, the proposed method is base on the SURF scale space representation to eliminate the redundant operation between image registration and fusion. The scale-invariant saliency map can be generated while the repeatable feature points are detected simultaneously. Meanwhile, due to the robust scale space, the image registration is accurate and the information of very different size objects is reserved well in the fused image. The experiments investigate how the parameters
(i.e., the number of octaves and layers in the scale space and the dimension of the feature descriptor) influence the fusion results in our algorithm. Meanwhile, we compare our method with several state-of-the-art methods and demonstrate our method is superior to others in the efficiency and visual performance. However, there are still some challenges that should be further explored, e.g., making the algorithm more robust to be applied to different practice and promoting the calculation to make it real-time.

\section*{Acknowledgements}
This research was partially supported by the Natural Science Foundation of Hunan Province, China (No.14JJ2008), the National Natural Science Foundation of China under Grant No. 61602522, No. 61573380, No. 61672542 and the Fundamental Research Funds of the Central Universities of Central South University under Grant No. 2018zzts577.

\section*{References}
\bibliographystyle{plainnat}
\bibliography{reference2}

\begin{thebibliography}{10}
\expandafter\ifx\csname url\endcsname\relax
  \def\url#1{\texttt{#1}}\fi
\expandafter\ifx\csname urlprefix\endcsname\relax\def\urlprefix{URL }\fi
\expandafter\ifx\csname href\endcsname\relax
  \def\href#1#2{#2} \def\path#1{#1}\fi

\bibitem{miao2011novel}
Q.-g. Miao, C.~Shi, P.-f. Xu, M.~Yang, Y.-b. Shi, A novel algorithm of image
  fusion using shearlets, Optics Communications 284~(6) (2011) 1540--1547.

\bibitem{li2017pixel}
S.~Li, X.~Kang, L.~Fang, J.~Hu, H.~Yin, Pixel-level image fusion: A survey of
  the state of the art, Information Fusion 33 (2017) 100--112.

\bibitem{li2013imageGFF}
S.~Li, X.~Kang, J.~Hu, Image fusion with guided filtering, IEEE Transactions on
  Image Processing 22~(7) (2013) 2864--2875.

\bibitem{liu2015multi}
Y.~Liu, S.~Liu, Z.~Wang, Multi-focus image fusion with dense sift, Information
  Fusion 23 (2015) 139--155.

\bibitem{zhou2014multi}
Z.~Zhou, S.~Li, B.~Wang, Multi-scale weighted gradient-based fusion for
  multi-focus images, Information Fusion 20 (2014) 60--72.

\bibitem{liu2015automatic}
Y.~Liu, F.~Yu, An automatic image fusion algorithm for unregistered multiply
  multi-focus images, Optics Communications 341 (2015) 101--113.

\bibitem{laliberte2003registration}
F.~Lalibert{\'e}, L.~Gagnon, Y.~Sheng, Registration and fusion of retinal
  images-an evaluation study, IEEE Transactions on Medical Imaging 22~(5)
  (2003) 661--673.

\bibitem{kessler2006image}
M.~L. Kessler, Image registration and data fusion in radiation therapy, The
  British journal of radiology 79~(special\_issue\_1) (2006) S99--S108.

\bibitem{Gong2014A}
M.~Gong, S.~Zhao, L.~Jiao, D.~Tian, S.~Wang, A novel coarse-to-fine scheme for
  automatic image registration based on sift and mutual information, IEEE
  Transactions on Geoscience and Remote Sensing 52~(7) (2014) 4328--4338.

\bibitem{Guo2017Automatic}
F.~Guo, X.~Zhao, B.~Zou, Y.~Liang, Automatic retinal image registration using
  blood vessel segmentation and sift feature, International Journal of Pattern
  Recognition and Artificial Intelligence 31~(11) (2017) 10.

\bibitem{bay2006surf}
H.~Bay, T.~Tuytelaars, L.~Van~Gool, Surf: Speeded up robust features, in:
  European conference on computer vision, Springer, 2006, pp. 404--417.

\bibitem{zitova2003image}
B.~Zitova, J.~Flusser, Image registration methods: a survey, Image and vision
  computing 21~(11) (2003) 977--1000.

\bibitem{goshtasby20052}
A.~A. Goshtasby, 2-D and 3-D image registration: for medical, remote sensing,
  and industrial applications, John Wiley \& Sons, 2005.

\bibitem{hill2001medical}
D.~L. Hill, P.~G. Batchelor, M.~Holden, D.~J. Hawkes, Medical image
  registration, Physics in medicine \& biology 46~(3) (2001) R1.

\bibitem{goshtasby2007image}
A.~A. Goshtasby, S.~Nikolov, Image fusion: advances in the state of the art
  (2007).

\bibitem{pajares2004wavelet}
G.~Pajares, J.~M. De~La~Cruz, A wavelet-based image fusion tutorial, Pattern
  recognition 37~(9) (2004) 1855--1872.

\bibitem{liu2015general}
Y.~Liu, S.~Liu, Z.~Wang, A general framework for image fusion based on
  multi-scale transform and sparse representation, Information Fusion 24 (2015)
  147--164.

\bibitem{aslantas2014pixel}
V.~Aslantas, A.~N. Toprak, A pixel based multi-focus image fusion method,
  optics communications 332 (2014) 350--358.

\bibitem{Liu2018Deep}
Y.~Liu, X.~Chen, Z.~Wang, Z.~J. Wang, R.~K. Ward, X.~Wang, Deep learning for
  pixel-level image fusion: Recent advances and future prospects, Information
  Fusion.

\bibitem{chen2015sirf}
C.~Chen, Y.~Li, W.~Liu, J.~Huang, Sirf: simultaneous satellite image
  registration and fusion in a unified framework, IEEE Transactions on Image
  Processing 24~(11) (2015) 4213--4224.

\bibitem{chen2011maximum}
S.~Chen, Q.~Guo, H.~Leung, E.~Bosse, A maximum likelihood approach to joint
  image registration and fusion, IEEE Transactions on Image Processing 20~(5)
  (2011) 1363--1372.

\bibitem{ofir2017registration}
N.~Ofir, S.~Silberstein, D.~Rozenbaum, S.~D. Bar, Registration and fusion of
  multi-spectral images using a novel edge descriptor, arXiv preprint
  arXiv:1711.01543.

\bibitem{He2013Guided}
K.~He, J.~Sun, X.~Tang, Guided image filtering, IEEE Transactions on Pattern
  Analysis and Machine Intelligence 35~(6) (2013) 1397--1409.

\bibitem{Fischler1981Random}
M.~A. Fischler, R.~C. Bolles, Random sample consensus: a paradigm for model
  fitting with applications to image analysis and automated cartography, ACM,
  1981.

\bibitem{Pantanowitz2009The}
L.~Pantanowitz, M.~Hornish, R.~A. Goulart, The impact of digital imaging in the
  field of cytopathology, Cytojournal 6~(1) (2009) 59--70.

\end{thebibliography}

\end{document}